\documentclass[ShortAfour,twocolumn]{sagej}

\usepackage{moreverb,url}
\usepackage{verbatim}
\usepackage{graphicx}
\usepackage{gensymb}
\usepackage{color,soul}
\usepackage{subcaption}
\usepackage{rotating}
\usepackage{multirow}
\usepackage[round,authoryear]{natbib}
\usepackage{tikz}
\usetikzlibrary{arrows,shapes,calc}
\usepackage{colortbl}
\usepackage{cleveref}
\usepackage{pdfpages}

\def\volumeyear{2017}
\def\journalname{The International Journal of Robotics Research}

\begin{document}

\runninghead{Schenck and Fox}

\title{Perceiving and Reasoning About Liquids Using Fully Convolutional Networks}

\author{Connor Schenck\affilnum{1} and Dieter Fox\affilnum{1}}

\affiliation{
  {\bf Acknowledgments:} \\
  This work was funded in part by the National Science Foundation under contract
  number NSF-NRI-1525251. We would also like to thank Sudharsan Prabu for
  helping to collect data for this paper. \\
  \\
  \affilnum{1}Paul G.~Allen School for Computer Science \& Engineering, University of Washington, Seattle, WA, USA}

\corrauth{Connor Schenck}

\email{schenckc@cs.washington.edu}

\begin{abstract}
  Liquids are an important part of many common manipulation tasks in human
  environments.  If we wish to have robots that can accomplish these types of
  tasks, they must be able to interact with liquids in an intelligent manner.
  In this paper, we investigate ways for robots to perceive and reason about
  liquids.  That is, a robot asks the questions {\it What in the visual data
    stream is liquid?} and {\it How can I use that to infer all the potential
    places where liquid might be?}  We collected two datasets to evaluate these
  questions, one using a realistic liquid simulator and another on our robot.
  We used fully convolutional neural networks to learn to detect and track
  liquids across pouring sequences.  Our results show that these networks are
  able to perceive and reason about liquids, and that integrating temporal
  information is important to performing such tasks well.
\end{abstract}


\maketitle

\section{Introduction}

% Motivate
Liquids are ubiquitous in human environments.  Humans perform many of their
daily actions using liquids, whether it is pouring coffee, mixing ingredients
for a recipe, or washing their hands.  Any general purpose robot that will
operate in a human environment should be able to robustly handle liquids.
This task poses different challenges than object manipulation, since
liquids follow complicated dynamics and aren't necessarily divisible into
well-defined objects.

Before a robot can even begin to manipulate liquids, it first must be able
to perceive and reason about them.  For example, solving tasks such as pouring
requires both robust control {\it and} detection of liquid during the pouring
operation.  Thus, controlling liquids requires close closed-loop sensory
feedback to perform well.  This is a difficult problem in itself.  For example,
many liquids are transparent, making them hard to see in images.
Additionally, many approaches to finding rigid objects in a scene rely
on using a depth sensor, which is unsuitable for liquids as many of them are not
visible on depth sensors.  In this paper, we investigate ways to solve this
task using deep learning techniques.

% Introduce problem
Specifically, we examine the problems of {\it perceiving} and {\it reasoning}
about liquids.  That is, we ask the questions {\it Where in the raw visual data
  stream is liquid?} and {\it Can that be used to infer all places where liquid
  might be?}  To solve these problems, we take advantage of recent advances in
the field of deep learning.  This approach has been extremely successful in
various areas of computer vision, including classification
\citep{krizhevsky2012imagenet}, semantic labeling \citep{farabet2013learning},
and pose regression \citep{girshick2011efficient}, and it enabled computers to
successfully play Atari games from raw image data \citep{guo2014} and train
end-to-end policies on robots \citep{levine2015}.  The ability of deep networks
to process and make sense of raw visual data makes them a good fit for
perceiving and reasoning about liquids.

% How we do it
In this paper, we focus on the task of pouring as our exemplar task for learning about liquids.  
While researchers have already worked on robotic pouring tasks, previous techniques made simplifying assumptions, such as replacing water by an easily visible granular medium~\citep{yamaguchi2016c}, restricting the setting such that no perceptual feedback is necessary~\citep{langsfeld2014,okada2006,tamosiunaite2011,cakmak2012}, requiring highly accurate force sensors~\citep{rozo2013}, detecting moving liquid in front of a relatively static background~\citep{yamaguchi2016}, or dealing with simulated liquids only~\citep{kunze2014,kunze2015}.  
Here, we show how fully-convolutional deep networks (FCNs) can be trained to robustly perceive liquids and how they can be modified to peform better at generalization.  
To collect the large amounts of data necessary to train these deep networks, we utilize a realistic liquid simulator to generate a simulated dataset and a thermal camera to automatically label water pixels in a dataset collected on the real robot.

% Summarize results
Our results show that the methodology we propose in this paper is able to both
{\it perceive} and {\it reason} about liquids.  Specifically, they show that
recurrent networks are well-suited to these tasks, as they are able to integrate
information over time in a useful manner.  We also show that, with the right
type of input image, our neural networks can generalize to new data with objects
that are not included in the training set.  These results strongly suggest that
our deep learning approach is useful in a robotics context, which we demonstrate
in a closed-loop water pouring experiment.

The main contributions of this paper are:
\begin{itemize}
\item A deep learning framework for perceiving and reasoning about liquids based
  on raw visual data.
\item An approach for automatic, pixel-level labeling of real, visual data using
  a thermal camera along with heated liquid.
\item Two fully labeled datasets containing videos generated by a realistic liquid
  simulator and our novel thermal-visual imagery approach. 
\item An extensive experimental evaluation investigating the pros and cons of
  different deep network structures and demonstrating the superior performance
  of our approach, ultimately enabling a closed-loop water pouring system.
\end{itemize}

% Summarize paper
The rest of this paper is laid out as follows.  The next section discusses
relevant work related to ours.  The following section details the exact tasks we
investigate in this paper.  The sections after that describe how we generate our
simulated dataset and performed the pouring trials on our robot, followed by a
discussion of our learning methodology.  We then describe how we evaluate our
networks and present experimental results.  And finally, the last section
concludes the paper and summarizes the results.

\section{Related Work}

% Humans/infants and liquids psych stuff
Humans interact with liquids from a young age. 
Studies have shown that even infants can distinguish between rigid objects and ``substances'', or liquids \citep{hespos2012}. 
They have also shown that infants as young as five months have knowledge about how substances behave and interact with solid objects \citep{hespos2016}.
Furthermore, infants as young as 10 months have the ability to distinguish quantities of non-cohesive substances as greater or less than, although the quantity ratio must be larger for substances than for solid objects, suggesting that humans use a different mechanism to quantify substances than to quantify objects \citep{vanmarle2011}. 
Further studies have shown a correlation between humans' understanding of fluid dynamics and what would be expected of a probabilistic fluid physics model \citep{bates2015}, which suggests that humans have more than a simple perceptual understanding of fluid physics.

% Prior pouring stuff with robots
However, there has been little work in robotics on creating deep understandings
of liquids.  Much of the work in robotics involving liquids focuses on the task
of pouring \citep{langsfeld2014,okada2006,tamosiunaite2011,cakmak2012}.  All of
these works, though, constrain the task space enough so as to preclude the need
for any direct perception of or detailed reasoning about liquids.  For example,
in \citep{cakmak2012} the focus was solely on learning the overall pouring
trajectory, requiring the robot only to upend the source over the target
without needing to know the specifics of the liquid dynamics.  Work by Rozo {\it
  et al.} \citep{rozo2013} did have the robot pour a specific amount of liquid
into the target, which is more challenging than simply dumping all the contents
from the source.  However, they used their robot's precise proprioceptive sensors
to measure the liquid amount, bypassing the need for any sort of liquid
understanding.  Many robots do not have precise proprioceptive sensors, making
relying on them for liquid perception infeasible.

% Kunze and Beetz
There has been some prior work in robotics focusing on physics-based
understanding of liquids.  The works by Kunze and Beetz
\citep{kunze2014,kunze2015} utilized a robotic physics simulator to reason about
the outcomes of different actions taken by the robot.  Specifically, the robot
was tasked with finding the best sequence of actions to fry pancakes, which
involved reasoning about the liquid pancake batter while pouring and mixing it.
However, they simulated the liquid batter as a collection of small balls which
had different physics than real pancake batter, and they did not connect the
simulated environment to any real-world sensory data.  This resulted in a system
that, while it worked well in simulation, does not translate directly to a real
environment.

% Yamaguchi and Atkenson stuff
Similar work by Yamaguchi and Atkenson also utilized a robotic simulator with small balls in place of liquid. 
In \citep{yamaguchi2015,yamaguchi2016b}, they utilize this simulator to learn fluid dynamics models and perform planning over pouring trajectories.
Similar to prior work on robotic pouring, the robot poured all the contents from the source to the target, however in this case they utilized the simulator to reason about spilled liquid.
They also applied this framework to a robot in a real environment \citep{yamaguchi2016c}, although they used a constant color background and fixed color features to detect the liquid.
In more recent work \citep{yamaguchi2016}, they utilized optical flow in addition to a set of heuristics to perceive the liquids in a real-world environment. 
By detecting the motion of the liquid as it fell, they were able to localize the liquid in the scene, although the output labels were imprecise and could only detect liquid in the air and not liquid resting in a container.
For the tasks in this paper, it is necessary to detect both stationary and moving liquid, thus it is unlikely that this method would perform well on its own.
However, the results in \citep{yamaguchi2016} show that their method works in at least some cases, so we evaluate combining the use of optical flow for detection of moving liquid with deep neural networks to achieve maximal performance on the task.

% Fluid sims and random forest learning, Ali's stuff on connecting real data to sim state
While \citep{kunze2014,kunze2015,yamaguchi2015,yamaguchi2016b} all used small balls in place of liquid in their simulations, work in computer graphics has developed realistic fluid simulations \citep{bridson2015}. 
Work by Ladick\'{y} {\it et al.} \citep{ladicky2015} used regression forests to learn the particle interactions in the fluid simulation, resulting in realistic fluid physics, showing that machine learning methods are capable of learning fluid dynamics.
Additionally, the work in \citep{kunze2014,kunze2015,yamaguchi2015,yamaguchi2016b} was ungrounded, i.e., it was disconnected from real-world sensory data. 
On the other hand, the work by Mottaghi {\it et al.} \citep{mottaghi2016,mottaghi2016b} has shown how physical scene models can be connected to real visual data.
Specifically, they utilize convolutional neural networks (CNNs) to convert an image into a description of a scene, and then apply Newtonian physics to understand what will happen in future timesteps.
In this paper we also utilize CNNs to convert raw sensory data into a labeled scene.

% Shane's water stuff, Stuff by that guy finding puddles for autonomous driving
There has been some work in robotics on {\it perceiving} liquids.
Rankin {\it et al.} \citep{rankin2010,rankin2011} investigated ways to detect pools of water from an unmanned ground vehicle navigating rough terrain. 
However, t‎hey detected water based on simple color features or sky reflections, and didn't reason about the dynamics of the water, instead treating it as a static obstacle. 
Griffith {\it et al.} \citep{griffith2012} learned to categorize objects based on their interactions with running water, although the robot did not detect or reason about the water itself, rather it used the water as a means to learn about the objects.
In contrast to \citep{griffith2012}, we use vision to directly detect the liquid itself, and unlike \citep{rankin2010,rankin2011}, we treat the liquid as dynamic and reason about it.

% Our prior work
This paper builds on our prior work \citep{schenckc2016b,schenckc2016c}. 
In \citep{schenckc2016b} we utilized CNNs to both detect and track liquids in a realistic fluid simulator. 
We found that recurrent CNNs are best suited to perceive and reason about liquids.
In this paper we show how deep neural networks can be utilized on not only simulated data, but also on data collected on a real robot.
We show in work concurrent to this that a robot can use the liquid perception and reasoning capabilities developed in this paper to solve a real robotic task \citep{schenckc2016c}, specifically learning to pour a specific amount of liquid from only raw visual data.

\section{Task Overview}
\label{sec:task}

In this paper we investigate the duel tasks of {\it perception} and {\it reasoning} about liquids. 
We define {\it perception} to be determining what in the raw sensory data is liquid, and what is not liquid.
We call this task {\it detection}.
We define {\it reasoning} to be, given labels for the visible liquid (i.e., a working detector), determining where all the liquid is, even if it may not be directly perceivable (e.g., liquid inside a container).
We call this task {\it tracking}.
For this paper, we focus on the task of pouring as it requires reasoning about both where the visible liquid is as well as where hidden liquid is.

We evaluate our neural networks on the tasks of {\it detection} and {\it tracking} in both simulation and on data collected on a real robot.
For the simulated dataset, we generated a large amount of pouring sequences using a realistic liquid simulator.
As it is simple to get the ground truth state from the simulator, we can easily evaluate both tasks on the simulated data.
For evaluations using real-world data, we carried out a series of pouring trials on our robot.
We use a thermal camera in combination with heated water to acquire the ground truth pixel labels.
However, this only gives labels for {\it visible} liquid, and not liquid occluded by the containers, so we evaluate only the task of {\it detection} on the robot data.

\section{Simulated Data Set}
\label{sec:sim_data}

\newlength{\objectsize}
\setlength{\objectsize}{3.8cm}
\begin{figure*}[t]
    \centering
    \setlength{\fboxsep}{0pt}
    \setlength{\fboxrule}{1pt}
    \setlength{\unitlength}{1.0cm}
    \begin{subfigure}{\objectsize}
        \fbox{\includegraphics[width=\objectsize]{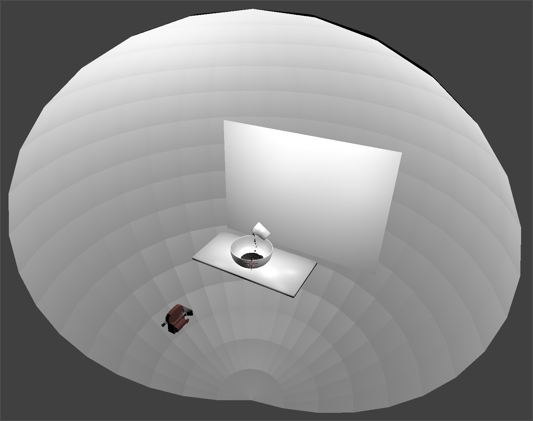}}
        \caption{{\it Untextured}}
        \label{fig:untextured}
    \end{subfigure}\hspace{0.2cm}%
    \begin{subfigure}{\objectsize}
        \fbox{\includegraphics[width=\objectsize]{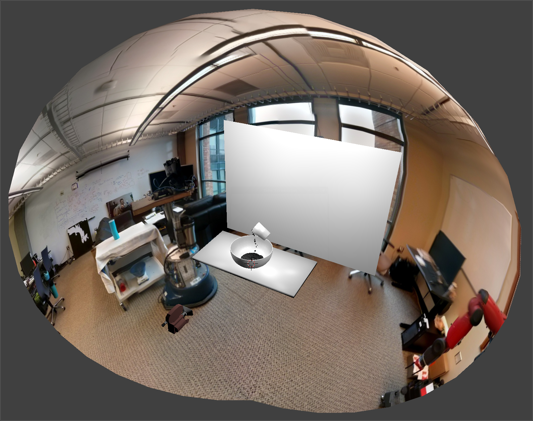}}
        \caption{{\it Background Texture}}
        \label{fig:background_texture}
    \end{subfigure}\hspace{0.2cm}%
    \begin{subfigure}{\objectsize}
        \fbox{\includegraphics[width=\objectsize]{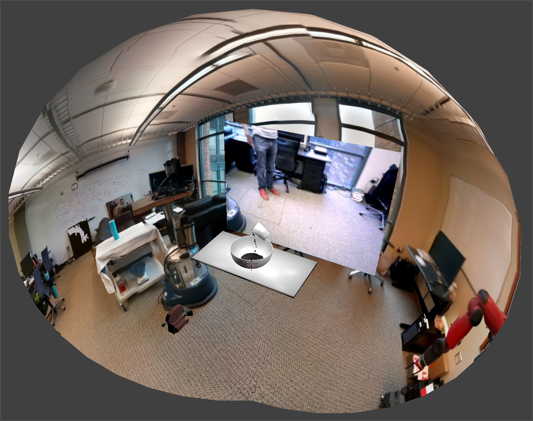}}
        \caption{{\it Background+Video}}
        \label{fig:background_video_plane}
    \end{subfigure}\hspace{0.2cm}%
    \begin{subfigure}{\objectsize}
        \fbox{\includegraphics[width=\objectsize]{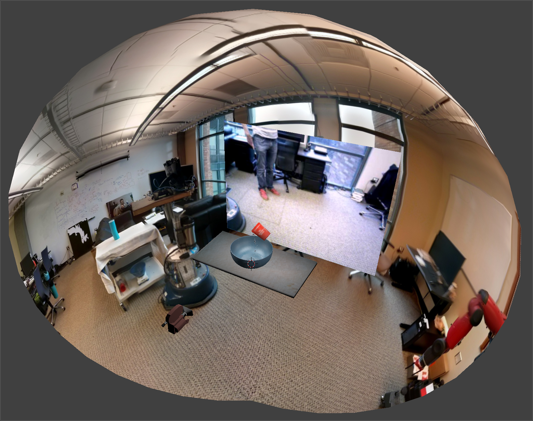}}
        \caption{{\it Fully Textured}}
        \label{fig:textured}
    \end{subfigure}
    \caption{The scene used to simulate pouring liquids. The background sphere is cut-away to show its interior. From left to right: The scene shown without any texture or materials; The background image sphere texture added; The video on the plane added in addition to the background texture; and The scene fully textured with all materials.}
    \label{fig:liquid_sim}
\end{figure*}

We use the simulated dataset generated in our prior work \citep{schenckc2016b} to evaluate our methodology.
The dataset contains 10,122 pouring sequences that are 15 seconds long each, for a total of 4,554,900 images.
Each sequence was generated using the 3D-modeling program Blender \citep{blender2016} and the library El'Beem for liquid simulation, which is based on the lattice-Boltzmann method for efficient, physically accurate liquid simulations \citep{korner2006}.

We divide the data generation into two steps: liquid simulation and rendering. 
Liquid simulation involves computing the trajectory of the mesh of the liquid over the course of the pour.
Rendering is converting the state of the simulation at each point in time into color images.
Liquid simulation is much more computationally intensive than rendering\footnote{Generating one 15 second sequence takes about 7.5 hours to simulate the liquid and an additional 0.5 hours to render it on our Intel Core i7 CPUs.}, so by splitting the data generation process into these two steps, we can simulate the trajectory of the liquid and then re-render it multiple times with different render settings (e.g., camera pose) to quickly generate a large amount of data.
We describe these two steps in the following sections.

\subsection{Liquid Simulation}

The simulation environment was set up as follows. 
A 3D model of the target container was placed on a flat plane parallel to the ground, i.e., the ``table.''
Above the target container and slightly to the side we placed the source container.
This setup is shown in Figure \ref{fig:untextured}.
The source container is pre-filled with a specific amount of liquid.
The source then rotates about the y-axis following a fixed trajectory such that the lip of the container turns down into the target container.
The trajectory of the liquid is computed at each timestep as the source container rotates.
Each simulation lasted exactly 15 seconds, or 450 frames at 30 frames per second.

For each simulation, we systematically vary 4 variables:
\begin{itemize}
    \item {\it Source Container} - {\it cup}, {\it bottle}, or {\it mug}
    \item {\it Target Container} - {\it bowl}, {\it dog dish}, or {\it fruit bowl}
    \item {\it Fill Amount} - 30\%, 60\%, or 90\%
    \item {\it Trajectory} - partial, hold, or dump
\end{itemize}
The 3 source containers we used are shown in Figures \ref{fig:sim_cup}, \ref{fig:sim_bottle}, and \ref{fig:sim_mug}, and the 3 target containers we used are shown in Figures \ref{fig:sim_bowl}, \ref{fig:sim_dish}, and \ref{fig:sim_fruit}.
Each source container was filled either 30\%, 60\%, or 90\% full at the start of each simulation.
The source was rotated along one of three trajectories: It was rotated until it was slightly past parallel with the table, held for 2 seconds, then rotated back to upright (partial); It was rotated until it was slightly past parallel with the table, where it stayed for the remainder of the simulation (hold); or It was rotated quickly until it was pointing nearly vertically down into the target container, remaining there until the simulation finished (dump).
The result was 81 liquid simulations (3 sources $\times$ 3 targets $\times$ 3 fill amounts $\times$ 3 trajectories).

\setlength{\objectsize}{2.5cm}
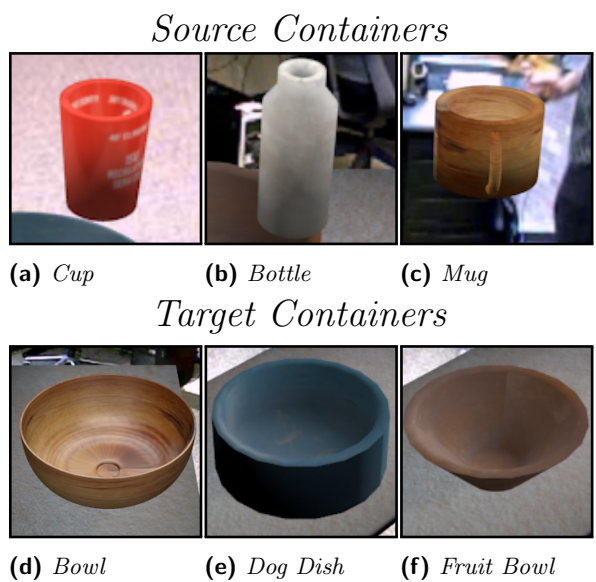
\begin{figure}
    \centering
    \setlength{\fboxsep}{0pt}
    \setlength{\fboxrule}{1pt}
    \setlength{\unitlength}{1.0cm}
    \begin{tikzpicture}
        \node[anchor=center,align=center] at (0.0,0.0) {{\it\Large Source Containers}};
    \end{tikzpicture}
    
    \begin{subfigure}{\objectsize}
        \fbox{\includegraphics[width=\objectsize]{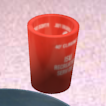}}
        \caption{{\it Cup}}
        \label{fig:sim_cup}
    \end{subfigure}\hspace{0.1cm}%
    \begin{subfigure}{\objectsize}
        \fbox{\includegraphics[width=\objectsize]{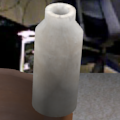}}
        \caption{{\it Bottle}}
        \label{fig:sim_bottle}
    \end{subfigure}\hspace{0.1cm}%
    \begin{subfigure}{\objectsize}
        \fbox{\includegraphics[width=\objectsize]{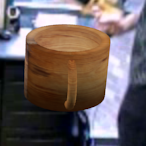}}
        \caption{{\it Mug}}
        \label{fig:sim_mug}
    \end{subfigure}
    
    \begin{tikzpicture}
        \node[anchor=center,align=center] at (0.0,0.0) {{\it\Large Target Containers}};
    \end{tikzpicture}
    
    \begin{subfigure}{\objectsize}
        \fbox{\includegraphics[width=\objectsize]{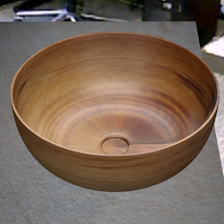}}
        \caption{{\it Bowl}}
        \label{fig:sim_bowl}
    \end{subfigure}\hspace{0.1cm}%
    \begin{subfigure}{\objectsize}
        \fbox{\includegraphics[width=\objectsize]{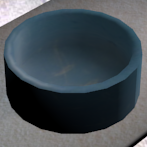}}
        \caption{{\it Dog Dish}}
        \label{fig:sim_dish}
    \end{subfigure}\hspace{0.1cm}%
    \begin{subfigure}{\objectsize}
        \fbox{\includegraphics[width=\objectsize]{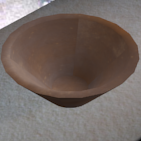}}
        \caption{{\it Fruit Bowl}}
        \label{fig:sim_fruit}
    \end{subfigure}
    
    \caption{The objects used to generate the simulated dataset. The first row are the three source containers. The last row are the 3 target containers. The objects are each shown here with 1 of their possible 7 textures.}
    \label{fig:sim_objects}
\end{figure}

\subsection{Rendering}

To generate rendered pouring sequences, we randomly select a simulation and render parameters\footnote{The number of parameters makes it infeasible to evaluate every possible combination.}.
We place the camera in the scene so that it is pointing directly at the table top where the target and source containers are.
In order to approximate realistic reflections on the liquid's surface, we enclose the scene in a sphere with a photo sphere taken in our lab set as the texture (shown in Figure \ref{fig:background_texture}).
Next we place a video of activity in our lab behind the table opposite the camera (shown in Figure \ref{fig:background_video_plane}).
We took videos such that they approximately match the location in the image on the background sphere behind the video plane.
We randomly select a texture for the source and target containers, and we render the liquid as 100\% transparent (but including reflections and refractions).
We also vary the reflectivity of the liquid as well as its index of refraction to simulate slight variations in the liquid type.
Figure \ref{fig:textured} shows the full scene with textures, video, and background sphere.

We randomly select from the following N parameters for each rendered sequence:
\begin{itemize}
    \item {\it Source Texture} - 7 preset textures
    \item {\it Target Texture} - 7 preset textures
    \item {\it Activity Video} - 8 videos
    \item {\it Liquid Reflectivity} - normal or none
    \item {\it Liquid Index-of-Refraction} - air-like, low-water, or normal-water
    \item {\it Camera Azimuth} - 8 azimuths
    \item {\it Camera Height} - high or low
    \item {\it Camera Distance} - close, medium, or far
\end{itemize}
There are 48 total camera viewpoints.
The camera azimuth is randomly selected from 1 of 8 possibilities spaced evenly around the table.
The height of the camera is selected such that it is either looking down into the target container at a 45 degree angle (high, lower-left image in Figure \ref{fig:sim_data_gen}) or it is level with the table looking directly at the side of the target (low, upper-left image in Figure \ref{fig:sim_data_gen}).
The camera is also placed either close to the table, far from the table, or in between.
The output of the rendering process is a series of color images, one for each frame of the sequence.

\subsection{Generating the Ground Truth}

\begin{figure}
    \centering
    \setlength{\fboxsep}{0pt}
    \setlength{\fboxrule}{1pt}
    \setlength{\unitlength}{1.0cm}
    \scalebox{0.65}{
    \begin{picture}(12.0,7.0)
        % I need to hide these here so that the stupid uploader doesn't forget to include these images.
        \put(0.0,0.0){\includegraphics[width=0.01cm]{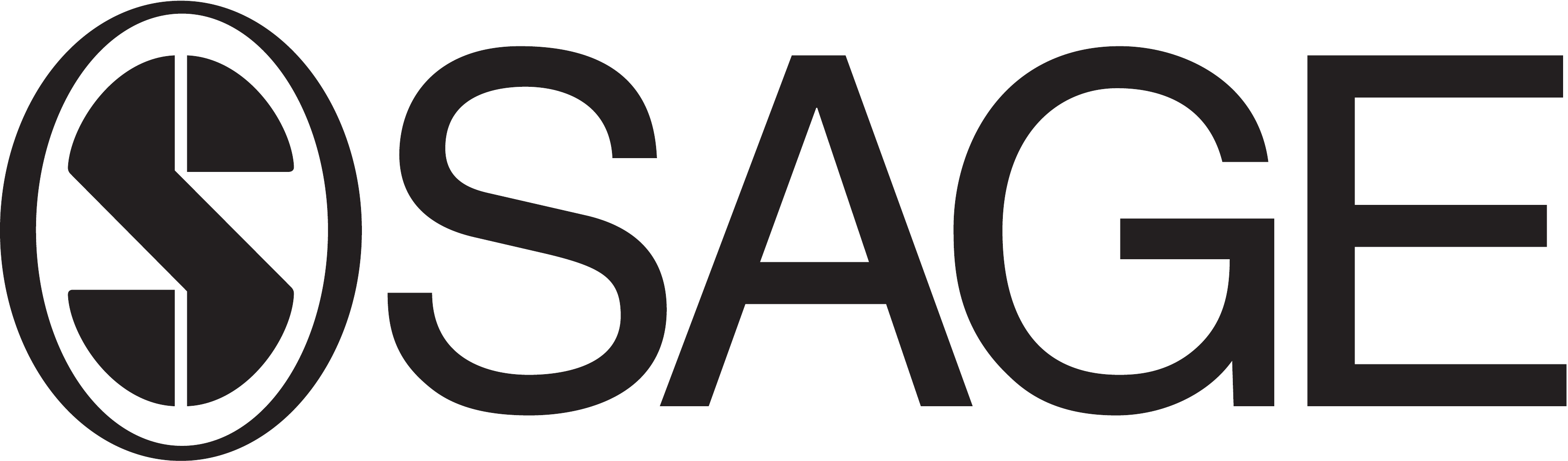}}
        \put(0.0,0.0){\includegraphics[width=0.01cm]{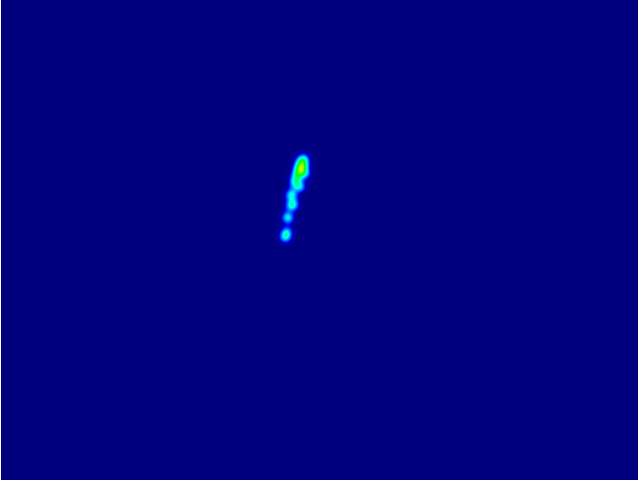}}
        
        \put(0.0,0.0){\fbox{\includegraphics[width=3.0cm]{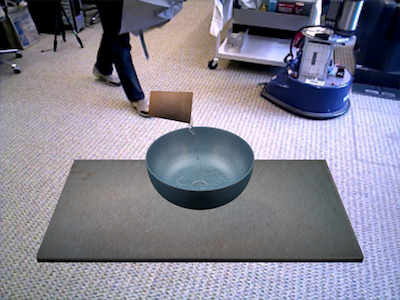}}}
        \put(3.0,0.0){\fbox{\includegraphics[width=3.0cm]{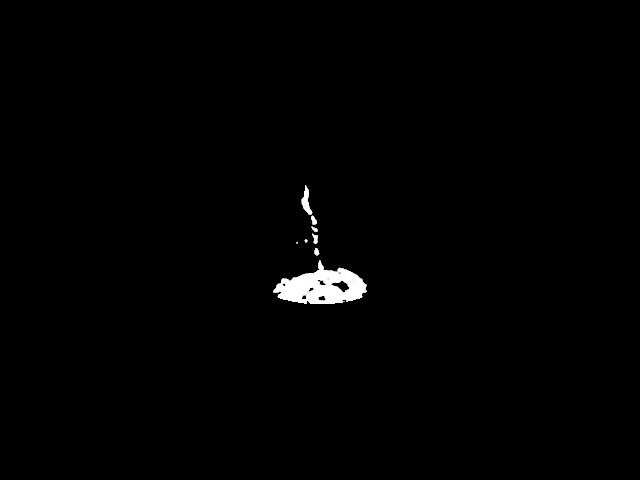}}}
        \put(6.0,0.0){\fbox{\includegraphics[width=3.0cm]{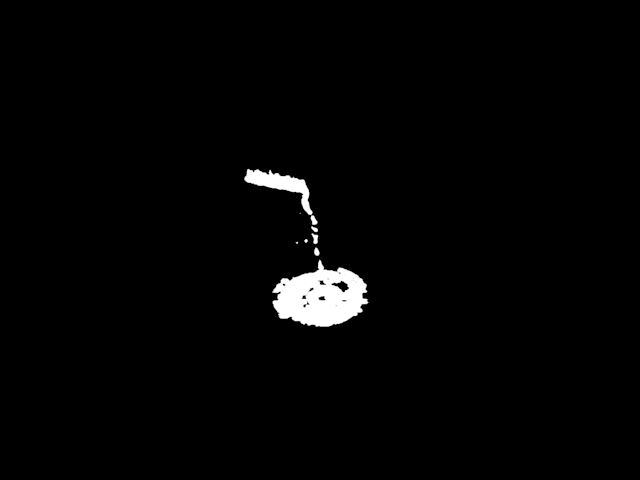}}}
        \put(9.0,0.0){\fbox{\includegraphics[width=3.0cm]{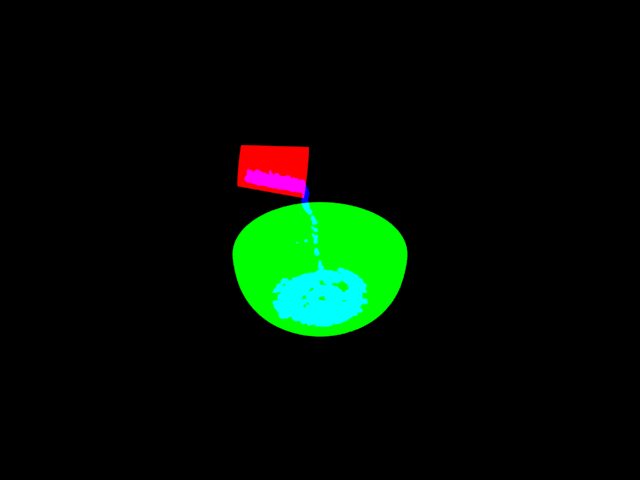}}}
        
        \put(0.0,2.25){\fbox{\includegraphics[width=3.0cm]{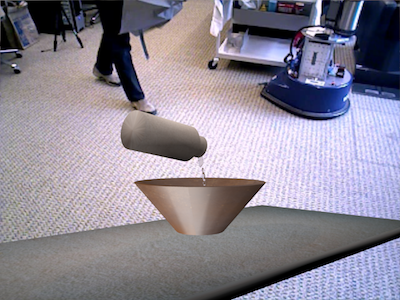}}}
        \put(3.0,2.25){\fbox{\includegraphics[width=3.0cm]{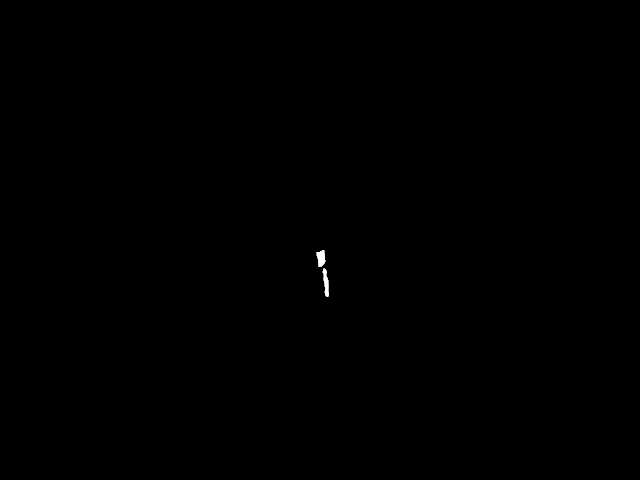}}}
        \put(6.0,2.25){\fbox{\includegraphics[width=3.0cm]{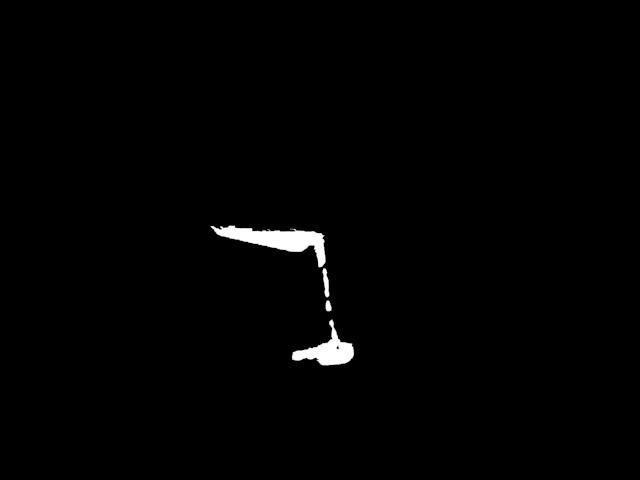}}}
        \put(9.0,2.25){\fbox{\includegraphics[width=3.0cm]{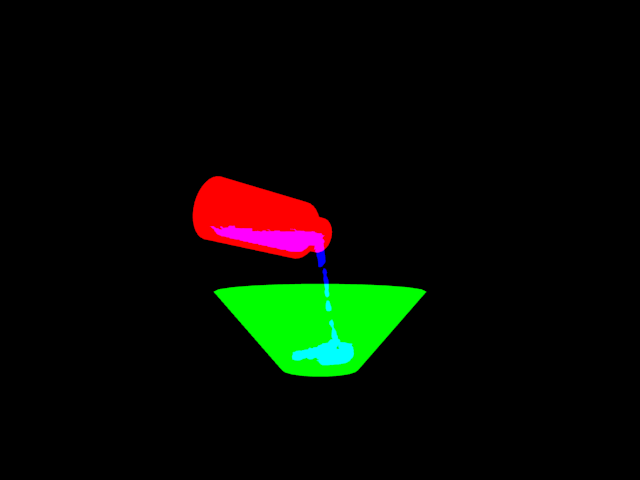}}}

        \put(0.0,4.5){\fbox{\includegraphics[width=3.0cm]{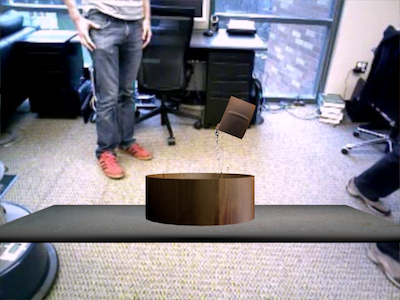}}}
        \put(3.0,4.5){\fbox{\includegraphics[width=3.0cm]{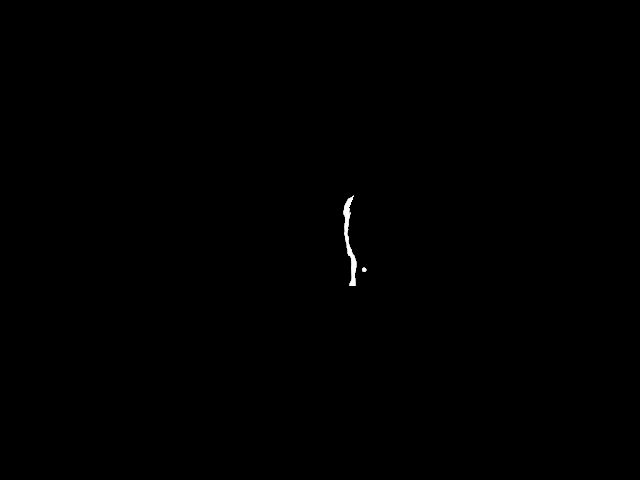}}}
        \put(6.0,4.5){\fbox{\includegraphics[width=3.0cm]{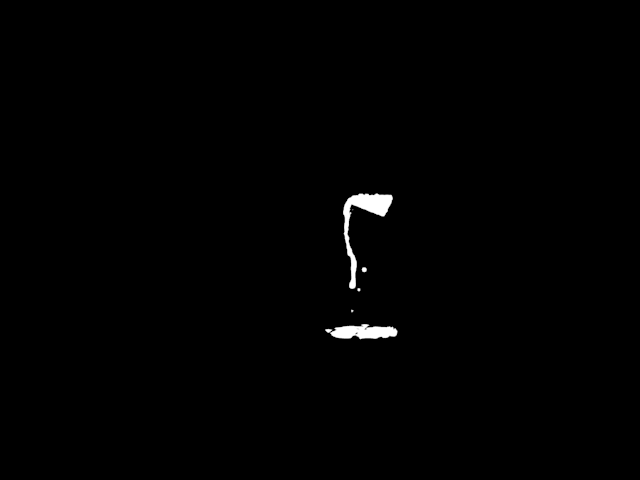}}}
        \put(9.0,4.5){\fbox{\includegraphics[width=3.0cm]{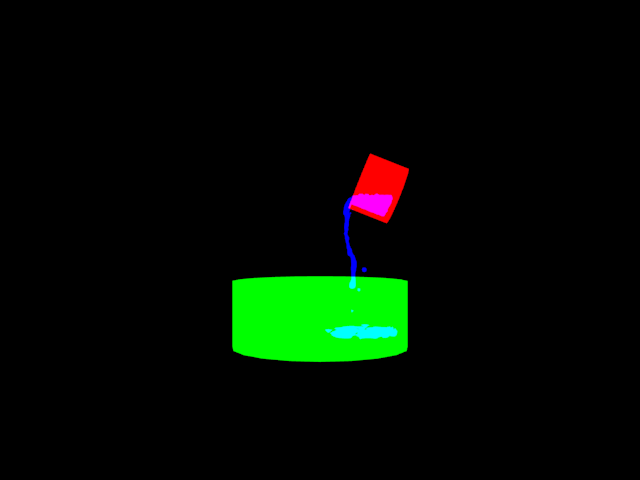}}}
        
        \put(0.0,7.0){\parbox{3.0cm}{{\begin{center}\bf\LARGE RGB\end{center}}}}
        \put(3.0,7.0){\parbox{3.0cm}{{\begin{center}\bf\LARGE Visible\end{center}}}}
        \put(6.0,7.0){\parbox{3.0cm}{{\begin{center}\bf\LARGE All\end{center}}}}
        \put(9.0,7.0){\parbox{3.0cm}{{\begin{center}\bf\LARGE Labels\end{center}}}}
    \end{picture}
    }
    \caption{Examples of frames from the simulated dataset. The left column is the raw RGB images generated by the renderer; the center-left column shows the ground truth liquid location for visible liquid; the center-right column shows the ground truth liquid location for all liquid in the scene; the right column shows the ground truth labeling output by the simulator.}
    \label{fig:sim_data_gen}
\end{figure}
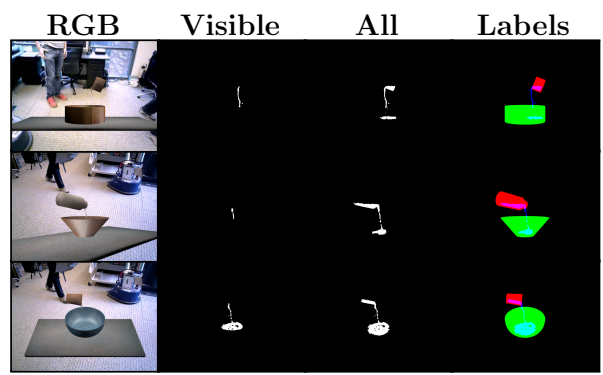

We generate the ground truth for each image in each rendered sequence as follows.
For each object (source container, target container, and liquid), we set that object to render as a solid color irrespective of lighting (red, green, and blue respectively).
Then we make all other objects in the scene invisible, and render the resulting scene.
We then combine the images for the objects as separate channels of a single image (right column of Figure \ref{fig:sim_data_gen}).

For the tasks of {\it detection} and {\it tracking}, we need to be able to distinguish between {\it visible} and {\it all} liquid respectively.
To do this, we render the scene again with each object rendered as its respective color, and then we encode which object is on top in the alpha channel of the ground truth image described in the last paragraph.
Some examples of the result are shown in Figure \ref{fig:sim_data_gen}.
The left column shows the rendered color image, the right column shows the ground truth pixel labels (absent the alpha channel), and the middle columns show the visible or all liquid.

\section{Robot Data Set}
\label{sec:robot_data}

\subsection{Robot}

\begin{figure}
    \centering
    \setlength{\fboxsep}{0pt}
    \setlength{\fboxrule}{1pt}
    \setlength{\unitlength}{1.0cm}
    \fbox{\includegraphics[width=5.0cm]{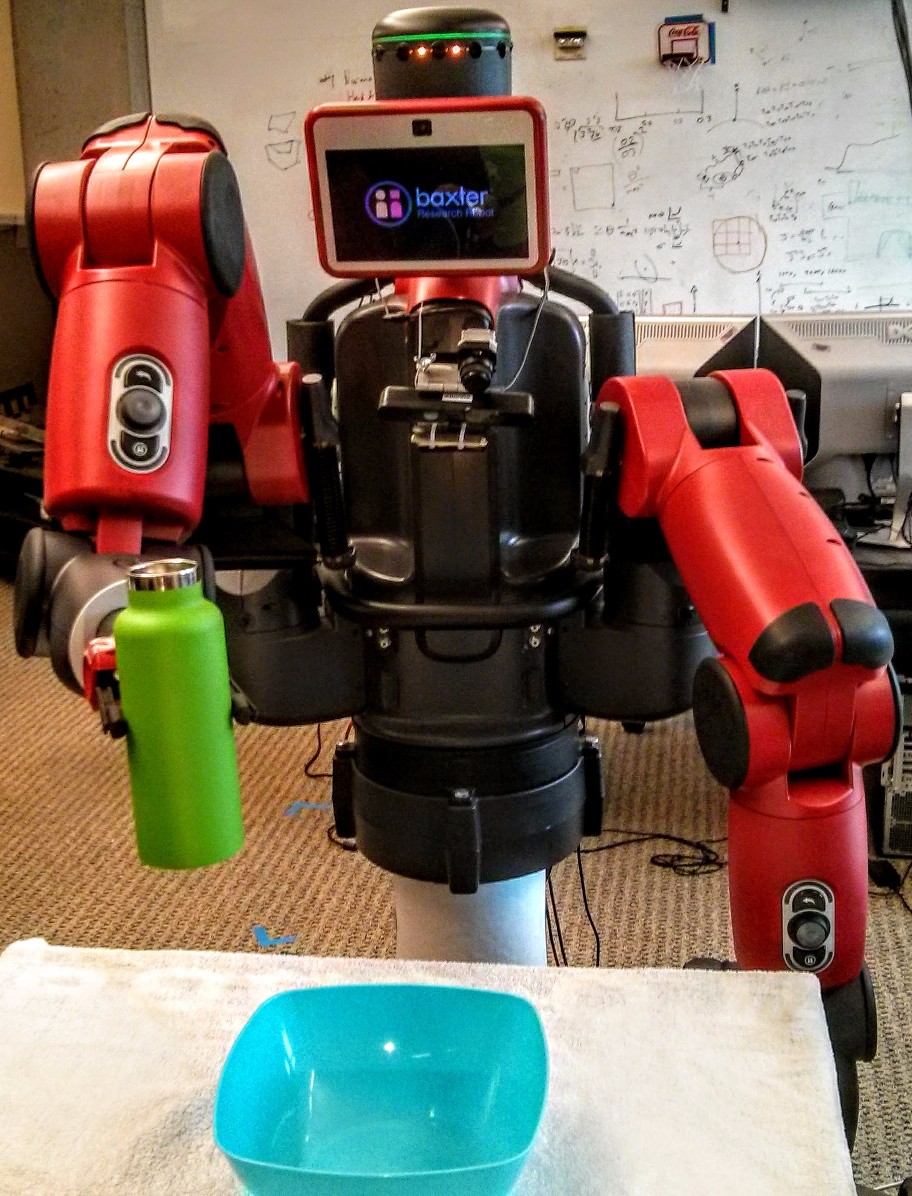}}
    \caption{The robot used in the experiments in this paper. It is shown here in front of a table, holding the {\it bottle} in its right gripper, with the {\it fruit bowl} placed on the table.}
    \label{fig:robot_setup}
\end{figure}

The robot used to collect the dataset is shown in Figure \ref{fig:robot_setup}.
It is a Rethink Robotics Baxter Research Robot, an upper-torso humanoid robot with 2 7-dof arms, each with a parallel gripper.
The robot is placed in front of a table with a towel laid over it to absorb spilled water.
The robot is controlled via joint velocity commands.
In the experiments in this paper, the robot uses only one of its arms at a time.
The arm is fixed above the target container and the robot controls the joint velocity of its last joint, i.e., the rotational angle of its wrist.

\subsection{Sensors}
The robot is equipped with a pair of cameras mounted to its front immediately below its screen.
The first camera is an Asus Xtion RGBD camera, capable of providing both color and depth images at $640{\times}480$ resolution and 30 Hz.
The second camera is an Infrared Cameras Inc. 8640P Thermographic camera, capable of providing thermal images at $640{\times}512$ resolution and 30 Hz.
The thermal camera is mounted immediately above the RGBD camera's color sensor, and is angled such that the two cameras view the same scene from largely similar perspectives.
The Baxter robot is also equipped with joint-torque sensors, however the signal from these sensors is too unreliable and so we did not use them in the experiments in this paper.

\subsubsection{Calibration of the Thermal Camera}

\setlength{\objectsize}{3.5cm}
\begin{figure*}[t]
    \centering
    \setlength{\fboxsep}{0pt}
    \setlength{\fboxrule}{1pt}
    \setlength{\unitlength}{1.0cm}
    \begin{subfigure}{\objectsize}
        \fbox{\includegraphics[width=\objectsize]{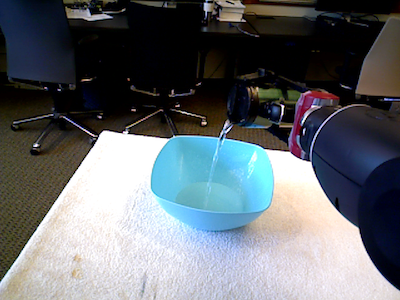}}
        \caption{{\it RGB}}
        \label{fig:calib_rgb}
    \end{subfigure}\hspace{0.2cm}%
    \begin{subfigure}{\objectsize}
        \fbox{\includegraphics[width=\objectsize]{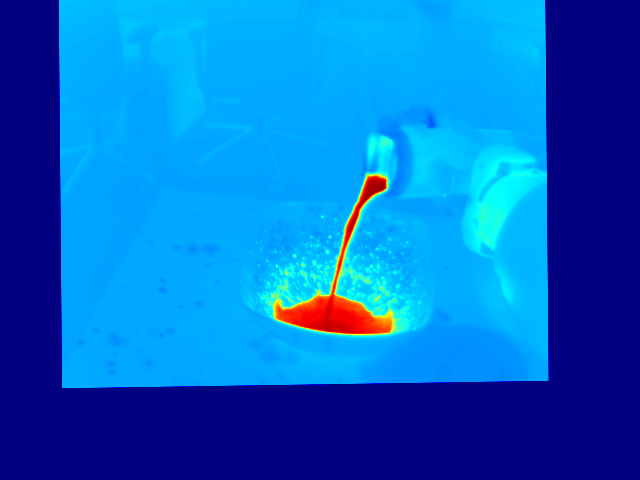}}
        \caption{{\it Thermal}}
        \label{fig:calib_therm}
    \end{subfigure}\hspace{0.2cm}%
    \begin{subfigure}{\objectsize}
        \fbox{\includegraphics[width=\objectsize]{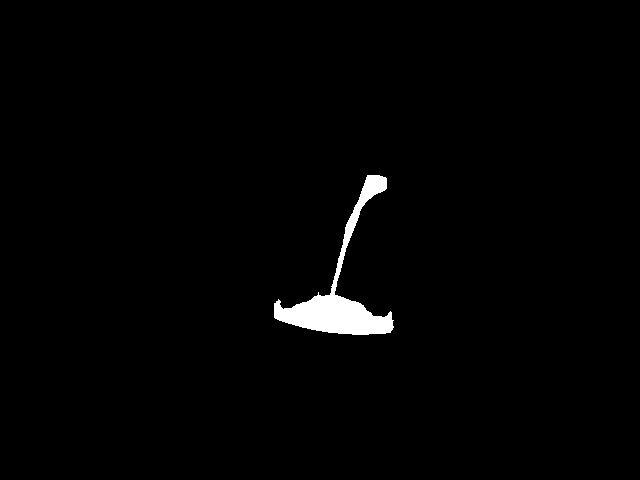}}
        \caption{{\it Threshold}}
        \label{fig:calib_gt}
    \end{subfigure}\hspace{0.2cm}%
    \begin{subfigure}{\objectsize}
        \fbox{\includegraphics[width=\objectsize]{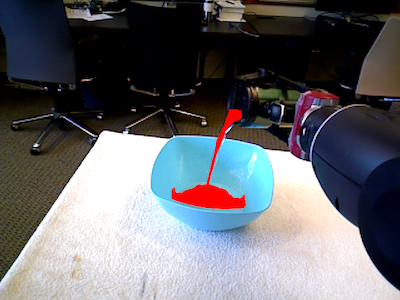}}
        \caption{{\it Overlay}}
        \label{fig:calib_overlay}
    \end{subfigure}
    \caption{An example of obtaining the ground truth liquid labels from the thermal camera. From left to right: The color image from the RGBD camera; The thermal image from the thermal camera transformed to the color pixel space; The result after thresholding the values in the thermal image; and An overlay of the liquid pixels onto the color image.}
    \label{fig:thermal_calib}
\end{figure*}

For our experiments, we use the thermal camera in combination with heated water to acquire the ground truth pixels labels for the liquid.
To do this, we must calibrate the thermal and RGBD cameras to each other.
In order to calibrate the cameras, we must know the correspondence between pixels in each image.
To get this correspondence, we use a checkerboard pattern printed on poster paper attached to an aluminum sheet.
We then mount a bright light to the robot's torso and shine that light on the checkerboard pattern while ensuring it is visible in both cameras.
The bright light is absorbed at differing rates by the light and dark squares of the pattern, resulting in a checkerboard pattern that is visible in the thermal camera\footnote{Albeit inverted as the black squares absorb more light than the white, thus appearing brighter in the thermal image. However we only care about the corners of the pattern, which are the same.}. 

We then use OpenCV's \texttt{findChessboardCorners} function to find the corners of the pattern in each image, resulting in a set of correspondence points $P^{therm}$ and $P^{RGB}$.
We compute the affine transform $T$ between the two sets using singular-value decomposition.
Thus to find the corresponding pixel from the thermal image to the RGB image, simply multiply as follows
\[ Tp^{therm} = p^{RGB} \]
where $p^{therm}$ is the xy coordinates of a pixel in the thermal image, and $p^{RGB}$ are its corresponding xy coordinates in the RGB image.

It should be noted that $T$ is only an affine transformation in pixel space, not a full registration between the two images.
That is, $T$ is only valid for pixels at the specific depth for which it was calibrated, and for pixels at different depths, $Tp^{therm}$ will not correspond to the same object in the RGB image as $p^{therm}$ in the thermal image.
While methods do exist to compute a full registration between RGB and thermal images \citep{pinggera2012}, they tend to be noisy and unreliable.
For our purposes, since the liquids are always a constant depth from the camera, we opted to use this affine transform instead, which is both faster and more reliable, resulting in better ground truth pixel labels.
While our RGBD camera does provide depth values at each pixel, the liquid does not appear in the depth readings and thus we could not use them to compute the full registration.
Figure \ref{fig:thermal_calib} shows an example of the correspondence between the thermal image and the RGB image.

\subsection{Objects}

\setlength{\objectsize}{2.5cm}
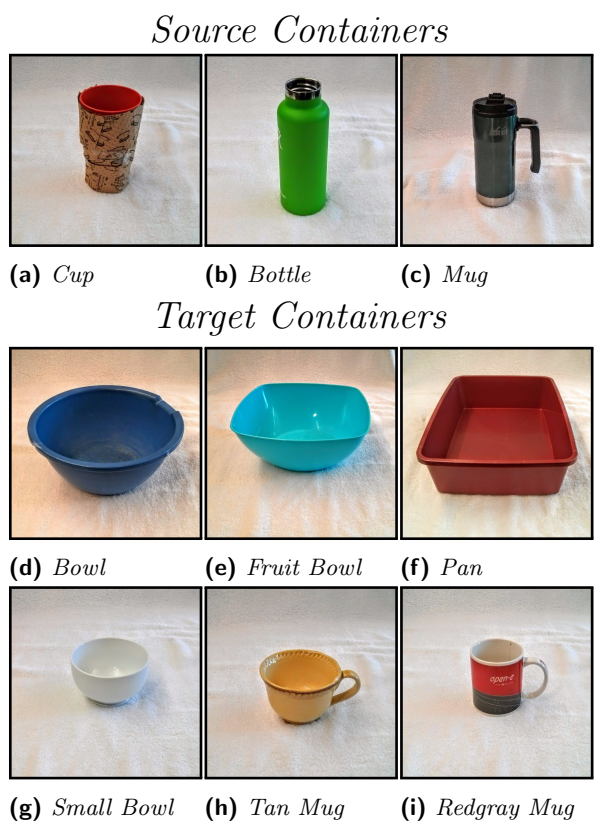
\begin{figure}
    \centering
    \setlength{\fboxsep}{0pt}
    \setlength{\fboxrule}{1pt}
    \setlength{\unitlength}{1.0cm}
    \begin{tikzpicture}
        \node[anchor=center,align=center] at (0.0,0.0) {{\it\Large Source Containers}};
    \end{tikzpicture}
    
    \begin{subfigure}{\objectsize}
        \fbox{\includegraphics[width=\objectsize]{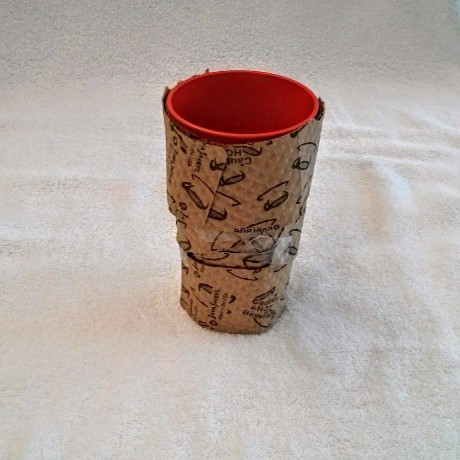}}
        \caption{{\it Cup}}
        \label{fig:cup}
    \end{subfigure}\hspace{0.1cm}%
    \begin{subfigure}{\objectsize}
        \fbox{\includegraphics[width=\objectsize]{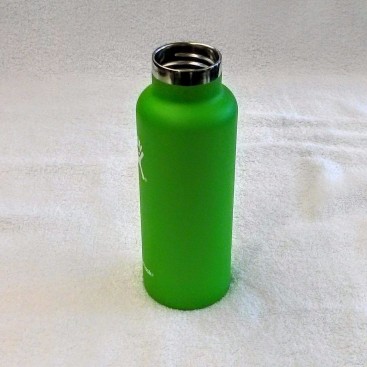}}
        \caption{{\it Bottle}}
        \label{fig:bottle}
    \end{subfigure}\hspace{0.1cm}%
    \begin{subfigure}{\objectsize}
        \fbox{\includegraphics[width=\objectsize]{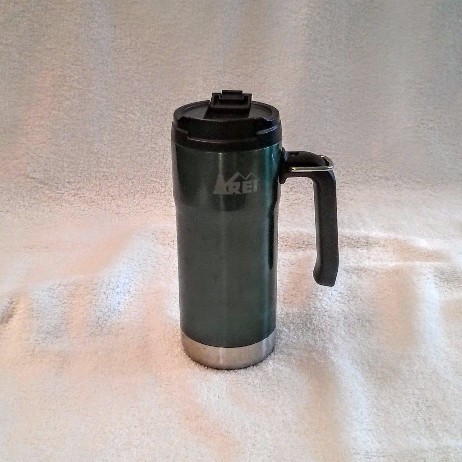}}
        \caption{{\it Mug}}
        \label{fig:mug}
    \end{subfigure}
    
    \begin{tikzpicture}
        \node[anchor=center,align=center] at (0.0,0.0) {{\it\Large Target Containers}};
    \end{tikzpicture}
    
    \begin{subfigure}{\objectsize}
        \fbox{\includegraphics[width=\objectsize]{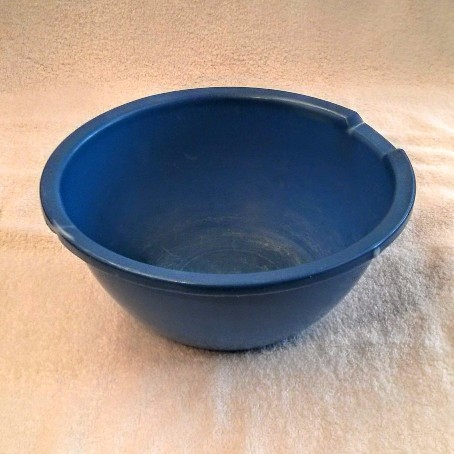}}
        \caption{{\it Bowl}}
        \label{fig:bowl}
    \end{subfigure}\hspace{0.1cm}%
    \begin{subfigure}{\objectsize}
        \fbox{\includegraphics[width=\objectsize]{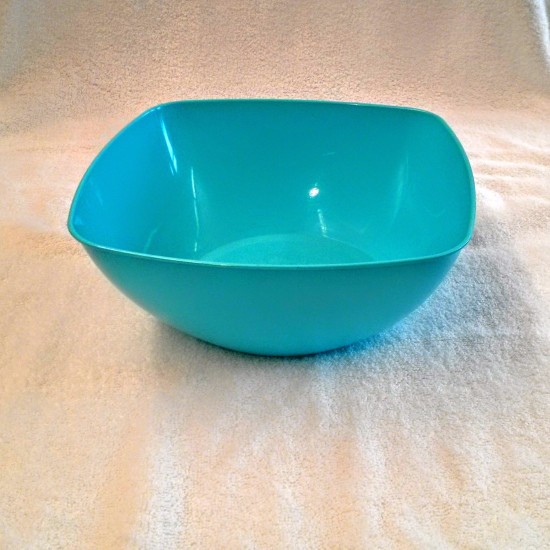}}
        \caption{{\it Fruit Bowl}}
        \label{fig:fruit_bowl}
    \end{subfigure}\hspace{0.1cm}%
    \begin{subfigure}{\objectsize}
        \fbox{\includegraphics[width=\objectsize]{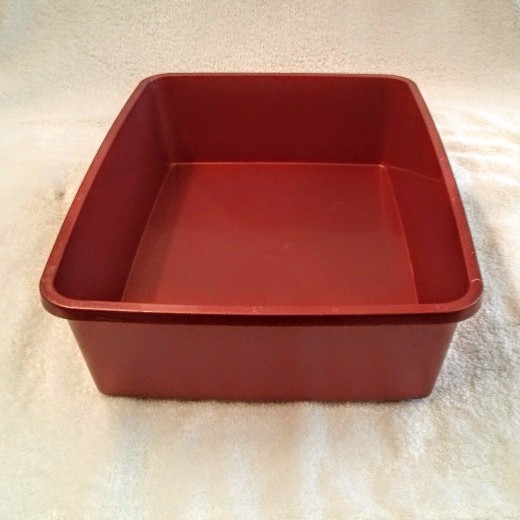}}
        \caption{{\it Pan}}
        \label{fig:pan}
    \end{subfigure}
    
    \begin{subfigure}{\objectsize}
        \fbox{\includegraphics[width=\objectsize]{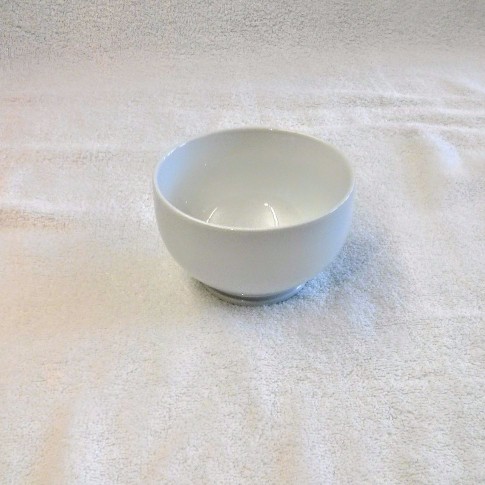}}
        \caption{{\it Small~Bowl}}
        \label{fig:small_bowl}
    \end{subfigure}\hspace{0.1cm}%
    \begin{subfigure}{\objectsize}
        \fbox{\includegraphics[width=\objectsize]{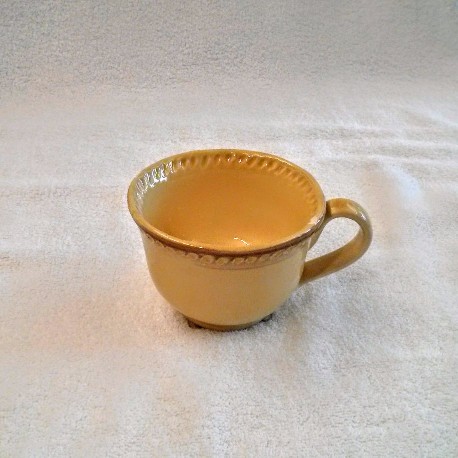}}
        \caption{{\it Tan Mug}}
        \label{fig:tan_mug}
    \end{subfigure}\hspace{0.1cm}%
    \begin{subfigure}{\objectsize}
        \fbox{\includegraphics[width=\objectsize]{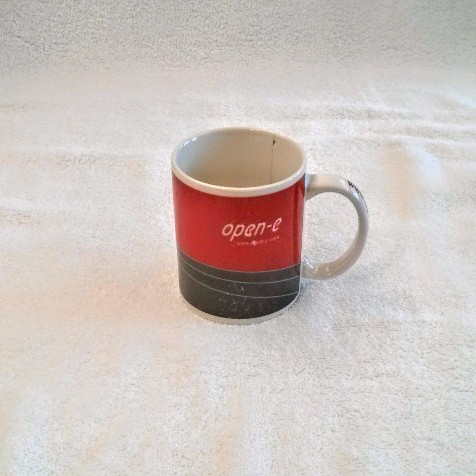}}
        \caption{{\it Redgray~Mug}}
        \label{fig:redgray_mug}
    \end{subfigure}
    
    \caption{The objects used to collect the dataset for this paper. The first row are the three source containers. The last two rows are the 6 target containers.}
    \label{fig:objects}
\end{figure}

For the robot dataset, we used two sets of objects: {\it source containers} and {\it target containers}.
We used 3 different source containers, the {\it cup}, the {\it bottle}, and the {\it mug}, shown in \ref{fig:cup}, \ref{fig:bottle}, and \ref{fig:mug}.
The {\it bottle} and {\it mug} were both thermally insulated, and we wrapped the {\it cup} in insulators.
This was done so that the robot could use the same source container from trial to trial without the object accumulating heat and appearing the same temperature as the liquid in the thermal image.
The only exception to this was the lid of the {\it mug}, which was not thermally insulated.
It was submersed in cold water between each trial to prevent heat build-up.

We used two different types of target containers, 3 large containers and 3 small.
The 3 large containers were the {\it bowl}, the {\it fruit bowl}, and the {\it pan} shown in Figures \ref{fig:bowl}, \ref{fig:fruit_bowl}, and \ref{fig:pan}.
The 3 small containers were the {\it small bowl}, the {\it tan mug} and the {\it redgray mug} shown in \ref{fig:small_bowl}, \ref{fig:tan_mug}, and \ref{fig:redgray_mug}.
Each target container was swapped out at the end of each trial to allow it time to dissipate the heat from the hot liquid.

\subsection{Data Collection}

We collected 1,009 pouring trials with our robot, generated by combining the data collected in our prior work \citep{schenckc2016c} with additional data collected for this paper.
For every trial on our robot we collected color, depth, and thermal images.
Additionally we collected 20 pouring trials for evaluating our methodology's generalization ability with objects not present in the training set.

We collected 648 pouring trails on our robot for use in this paper.
We fixed the robot's gripper over the target container and placed the source container in the gripper pre-filled with a specific amount of liquid.
The robot controlled the angle of the source by rotating its wrist joint.
We systematically varied 6 variables:
\begin{itemize}
 \item {\it Arm} - left or right
 \item {\it Source Container} - {\it cup}, {\it bottle}, or {\it mug}
 \item {\it Target Container} - {\it bowl}, {\it fruit bowl}, or {\it pan}
 \item {\it Fill Amount} - empty, 30\%, 60\%, or 90\%
 \item {\it Trajectory} - partial, hold, or dump
 \item {\it Motion} - minimal, moderate, or high
\end{itemize}
We used both arms, as well as varied the source containers.
We also used the 3 large target containers to contrast with the 3 small ones used in the prior dataset (described next).
In addition to various fill percents, we also included trials with no liquid to provide negative examples (which we use for both training and evaluating our networks).
The robot followed three fixed pouring trajectories: one in which it tilted the source to parallel with the ground and then returned to vertical; one in which it tilted the source to parallel with the ground and held it there; and one in which the robot quickly rotated the source to pointing almost vertically down into the target.
Finally, we added motion to the data.
For minimal motion, the only motion in the scene was the robot's with minimal background motion.
For moderate motion, a person moved around in the background of the scene while the robot was pouring.
For high motion, a person grasped and held the target container and actively moved it around while the robot poured into it.

\subsubsection{Prior Robot Data Collection}

In our prior work \citep{schenckc2016c} we collected 361 pouring trials.
We use that data as part of our dataset for this paper and briefly describe the data collection process here (refer to our prior work for more details).

The robot's gripper was fixed over the target container and it rotated only its wrist joint.
The source container was fixed in the robot's gripper and pre-filled with a specific amount of liquid.
The robot used its controller to attempt to pour a specific amount of liquid, resulting in trajectories where the robot would tilt the source container until some amount of liquid had transferred to the target and then the robot would tilt the source back upright.
We used only the {\it mug} as the source container for these trials and only the {\it small bowl}, {\it tan mug}, and {\it redgray mug} as target containers.
The robot used only its right arm.
We varied the target amount between 100 and 300 ml and the initial amount of liquid in the source between 300 and 400 ml.

\subsubsection{Test Data}

\setlength{\objectsize}{2.5cm}
\begin{figure}
    \centering
    \setlength{\fboxsep}{0pt}
    \setlength{\fboxrule}{1pt}
    \setlength{\unitlength}{1.0cm}
    \begin{subfigure}{\objectsize}
        \fbox{\includegraphics[width=\objectsize]{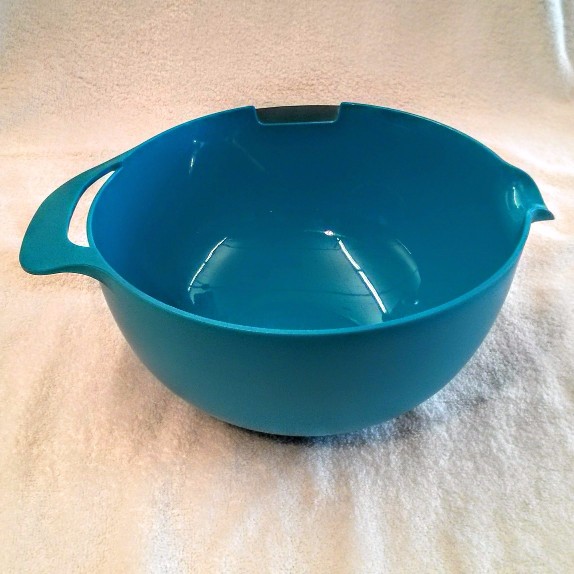}}
        \caption{{\it Blue Bowl}}
        \label{fig:blue_bowl}
    \end{subfigure}\hspace{0.1cm}%
    \begin{subfigure}{\objectsize}
        \fbox{\includegraphics[width=\objectsize]{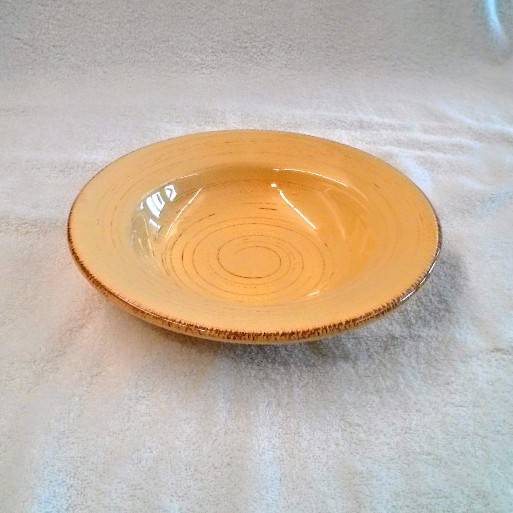}}
        \caption{{\it Tan Bowl}}
        \label{fig:tan_bowl}
    \end{subfigure}
    
    \begin{subfigure}{\objectsize}
        \fbox{\includegraphics[width=\objectsize]{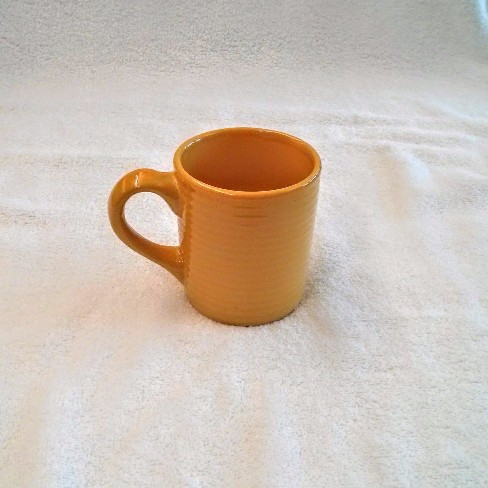}}
        \caption{{\it Gold Mug}}
        \label{fig:gold_mug}
    \end{subfigure}\hspace{0.1cm}%
    \begin{subfigure}{\objectsize}
        \fbox{\includegraphics[width=\objectsize]{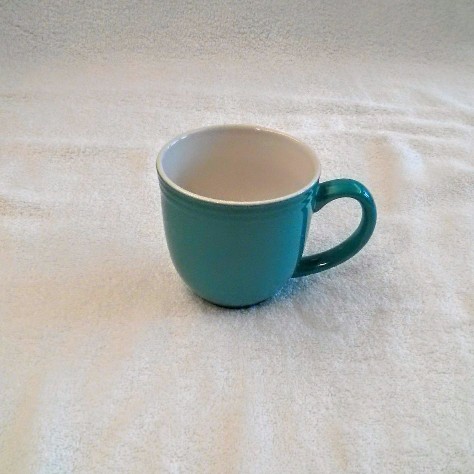}}
        \caption{{\it Teal Mug}}
        \label{fig:teal_mug}
    \end{subfigure}
    \caption{The target containers used to create the testing set.}
    \label{fig:test_objects}
\end{figure}

We also collected 20 pouring trials on our robot to evaluate our methodology's generalization ability.
We used the target containers in Figure~\ref{fig:test_objects} which were not included in the training datasets described in the previous sections.
For each object, we recorded 3 trials using the {\it mug} as the source container, the robot's right arm, and we filled the source initially 90\% full.
We collected one trial for each of the pouring trajectories described previously (partial, hold, and dump) with minimal background motion.
We collected 2 more trials with each object where we fixed the pouring trajectory (fixed as dump) and varied the motion between moderate and high.
Overall there were 5 trials per test object for a total of 20 test trials.

\subsection{Generating the Ground Truth from Thermal Images}

We process the thermal images into ground truth pixel labels as follows. 
First, we normalize the temperature values for each frame in the range 0 to 1. 
For all frames before liquid appears, we use the normalization parameters from the first frame with liquid.
We then threshold each frame at 0.6, that is, all pixels with values lower than 0.6 are labeled {\it not-liquid} and all pixels higher are labeled {\it liquid}.

While this results in a decent segmentation of the liquid, we can further improve it by removing erroneously labeled liquid pixels.
For example, during some sequences the robot briefly missed the target container, causing water to fall onto the table and be absorbed by the towel.
While this is still technically liquid, we do not wish to label it as such because after being absorbed by the towel, it's appearance qualitatively changes.
We use the PointCloud Library's plane fitting and point clustering functions to localize the object on the table from the depth image, and we remove points belonging to the table\footnote{We keep points above the lip of the target container in the image so as to not remove the stream of liquid as it transfers form the source to the target.}
Additionally, for some trials, the lid on the {\it mug} did not properly cool down between trials, and so for those trials we use a simple depth filter to remove pixels too close to the camera (the source container is slightly closer to the camera than the target).

\section{Learning Methodology}
\label{sec:methodology}

We utilize deep neural networks to learn the tasks of {\it detection} and {\it tracking}.
Specifically, we use fully-convolutional networks (FCNs) \citep{long2015}, that is, networks comprised of only convolutional layers (in addition to pooling and non-linear layers) and no fully-connected layers.
FCNs are well suited to the tasks in this paper because they produce pixel-wise labels and because they allow for variable sized inputs and outputs.
The following sections describe the different types of inputs and outputs for our networks, as well as the different network layouts.

\subsection{Network Input}
\label{sec:net_inputs}

\setlength{\objectsize}{2.5cm}
\begin{figure}
    \centering
    \setlength{\fboxsep}{0pt}
    \setlength{\fboxrule}{1pt}
    \setlength{\unitlength}{1.0cm}
    \begin{tikzpicture}
        \node[anchor=center,align=center] at (0.0,0.0) {{\it\Large Inputs}};
    \end{tikzpicture}
    
    \begin{subfigure}{\objectsize}
        \fbox{\includegraphics[width=\objectsize]{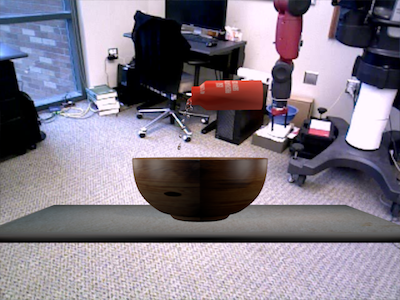}}
        \caption{{\it RGB}}
        \label{fig:rgb}
    \end{subfigure}\hspace{0.1cm}%
    \begin{subfigure}{\objectsize}
        \fbox{\includegraphics[width=\objectsize]{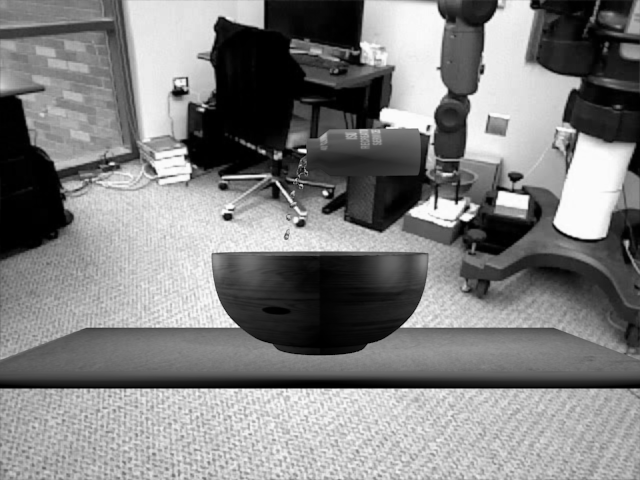}}
        \caption{{\it Grayscale}}
        \label{fig:gray}
    \end{subfigure}
       
    \begin{subfigure}{\objectsize}
        \fbox{\includegraphics[width=\objectsize]{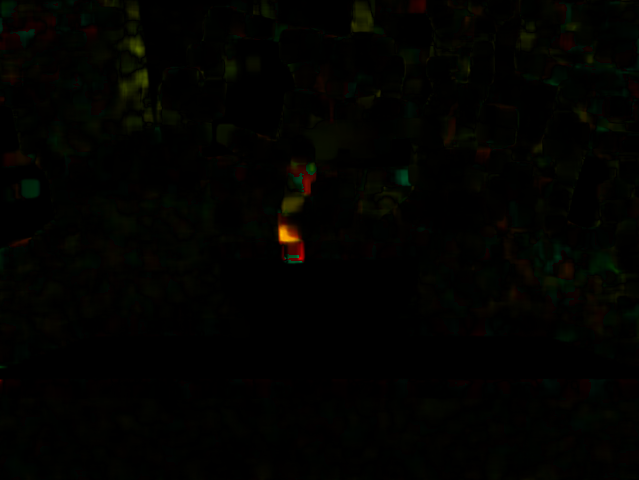}}
        \caption{{\it Optical Flow}}
        \label{fig:flow}
    \end{subfigure}\hspace{0.1cm}%  
    \begin{subfigure}{\objectsize}
        \fbox{\includegraphics[width=\objectsize]{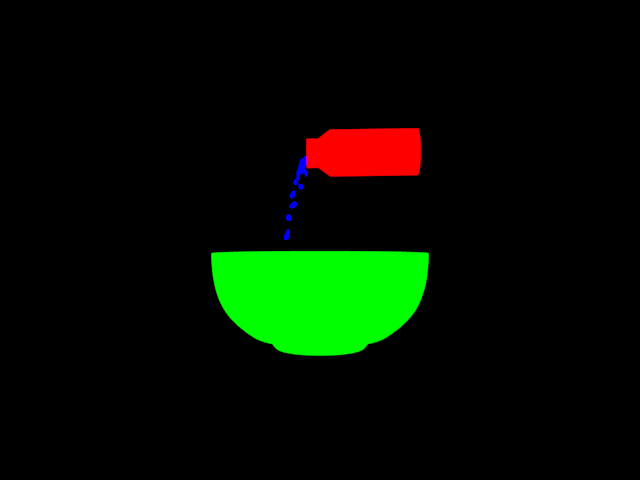}}
        \caption{{\it\scriptsize Visible Objects}}
        \label{fig:obj_vis}
    \end{subfigure}
      
    \begin{subfigure}{2\objectsize}
        \fbox{\includegraphics[width=\objectsize]{scene10_render_v3_N_NW_low_far_5_6_02_133_True_rgb300.png}\includegraphics[width=\objectsize]{scene10_render_v3_N_NW_low_far_5_6_02_133_True_flow300.png}}
        \caption{{\it RGB+Optical Flow}}
        \label{fig:rgbflow}
    \end{subfigure}
    
    \begin{subfigure}{2\objectsize}
        \fbox{\includegraphics[width=\objectsize]{scene10_render_v3_N_NW_low_far_5_6_02_133_True_gray300.png}\includegraphics[width=\objectsize]{scene10_render_v3_N_NW_low_far_5_6_02_133_True_flow300.png}}
        \caption{{\it Grayscale+Optical Flow}}
        \label{fig:grayflow}
    \end{subfigure}  
    
    \begin{tikzpicture}
        \node[anchor=center,align=center] at (0.0,0.0) {{\it\Large Outputs}};
    \end{tikzpicture}
    
    \begin{subfigure}{\objectsize}
        \fbox{\includegraphics[width=\objectsize]{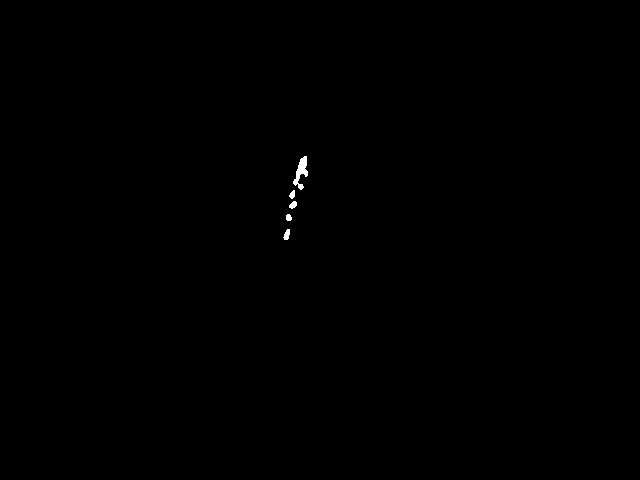}}
        \caption{{\it\small Visible Liquid}}
        \label{fig:visible}
    \end{subfigure}\hspace{0.1cm}%
    \begin{subfigure}{\objectsize}
        \fbox{\includegraphics[width=\objectsize]{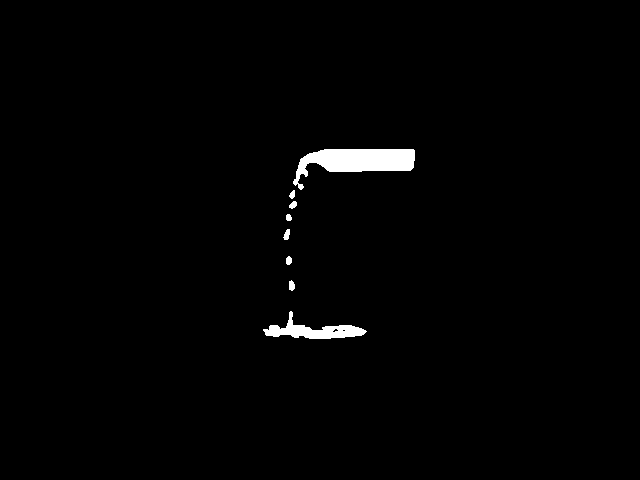}}
        \caption{{\it\footnotesize All Liquid}}
        \label{fig:all}
    \end{subfigure}
    
    \caption{Different images of the same frame from the same sequence. The upper part of this figure shows the different types of network inputs (RGB, grayscale, optical flow, visible objects, RGB+optical flow, and grayscale+optical flow). The lower part shows the types of desired network outputs (visible liquid and all liquid).}
    \label{fig:net_inout}
\end{figure}

We implemented 6 different types of input images to feed into our networks.
The first was the standard RGB image shown Figure \ref{fig:rgb}.
This is the type of FCN input most commonly seen in the literature \citep{havaei2015,romera2015}, and we use it as the primary type of input for all {\it detection} tasks on the simulated dataset.
However, since the robot dataset is one tenth the size of the simulated dataset, and thus is more prone to overfitting, we also desired to evaluate other types of images that may help counteract this tendency to overfit.
The most obvious type of image is grayscale, which was very commonly used in computer vision methods prior to CNNs \citep{forsyth2002}.
Figure \ref{fig:gray} shows a grayscale version of the RGB image in Figure \ref{fig:rgb}.

%            0.5, #pyr_scale
%            3, #levels
%            15,   #winsize
%            3,  #iterations
%            5,   #poly_n
%            1.2,   #poly_sigma
Inspired by prior work \citep{yamaguchi2016}, we also evaluated optical flow as an input to the networks.
We computed the dense optical flow for a given frame by calling OpenCV's \texttt{calcOpticalFlowFarneback} on that frame and the frame immediately prior (for the first frame we used the following frame instead).
For the parameters to the function \texttt{calcOpticalFlowFarneback}, we set the number of pyramid levels to 3 and the pyramid scale to 0.5, the window size to 15 and the number of iterations to 3, and the pixel neighborhood size to 5 with a standard deviation of 1.2.
Besides calling \texttt{calcOpticalFlowFarneback}, we did not perform any other filtering or smoothing on the optical flow output.

The output of the dense optical flow was an xy vector for each pixel, where the vector was the movement of that feature from the first frame to the second.
We converted each vector to polar coordinates (angle and magnitude), and further converted the angle to the sine and cosine values for the angle, resulting in three values for each pixel.
We store the resulting vectors in a three channel image, where the first channel is the sine of each pixel's angle, the second is the cosine, and the third is the magnitude.
An example is shown in Figure \ref{fig:flow} (converted to HSV for visualization purposes, where the angle is the hue and the magnitude is the value).
While \citep{yamaguchi2016} showed that optical flow at least correlates with moving liquid, it is not clear that flow by itself provides enough context to solve the {\it detection} problem.
Thus we also evaluate combining it with RGB (Figure \ref{fig:rgbflow}) and grayscale (Figure \ref{fig:grayflow}).

For the task of {\it tracking} we use pre-segmented images as input.
That is, we make the assumption that the robot already has a working detector that can identify the object label for each pixel in the image.
Each input pixel is labeled with the object that is visible at the pixel, which is represented as a one-hot vector (i.e., a binary vector where the index for the corresponding object label is 1 and all the other indices are 0).
An example of this is shown in Figure \ref{fig:obj_vis}.
When visualized, the labels for the source container become the red channel, for the target container become the green channel, and for the liquid become the blue channel.
Note that unlike the right column of Figure \ref{fig:sim_data_gen}, here the network only gets labels for the object that is ``on top'' at each pixel and cannot see objects occluded by other objects, e.g., cannot see the liquid in either container.

\subsection{Network Output}

The desired output of the network is fixed based on the task.
For {\it detection} the network should output the locations of the {\it visible} liquid in the scene.
An example of this is shown in Figure \ref{fig:visible}.
Note that in the case of Figure \ref{fig:visible}, most of the liquid is occluded by the containers, so here the robot is detecting primarily the flow of liquid as it transfers from the source to the target container.

For {\it tracking}, the desired output is the location of {\it all} liquid in the scene, including liquid occluded by the containers.
Here the network must learn to infer where liquid is in the scene based on other clues, such as determining the level of liquid in the source container based on the stream of liquid that is visible coming from the opening.
An example of this is shown in Figure \ref{fig:all}.
We should note that for our two datasets, it is only possible to get the ground truth location of all liquid from the simulated dataset because the simulator allows us to directly see the state of the environment, whereas on the robot dataset, the thermal camera only allows us to see the visible liquid and not liquid occluded by the containers.

The output of each network is a pixel-wise label confidence image, i.e., for each pixel, the network outputs its confidence in $[0,1]$ that that pixel is either {\it liquid} or {\it not-liquid}.

\subsection{Network Layouts}

All of the networks we use in this paper are fully-convolutional networks (FCNs).
That is, they do not have any fully-connected layers, which means each intermediate piece of data in the network maintains the image structure from the original input.
This makes FCNs well-suited for tasks which require pixel labels of the pixels from the input image, which both our tasks {\it detection} and {\it tracking} require.
Additionally, they allow variable sized input and outputs, which we take advantage of during training of our networks (described later in the evaluation section).

We use the Caffe deep learning framework \citep{jia2014} to implement our networks

\subsubsection{Input Blocks}

\def\setcubesizes[#1,#2,#3] {
    \pgfmathsetmacro{\cubex}{#1*1.7}\pgfmathsetmacro{\cubey}{#2*1.4}\pgfmathsetmacro{\cubez}{#3*1.85} %
}

\def\drawcubefront[#1,#2,#3,#4,#5] {
    \setcubesizes[0.0,#3,#4] %
    \draw[black,fill=#5] (#1,#2,0) -- ++(0,0,-\cubez) -- ++(0,\cubey,0) -- ++(0,0,\cubez) -- cycle; %
}

\def\drawlayer[#1,#2,#3,#4,#5,#6,#7] {
    \drawcubefront[#1,#2,#3,#4,#5] %
    \node[anchor=west] at ({#1-0.05pt},{#2+\cubey/2+0.05pt},0) {\rotatebox{90}{\centering\tiny \scalebox{0.8}{{\color{white}{\bf #6}\hspace{-1pt}{\it #7}}}}}; %
}

\def\drawcube[#1,#2,#3,#4,#5,#6] {
    \setcubesizes[#3,#4,#5] %
    \draw[black,fill=#6!90!white] ({#1+\cubex},#2,0) -- ++(0,0,-\cubez) -- ++(0,\cubey,0) -- ++(0,0,\cubez) -- cycle; %
    \draw[black,fill=#6!60!white] (#1,{#2+\cubey},0) -- ++(\cubex,0,0) -- ++(0,0,-\cubez) -- ++(-\cubex,0,0) -- cycle; %
    \draw[black,fill=#6!30!white] (#1,#2,0) -- ++(\cubex,0,0) -- ++(0,\cubey,0) -- ++(-\cubex,0,0) -- cycle; %
}

\def\drawblob[#1,#2,#3,#4,#5,#6] {
    \drawcube[#1,#2,#3,#4,#5,gray] %
    \node[anchor=south] at ({#1+\cubex/2},#2,0) {\rotatebox{90}{{\tiny #6}}}; %
}

\def\drawblock[#1,#2,#3,#4,#5,#6] {
    \drawcube[#1,#2,#3,#4,#5,blue] %
    \node[anchor=center,align=center] at ({#1+\cubex/2},{#2+\cubey/2},0) {{\bf #6}}; %
}

\def\drawinputblob[#1,#2,#3,#4,#5,#6] {
    \drawcube[#1,#2,#3,#4,#5,gray] %
    \path ({#1+\cubex}, {#2+\cubey/2}, 0) -- node[sloped,xslant=1.0] {\includegraphics[width=#5cm,height=#4cm]{#6}} ({#1+\cubex}, {#2+\cubey/2}, {0-\cubez}); %
}

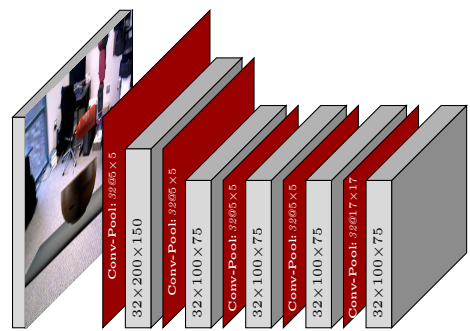
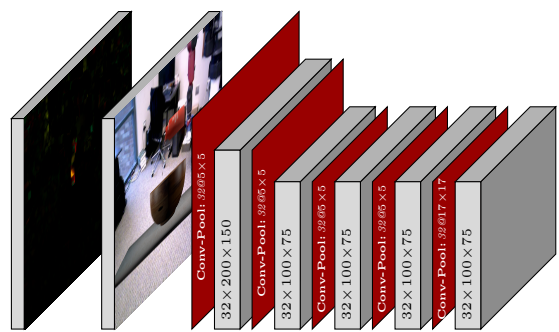
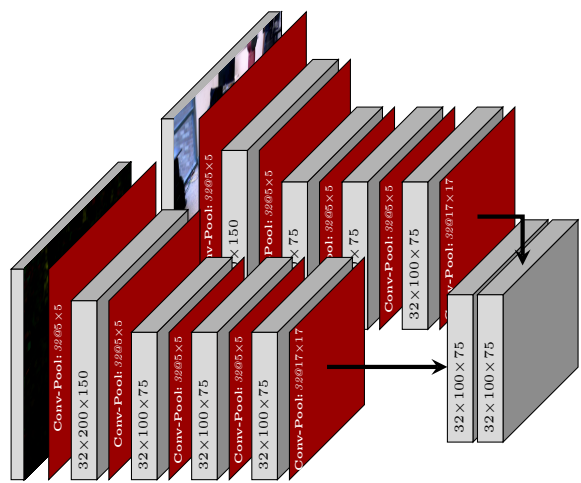
\begin{figure}

\begin{subfigure}{\columnwidth}
\begin{center}
\begin{tikzpicture}
    \drawinputblob[0.0,0.0,0.1,2.0,2.0,scene10_render_v3_N_NW_low_far_5_6_02_133_True_rgb300.png]
    \drawlayer[1.2,0.0,2.0,2.0,red!60!black, Conv-Pool:, 32@$5{\times}5$]
    \drawblob[1.5,0.0,0.2,1.7,1.7,$32{\times}200{\times}150$]
    \drawlayer[2.0,0.0,1.7,1.7,red!60!black, Conv-Pool:, 32@$5{\times}5$]
    \drawblob[2.3,0.0,0.2,1.4,1.4,$32{\times}100{\times}75$]
    \drawlayer[2.8,0.0,1.4,1.4,red!60!black, Conv-Pool:, 32@$5{\times}5$]
    \drawblob[3.1,0.0,0.2,1.4,1.4,$32{\times}100{\times}75$]
    \drawlayer[3.6,0.0,1.4,1.4,red!60!black, Conv-Pool:, 32@$5{\times}5$]
    \drawblob[3.9,0.0,0.2,1.4,1.4,$32{\times}100{\times}75$]
    \drawlayer[4.4,0.0,1.4,1.4,red!60!black, Conv-Pool:, 32@$17{\times}17$]
    \drawblob[4.7,0.0,0.2,1.4,1.4,$32{\times}100{\times}75$]
\end{tikzpicture}
\end{center}
\vspace{-0.5cm}
\caption{Standard Input Block}
\label{fig:standard_input}
\end{subfigure}

\begin{subfigure}{\columnwidth}
\begin{center}
\begin{tikzpicture}
    \drawinputblob[-1.2,0.0,0.1,2.0,2.0,scene10_render_v3_N_NW_low_far_5_6_02_133_True_flow300.png]
    \drawinputblob[0.0,0.0,0.1,2.0,2.0,scene10_render_v3_N_NW_low_far_5_6_02_133_True_rgb300.png]
    \drawlayer[1.2,0.0,2.0,2.0,red!60!black, Conv-Pool:, 32@$5{\times}5$]
    \drawblob[1.5,0.0,0.2,1.7,1.7,$32{\times}200{\times}150$]
    \drawlayer[2.0,0.0,1.7,1.7,red!60!black, Conv-Pool:, 32@$5{\times}5$]
    \drawblob[2.3,0.0,0.2,1.4,1.4,$32{\times}100{\times}75$]
    \drawlayer[2.8,0.0,1.4,1.4,red!60!black, Conv-Pool:, 32@$5{\times}5$]
    \drawblob[3.1,0.0,0.2,1.4,1.4,$32{\times}100{\times}75$]
    \drawlayer[3.6,0.0,1.4,1.4,red!60!black, Conv-Pool:, 32@$5{\times}5$]
    \drawblob[3.9,0.0,0.2,1.4,1.4,$32{\times}100{\times}75$]
    \drawlayer[4.4,0.0,1.4,1.4,red!60!black, Conv-Pool:, 32@$17{\times}17$]
    \drawblob[4.7,0.0,0.2,1.4,1.4,$32{\times}100{\times}75$]
\end{tikzpicture}
\end{center}
\vspace{-0.5cm}
\caption{Early Fusion Input Block}
\label{fig:early_fusion}
\end{subfigure}

\begin{subfigure}{\columnwidth}
\begin{center}
\begin{tikzpicture}[>=stealth]
    % Upper net
    \drawinputblob[0.7,0.0,0.1,2.0,2.0,scene10_render_v3_N_NW_low_far_5_6_02_133_True_rgb300.png]
    \drawlayer[1.2,0.0,2.0,2.0,red!60!black, Conv-Pool:, 32@$5{\times}5$]
    \drawblob[1.5,0.0,0.2,1.7,1.7,$32{\times}200{\times}150$]
    \drawlayer[2.0,0.0,1.7,1.7,red!60!black, Conv-Pool:, 32@$5{\times}5$]
    \drawblob[2.3,0.0,0.2,1.4,1.4,$32{\times}100{\times}75$]
    \drawlayer[2.8,0.0,1.4,1.4,red!60!black, Conv-Pool:, 32@$5{\times}5$]
    \drawblob[3.1,0.0,0.2,1.4,1.4,$32{\times}100{\times}75$]
    \drawlayer[3.6,0.0,1.4,1.4,red!60!black, Conv-Pool:, 32@$5{\times}5$]
    \drawblob[3.9,0.0,0.2,1.4,1.4,$32{\times}100{\times}75$]
    \drawlayer[4.4,0.0,1.4,1.4,red!60!black, Conv-Pool:, 32@$17{\times}17$]
    
    \pgfmathsetmacro{\lowery}{-2.0}
    \pgfmathsetmacro{\lowxoff}{2.0}
    \draw[black,ultra thick,dotted] (0.85,0.0) -- ++(-\lowxoff,\lowery);
    
    % Lower net
    \drawinputblob[0.7-\lowxoff,\lowery,0.1,2.0,2.0,scene10_render_v3_N_NW_low_far_5_6_02_133_True_flow300.png]
    \drawlayer[1.2-\lowxoff,\lowery,2.0,2.0,red!60!black, Conv-Pool:, 32@$5{\times}5$]
    \drawblob[1.5-\lowxoff,\lowery,0.2,1.7,1.7,$32{\times}200{\times}150$]
    \drawlayer[2.0-\lowxoff,\lowery,1.7,1.7,red!60!black, Conv-Pool:, 32@$5{\times}5$]
    \drawblob[2.3-\lowxoff,\lowery,0.2,1.4,1.4,$32{\times}100{\times}75$]
    \drawlayer[2.8-\lowxoff,\lowery,1.4,1.4,red!60!black, Conv-Pool:, 32@$5{\times}5$]
    \drawblob[3.1-\lowxoff,\lowery,0.2,1.4,1.4,$32{\times}100{\times}75$]
    \drawlayer[3.6-\lowxoff,\lowery,1.4,1.4,red!60!black, Conv-Pool:, 32@$5{\times}5$]
    \drawblob[3.9-\lowxoff,\lowery,0.2,1.4,1.4,$32{\times}100{\times}75$]
    \drawlayer[4.4-\lowxoff,\lowery,1.4,1.4,red!60!black, Conv-Pool:, 32@$17{\times}17$]
    
    % Combined net
    \pgfmathsetmacro{\lowery}{\lowery-1.0}
    %\draw[black,ultra thick,dotted] (4.4,\lowery/2+0.05) -- ++(1.3,0.0);
    \drawblob[4.5,\lowery/2,0.2,1.4,1.4,$32{\times}100{\times}75$]
    \drawblob[4.9,\lowery/2,0.2,1.4,1.4,$32{\times}100{\times}75$]
    \draw[black,ultra thick,->] (4.9,1.5) -- ++(0.6,0.0) -- ++(0.0,\lowery+2.35);
    \draw[black,ultra thick,->] (4.9-\lowxoff,\lowery+1.0+1.5) -- ++(1.6,0.0);
\end{tikzpicture}
\end{center}
\vspace{-0.5cm}
\caption{Late Fusion Input Block}
\label{fig:late_fusion}
\end{subfigure}

\caption{The 3 different types of input blocks. The first is used when the network takes only a single type of input; the second two are used when combining two different types of input. Here gray boxes are the feature representations at each level of the network, and the colored squares are the layers that operate on each representation. Gray boxes immediately adjacent indicate channel-wise concatenation.}
\label{fig:input_blocks}

\end{figure}

Each network we implement is built from one or more input blocks.
Input blocks are combinations of network layers with different types of input.
Essentially each is the beginning part of an FCN.
We split our description of our neural networks into input blocks and network types (below) to simplify it.
We combine our different types of input blocks with our different network types to create a combinatorially larger number of networks, which we then use to solve the tasks of {\it detection} and {\it tracking}.

Figure \ref{fig:input_blocks} shows the 3 different types of input blocks that we use in this paper.
The first (Figure \ref{fig:standard_input}) is the standard input block used by most of our networks.
It takes as input a single image, which it then passes through 5 conv-pool layers, which apply a convolution, then a rectified linear filter, and finally a max pooling operation.
The first two conv-pool layers have a stride of two for the max pooling operation; all other layers have a stride of one.
The output of this input block is a tensor with shape $32\times\frac{H}{4}\times\frac{W}{4}$ where $H$ and $W$ are the height and width of the input image respectively.

The second two input blocks are used for networks that take two different types of images as input (e.g., RGB and optical flow).
Figure \ref{fig:early_fusion} shows the early-fusion approach, which combines the two images channel-wise and feeds them into a block otherwise identical to the standard input block.
Figure \ref{fig:late_fusion} shows the late-fusion approach, which feeds each image into separate copies of the standard input block, and then concatenates the resulting tensors channel-wise, resulting in a $64\times\frac{H}{4}\times\frac{W}{4}$ tensor.
Some work in the literature has suggested that the late-fusion approach tends to perform better than the early-fusion approach\citep{valada2016}, however in this paper we evaluate this premise on our own tasks.

\subsubsection{Network Types}

\begin{figure*}
\begin{subfigure}{\textwidth}
\begin{center}
\begin{tikzpicture}
    \drawblock[1.5,0.0,2.0,1.4,1.4,Input Block]
    \drawlayer[5.2,0.0,1.4,1.4,magenta!70!black, Conv:, 64@$1{\times}1$]
    \drawblob[5.5,0.0,0.4,1.4,1.4,$64{\times}100{\times}75$]
    \drawlayer[6.3,0.0,1.4,1.4,magenta!70!black, Conv:, 64@$1{\times}1$]
    \drawblob[6.6,0.0,0.4,1.4,1.4,$64{\times}100{\times}75$]
    \drawlayer[7.4,0.0,1.4,1.4,orange!80!black, $Conv^\top$:, 2@$16{\times}16$]
    \drawinputblob[7.7,0.0,0.1,2.0,2.0,scene10_render_v3_N_NW_low_far_5_6_02_133_True_visible_pred300.png]
\end{tikzpicture}
\end{center}
\vspace{-1.0cm}
\caption{FCN}
\label{fig:fcn}
\end{subfigure}

\begin{subfigure}{\textwidth}
\begin{center}
\begin{tikzpicture}[>=stealth]
    \drawblock[-2.0,-2.0,2.0,1.4,1.4,\parbox{4cm}{\begin{center}Input Block\\$t=k$\end{center}}]
    \draw[black,line width=2.0mm,dotted] (0.0,2.0) -- ++(0.0,-1.5);
    \drawblock[-2.0,2.0,2.0,1.4,1.4,\parbox{4cm}{\begin{center}Input Block\\$t=1$\end{center}}]
    \drawblob[3.0,0.0,0.2,1.4,1.4,$32N_I{\times}100{\times}75$]
    \draw[black,line width=1.0mm,dotted] (3.8,1.4) -- ++(2.6,0.0);
    \drawblob[4.6,0.0,0.2,1.4,1.4,$32N_I{\times}100{\times}75$]
    \draw[black,ultra thick,->] (2.0,3.5) -- ++(1.7,0.0) -- ++(0.0,-1.0);
    \draw[black,ultra thick,->] (2.0,-0.5) -- ++(2.8,0.0) -- ++(0.0,0.5);
    
    \drawlayer[5.2,0.0,1.4,1.4,magenta!70!black, Conv:, 64@$1{\times}1$]
    \drawblob[5.5,0.0,0.4,1.4,1.4,$64{\times}100{\times}75$]
    \drawlayer[6.3,0.0,1.4,1.4,magenta!70!black, Conv:, 64@$1{\times}1$]
    \drawblob[6.6,0.0,0.4,1.4,1.4,$64{\times}100{\times}75$]
    \drawlayer[7.4,0.0,1.4,1.4,orange!80!black, $Conv^\top$:, 2@$16{\times}16$]
    \drawinputblob[7.7,0.0,0.1,2.0,2.0,scene10_render_v3_N_NW_low_far_5_6_02_133_True_visible_pred300.png]
\end{tikzpicture}
\end{center}
\vspace{-1.0cm}
\caption{MF-FCN}
\label{fig:mf-fcn}
\end{subfigure}

\begin{subfigure}{\textwidth}
\begin{center}
\begin{tikzpicture}[>=stealth]
    \drawblock[-0.5,0.0,2.0,1.4,1.4,Input Block]
    \draw[black,ultra thick,->] (3.5,1.5) -- ++(1.2,0.0);
    
    \pgfmathsetmacro{\lowery}{-1.5}
    \pgfmathsetmacro{\lowxoff}{3.5}    
    % Lower net
    \drawinputblob[0.7-\lowxoff,\lowery,0.1,2.0,2.0,scene10_render_v3_N_NW_low_far_5_6_02_133_True_visible_pred300.png]
    \drawlayer[1.2-\lowxoff,\lowery,2.0,2.0,red!60!black, Conv-Pool:, 20@$5{\times}5$]
    \drawblob[1.45-\lowxoff,\lowery,0.15,1.7,1.7,$20{\times}200{\times}150$]
    \drawlayer[1.9-\lowxoff,\lowery,1.7,1.7,red!60!black, Conv-Pool:, 20@$5{\times}5$]
    \drawblob[2.15-\lowxoff,\lowery,0.15,1.4,1.4,$20{\times}100{\times}75$]
    \drawlayer[2.6-\lowxoff,\lowery,1.4,1.4,red!60!black, Conv-Pool:, 20@$5{\times}5$]
    \draw[black,ultra thick,->] (3.15-\lowxoff,\lowery+1.2) -- ++(5.7,0.0) -- ++(0.0,0.3);
    
    % LSTM input blobs
    \drawblob[4.7,0.0,0.2,1.4,1.4,$32N_I{\times}100{\times}75$]
    \drawblob[5.2,0.0,0.15,1.4,1.4,$20{\times}100{\times}75$]
    \drawblob[5.6,0.0,0.15,1.4,1.4,$20{\times}100{\times}75$]
    \drawblob[5.6,2.0,0.15,1.4,1.4,$20{\times}100{\times}75$]
    
    \node[anchor=east,align=right] at (5.0,-1.0) {{\footnotesize Previous Recurrent State}};
    \draw[black,ultra thick,->] (5.0,-1.0) -- ++(0.75,0.0) -- ++(0.0,1.0);
    \node[anchor=west,align=left] at (7.0,-1.0) {{\footnotesize Next Recurrent State}};
    \draw[black,ultra thick,<-] (7.0,-1.0) -- ++(-0.55,0.0) -- ++(0.0,1.0);
    \node[anchor=east,align=right] at (4.0,3.8) {{\footnotesize Previous Cell State}};
    \draw[black,ultra thick,->] (4.0,3.8) -- ++(1.6,0.0);
    \node[anchor=west,align=left] at (8.0,4.7) {{\footnotesize Next Cell State}};
    
    % Last part of net
    \drawlayer[6.0,0.0,2.83,1.4,green!70!black, LSTM:, 4 gates with 20 $1{\times}1$ kernels each]
    \drawblob[6.3,0.0,0.15,1.4,1.4,$20{\times}100{\times}75$]
    \drawblob[6.3,2.0,0.15,1.4,1.4,$20{\times}100{\times}75$]
    \drawlayer[6.7,0.0,1.4,1.4,magenta!70!black, Conv:, 20@$1{\times}1$]
    \drawblob[7.0,0.0,0.15,1.4,1.4,$20{\times}100{\times}75$]
    \drawlayer[7.4,0.0,1.4,1.4,orange!80!black, $Conv^\top$:, 2@$16{\times}16$]
    \drawinputblob[7.7,0.0,0.1,2.0,2.0,scene10_render_v3_N_NW_low_far_5_6_02_133_True_visible_pred300.png]
    
    \draw[black,ultra thick,->] (7.0,3.5) -- ++(0.8,0.0) -- ++(1.0,1.0);
    
    % Node for pushing everything to the right so it doesn't overlap with the caption.
    \node[] at (-4.0,0.0) {};
\end{tikzpicture}
\end{center}
\vspace{-1.0cm}
\caption{LSTM-FCN}
\label{fig:lstm-fcn}
\end{subfigure}

\caption{The three types of networks we tested. The first is a standard FCN. The second is an FCN that takes in a series of consecutive frames. The final is a recurrent network that uses an LSTM layer to enable the recurrence. As in Figure \ref{fig:input_blocks}, the gray boxes are the feature representations at each level of the network, and the colored squares are the layers that operate on each representation. Gray boxes immediately adjacent indicate channel-wise concatenation (the dashed line in the MF-FCN indicates a concatenation over the range of inputs). $N_I$ indicates the size of the output of the input block, with $N_I = 2$ for the late-fusion blocks and $N_I = 1$ for all other blocks. The LSTM-FCN takes its own output from the previous timestep as input (lower-left), convolves it through 3 layers, and concatenates it with the output of the input block. The LSTM layer is implemented using the layout described in Figure 1 of \citep{greff2015}}
\end{figure*}
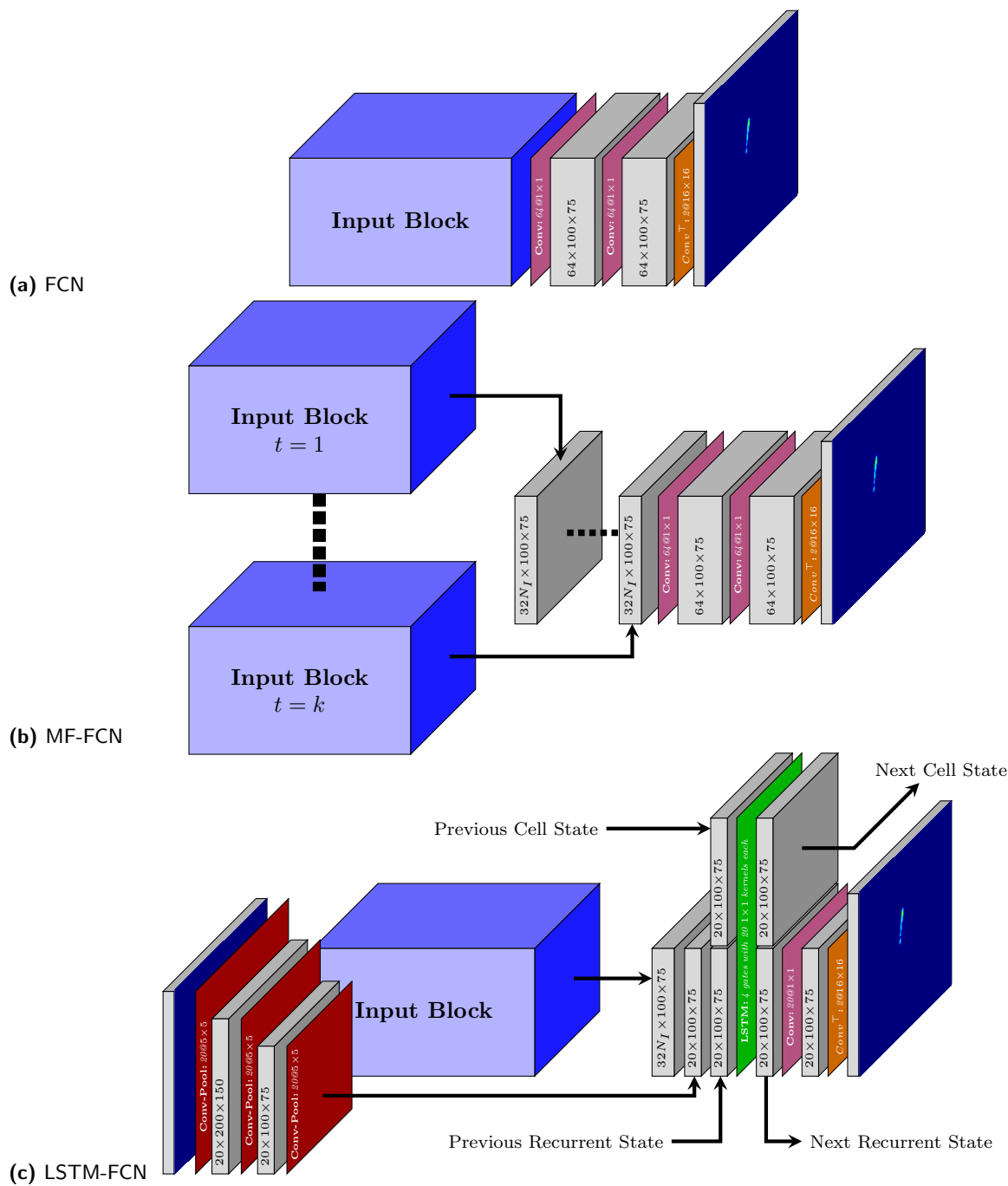

We use 3 different types of networks in this paper:
\begin{description}
    \item [FCN] The first is a standard FCN shown in Figure \ref{fig:fcn}. It takes the output of the input block and passes it through 2 $1{\times}1$ convolutional layers and a final transposed convolution\footnote{Sometimes referred to in the literature as upsampling or deconvolution.} layer (written as $Conv^\top$ in the figure). The $1{\times}1$ convolutional layers take the place of fully-connected layers in a standard neural network. They take only the channels for a single ``pixel'' of the input tensor, acting similar to a fully-connected layer on a network that takes the image patch of the response region for that pixel. Each $1{\times}1$ convolutional layer is followed by a rectified linear filter.
    \item [MF-FCN] The second network type is a multi-frame FCN shown in Figure \ref{fig:mf-fcn}. It takes as input a series of sequential frames, and so has an input block for each frame. Each input block shares parameters, e.g., the first convolutional layer in the first input block has the exact same kernels as the first convolutional layer in the second input block, and so on. The output tensors of all the input blocks are concatenated together channel-wise. This is then feed to a network structured identical to the structure for the previous network (two $1{\times}1$ convolutional layers followed by a transposed convolution layer).
    \item [LSTM-FCN] The last network type is a recurrent network that utilizes a long short-term memory (LSTM) layer \citep{hochreiter1997} shown in Figure \ref{fig:lstm-fcn}. It takes the recurrent state, the cell state, and the output image from the previous timestep in addition to the frame from the current timestep as input. The output tensor from the input block is concatenated channel-wise with the recurrent state, and with the output image from the previous timestep after it has been passed through 3 conv-pool layers. The resulting tensor is then fed into the LSTM layer along with the cell state. The LSTM layer uses the cell state to ``gate'' the other inputs, that is, the cell state controls how the information in the other inputs passes through the LSTM. The resulting output we refer to as the ``recurrent state'' because it is feed back into the LSTM on the next timestep. The LSTM layer also updates the cell state for use on the next timestep. In addition to being used in the next timestep, this recurrent state is also fed through a $1{\times}1$ convolutional layer and then a transposed convolution layer to generate the output image for this timestep. To maintain the fully-convolutional nature of our network, we replace all the gates in the LSTM layer with $1{\times}1$ convolutional layers. Please refer to Figure~1 of \citep{greff2015} for a more detailed description of the LSTM layer.
\end{description}

\section{Evaluation}
\label{sec:evaluation}

\subsection{Simulated Data Set}

We evaluate all three of our network types on the tasks of {\it detection} and {\it tracking} on the simulated dataset.
For this dataset, we use only single input image types, so all networks are implemented with the standard input block (Figure \ref{fig:standard_input}).
We report the results as precision and recall curves, that is, for every value between 0 and 1, we threshold the confidence of the network's labels and compute the corresponding precision and recall based on the pixel-wise accuracy.
We also report the area-under-curve score for the precision and recall curves.
Additionally, we report precision and recall curves for various amounts of ``slack,'' i.e., we count a positive classification as correct if it is within $n$ pixels of a true positive pixel, where $n$ is the slack value.
This slack evaluation allows us to differentiate networks that are able to detect or track the liquid, albeit somewhat imprecisely, versus networks that fire on parts of the image not close to liquid.

We evaluate our networks on two subsets of the simulated dataset: the {\it fixed-view} set and the {\it multi-view} set.
The {\it fixed-view} set contains all the data for which the camera was directly across from the table (camera azimuth of 0 or 180 degrees) and the camera was level with the table (low camera height), or 1,266 of the pouring sequences.
Due to the cylindrical shape of all the source and target containers, this is the set of data for which the mapping from the full 3D state of the simulator to a 2D representation is straightforward, which is useful for our networks as they operate only on 2D images.
The {\it multi-view} set contains all data from the simulated dataset, including all camera viewpoints.
The mapping from 3D to 2D for this set is not as straightforward.

\subsubsection{Detection}

For the task of {\it detection}, we trained all three networks in a similar manner.
Due to the fact that the vast majority of pixels in any sequence are {\it not-liquid} pixels, we found that trying to train directly on the full pouring sequences resulted in networks that settled in a local minima classifying all pixels as {\it not-liquid}.
Instead, we first pre-train each network for 61,000 iterations on crops of the images and sequences around areas with large amounts of liquid (due to the increased complexity of the LSTM-FCN, we initialize the pre-training LSTM-FCN with the weights of the pre-trained single-frame FCN).
We then train the networks for an additional 61,000 iterations on full images and sequences.
This is only possible because our networks are fully-convolutional, which allows them to have variable sized inputs and outputs.
Additionally, we also employ gradient weighting to counteract the large imbalance between positive and negative pixels.
We multiply the gradient from each {\it not-liquid} pixel by 0.1 so that the error from the {\it liquid} pixels has a larger effect on the learned weights.

The full input images to our networks were scaled to $400{\times}300$.
The crops taken from these images were $160{\times}160$.
The single-frame networks were trained with a batch size of 32.
The multi-frame networks were given a window of 32 frames as input and were trained with a batch size of 1.
The LSTM networks were unrolled for 32 frames during training (i.e., the gradients were propagated back 32 timesteps) and were trained with a batch size of 5.
We used the mini-batch gradient descent method Adam \citep{kingma2014} with a learning rate of 0.0001 and default momentum values.
All error signals were computed using the softmax with loss layer built-into Caffe \citep{jia2014}.

\subsubsection{Tracking}

For the task of {\it tracking}, we trained the networks on segmented object labels (Figure \ref{fig:obj_vis}).
That is, assuming we already have good detectors for what is visible in the scene, can the robot find the liquid that is not visible?
Note that here we use the ground truth labels as shown in Figure \ref{fig:obj_vis} and not the output of the detection network as input to the tracking network, however we do also evaluate combining the two (as described in the next section).
Since the input image is already somewhat structured, we scale it down to $130{\times}100$.
Unlike for {\it detection}, here we don't pre-train the networks on crops, but we do utilize the same gradient weighting scheme.
We use the same training parameters as for {\it detection} with the exception that we unroll the LSTM network for 160 timesteps.
For {\it tracking}, we use only the {\it fixed-view} set.

\subsubsection{Combined Detection \& Tracking}

Finally, we also evaluate performing combined {\it detection} \& {\it tracking} with a single network.
The networks take in the same $400{\times}300$ images that the {\it detection} networks take, and output the location of {\it all} liquid in the scene.
We initialize these networks with the weights of their corresponding {\it detection} network and train them on full images.
We use the same gradient weighting scheme as for the two tasks separately.
We train the networks for combined {\it detection} \& {\it tracking} using the same learning parameters as for training the {\it detection} networks.

\subsection{Robot Data Set}

For the robot dataset, we evaluate our networks only on the task of {\it detection} because our thermal camera can only see {\it visible} liquid and not liquid occluded by the containers.
However, {\it detection} on the robot dataset is more challenging than on the simulated dataset as there is less data to train on.
This is a general problem in robotics with deep learning.
Deep neural networks require vast amounts of data to train on, but it is difficult to collect this much data on a robot.
While there have been some proposed solutions for specific problems \citep{levine2015,tzeng2015}, there is no generally accepted methodology for solving this issue.
Here we evaluate utilizing different types of input images to help prevent the networks from overfitting on the smaller amount of data.

Specifically, we train networks for each of the following input types (with the corresponding input block in parentheses):
\begin{itemize}
    \item RGB (standard input block)
    \item Grayscale (standard input block)
    \item Optical Flow (standard input block)
    \item RGB+Optical Flow (early-fusion input block)
    \item Grayscale+Optical Flow (early-fusion input block)
    \item RGB+Optical Flow (late-fusion input block)
    \item Grayscale+Optical Flow (late-fusion input block)
\end{itemize}
We train LSTM networks on all of these different types of inputs, as well as the single-frame networks since they are necessary to initialize the weights of the LSTM networks.
We use the same learning parameters and training methodology as for {\it detection} on simulated data (pre-training on crops, gradient weighting, etc.).
For brevity, we report our results as area under the curve for the precision and recall curves for each network.

Unlike for the simulated dataset, where the train and test sets are created by dividing the dataset, for the robot dataset, we created an explicit test set.
To test the robot's generalization ability, we used target containers that did not appear in the train set.
We train all networks on the entire dataset and test on this explicit test set.
To gauge the extent to which our networks overfit to their training set, we report the performance of the networks on both the train set and the test set.

\subsection{Baseline for the Robot Data Set}
\label{sec:baseline}

For comparison, we implement as a baseline the liquid detection methodology described in \citep{yamaguchi2016} for the {\it detection} task on the real robot dataset.
We briefly describe that implementation here.
For each image in a sequence, we compute the dense optical flow using the same methodology as for the neural network method.
Next, we compute the magnitude of the flow vector for each pixel, and create a resulting flow magnitude image.
We then perform the following steps to filter the image as described in Section II.A of \citep{yamaguchi2016}:
\begin{enumerate}
    \item Erode the image with a square kernel of size 3.
    \item Dilate the image with a square kernel of size 3.
    \item \label{itm:A} Apply a temporal filter with size 5 to each pixel, replacing the value in the pixel with the OR of all the pixels covered by the filter.
    \item Dilate the image with a square kernel of size 7.
    \item Erode the image with a square kernel of size 11.
    \item Dilate the image with a square kernel of size 13.
    \item \label{itm:B} Convolve the image with a $12{\times}1$ filter (12 pixels high, 1 wide) where each value in the filter is 1/12.
    \item Use the result of the prior step to apply as a mask to the result of step \ref{itm:A}.
    \item Apply the same filter to the result as in step \ref{itm:B}.
    \item Scale the magnitudes in the resulting image to be in the range 0 to 1.
\end{enumerate}
Note that some of the hyper parameters we used are adjusted from the values used in \citep{yamaguchi2016} to account for the difference in image sizes ($640{\times}480$ vs. $400{\times}300$ in this paper).

There are two primary differences between our implementation and the implementation in \citep{yamaguchi2016}. 
The first is the way in which background motion is removed. 
In that paper, the authors utilized stereo RGB cameras to localize the optical flow in 3D, and then fixed a region of interest around the liquid, removing all motion not in that region.
In our work, we use a single camera, however our camera also uses structured infrared light combined with an infrared camera to determine the depth of each point in the image.
In order to remove background motion, we generate a mask by including only pixels whose value is closer than one meter from the camera.
We then smooth this mask by eroding, then dilating twice, then eroding again, all with a square kernel of size 7.
This mask is applied to the optical flow before applying the filter steps above.

The second difference between our implementation and that in \citep{yamaguchi2016} is that, in order to be comparable to our methodology, it must compute a distribution over the class labels, rather than a single label.
In \citep{yamaguchi2016} they compute only a binary mask for each image.
However, in the following section we utilize precision-recall curves to compare our methods, which requires a probability distribution over class labels to compute.
We approximate this distribution using the magnitude of the flow at each pixel, that is, the more a pixel is moving, the more likely it is liquid.

\section{Results}
\label{sec:results}

\subsection{Simulated Data Set}
\label{sec:results_sim}

\subsubsection{Detection}

\begin{figure}
    \centering
    \setlength{\fboxsep}{0pt}
    \setlength{\fboxrule}{1pt}
    \setlength{\unitlength}{1.0cm}
    \scalebox{0.8}{
    \begin{picture}(10.0,7.7)
        \put(0.0,0.0){\fbox{\includegraphics[width=2.0cm]{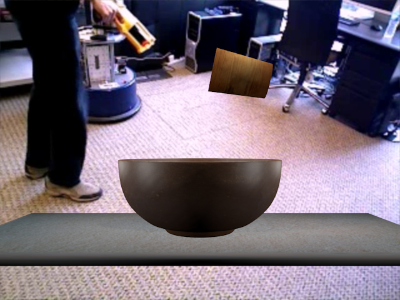}}}
        \put(2.0,0.0){\fbox{\includegraphics[width=2.0cm]{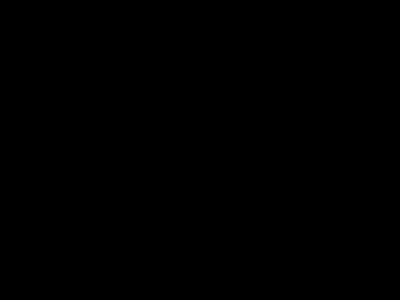}}}
        \put(4.0,0.0){\fbox{\includegraphics[width=2.0cm]{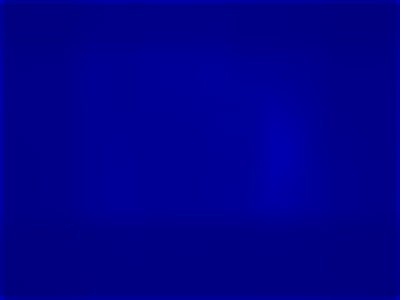}}}
        \put(6.0,0.0){\fbox{\includegraphics[width=2.0cm]{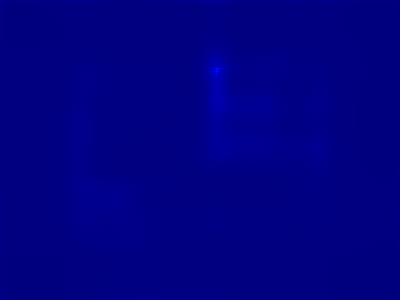}}}
        \put(8.0,0.0){\fbox{\includegraphics[width=2.0cm]{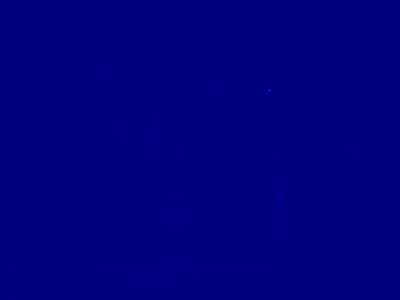}}}
        
        \put(0.0,1.55){\fbox{\includegraphics[width=2.0cm]{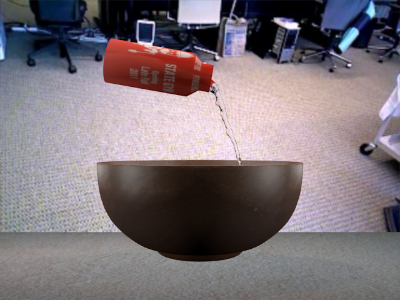}}}
        \put(2.0,1.55){\fbox{\includegraphics[width=2.0cm]{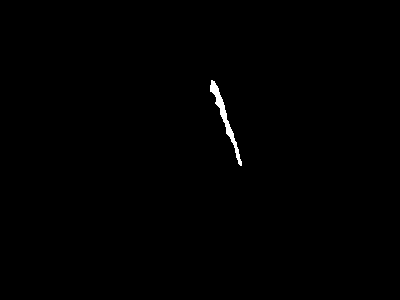}}}
        \put(4.0,1.55){\fbox{\includegraphics[width=2.0cm]{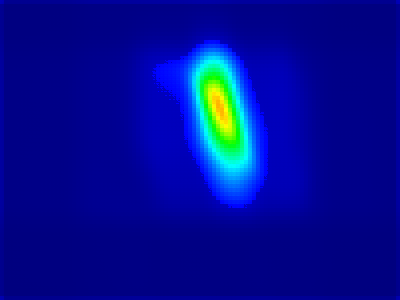}}}
        \put(6.0,1.55){\fbox{\includegraphics[width=2.0cm]{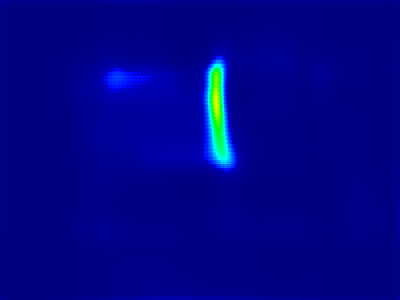}}}
        \put(8.0,1.55){\fbox{\includegraphics[width=2.0cm]{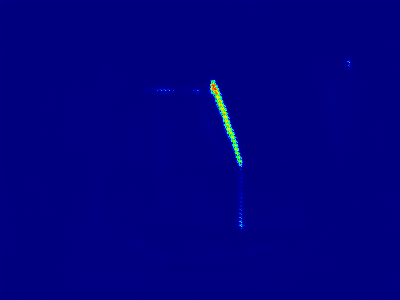}}}

        \put(0.0,3.1){\fbox{\includegraphics[width=2.0cm]{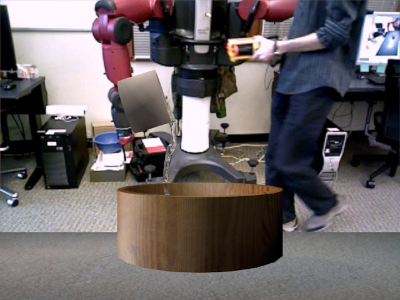}}}
        \put(2.0,3.1){\fbox{\includegraphics[width=2.0cm]{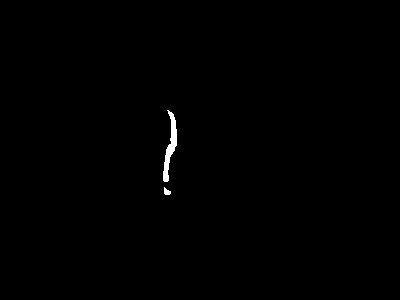}}}
        \put(4.0,3.1){\fbox{\includegraphics[width=2.0cm]{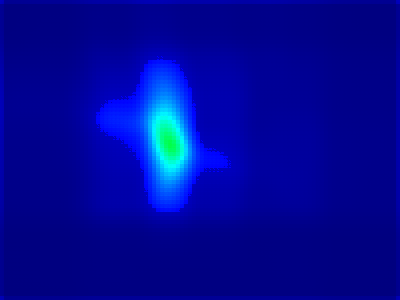}}}
        \put(6.0,3.1){\fbox{\includegraphics[width=2.0cm]{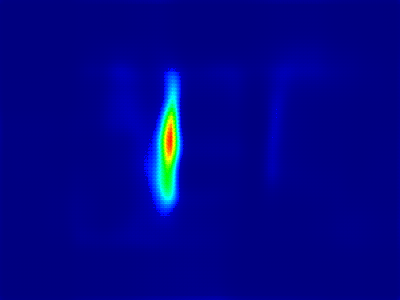}}}
        \put(8.0,3.1){\fbox{\includegraphics[width=2.0cm]{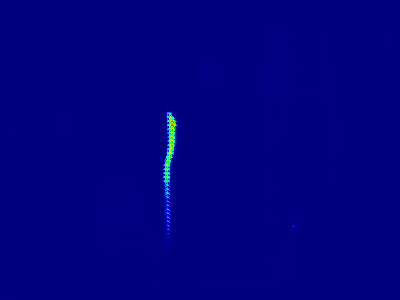}}}
        
        \put(0.0,4.65){\fbox{\includegraphics[width=2.0cm]{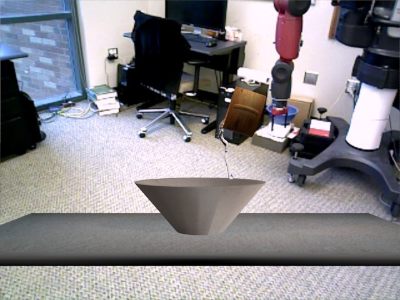}}}
        \put(2.0,4.650){\fbox{\includegraphics[width=2.0cm]{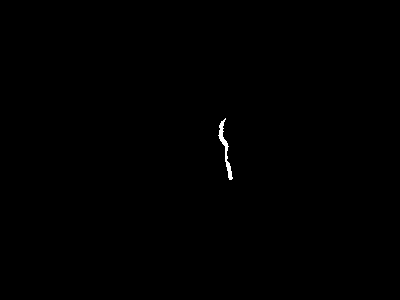}}}
        \put(4.0,4.65){\fbox{\includegraphics[width=2.0cm]{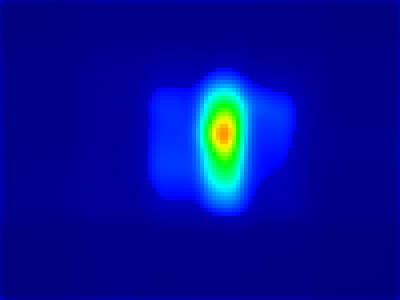}}}
        \put(6.0,4.65){\fbox{\includegraphics[width=2.0cm]{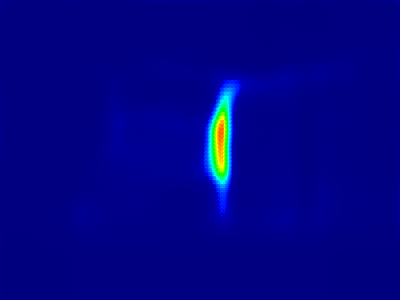}}}
        \put(8.0,4.65){\fbox{\includegraphics[width=2.0cm]{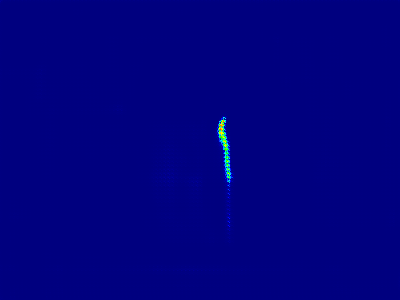}}}
        
        \put(0.0,6.2){\fbox{\includegraphics[width=2.0cm]{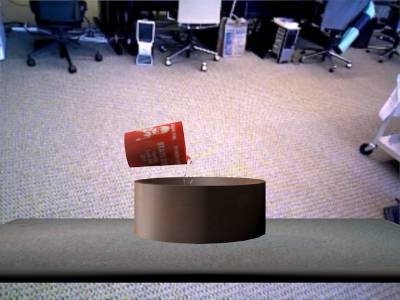}}}
        \put(2.0,6.2){\fbox{\includegraphics[width=2.0cm]{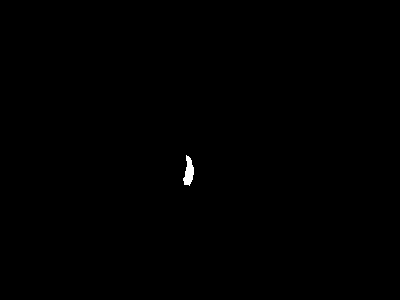}}}
        \put(4.0,6.2){\fbox{\includegraphics[width=2.0cm]{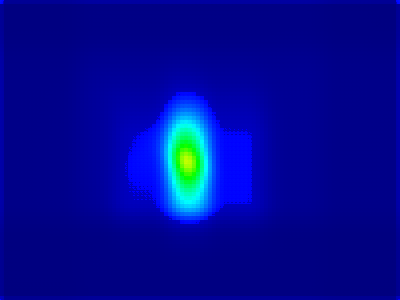}}}
        \put(6.0,6.2){\fbox{\includegraphics[width=2.0cm]{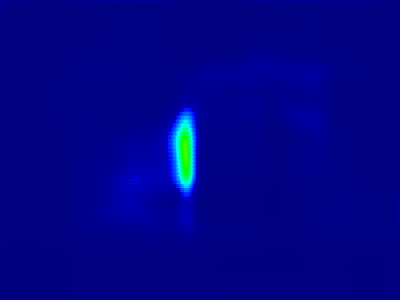}}}
        \put(8.0,6.2){\fbox{\includegraphics[width=2.0cm]{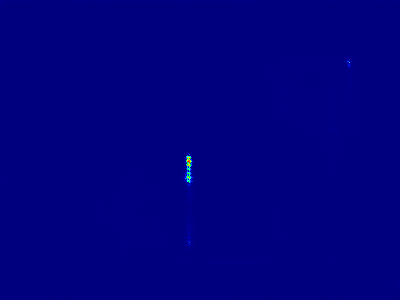}}}
        
        \put(0.5,7.9){{\bf Input}}
        \put(2.4,7.9){{\bf Labels}}
        \put(4.6,7.9){{\bf FCN}}
        \put(6.25,7.9){{\bf MF-FCN}}
        \put(8.05,7.9){{\bf LSTM-FCN}}
    \end{picture}
    }
    \caption{Example frames from the 3 network types on the {\it detection} task on the simulated dataset. The sequences shown here were randomly selected from the test set and the frame with the largest amount of liquid visible was selected. The last sequence was selected to show how the networks perform when no liquid is present.}
    \label{fig:results_sim_detection1}
\end{figure}
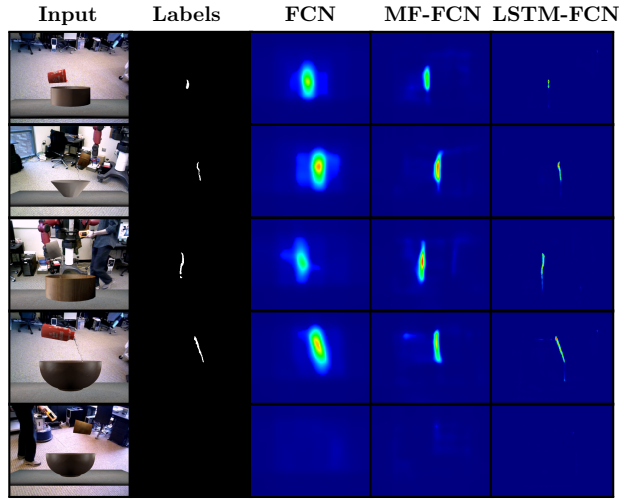

\setlength{\objectsize}{4.0cm}
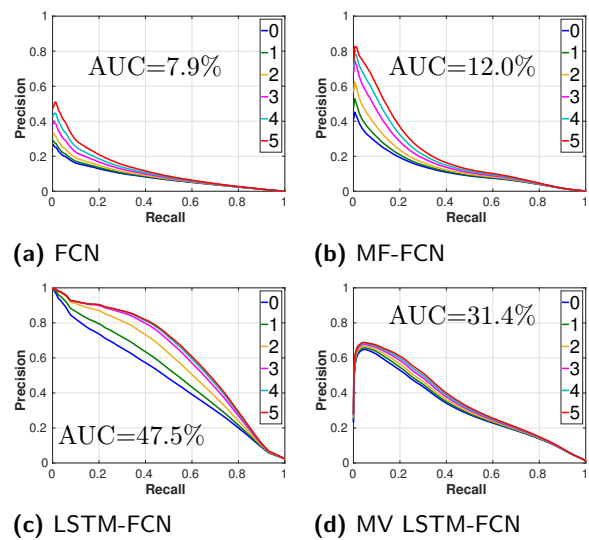
\begin{figure}
    \centering
    \setlength{\unitlength}{1.0cm}
    \begin{subfigure}{\objectsize}
        \begin{picture}(4.0,3.0)
            \put(0.0,0.0){\includegraphics[width=4.0cm]{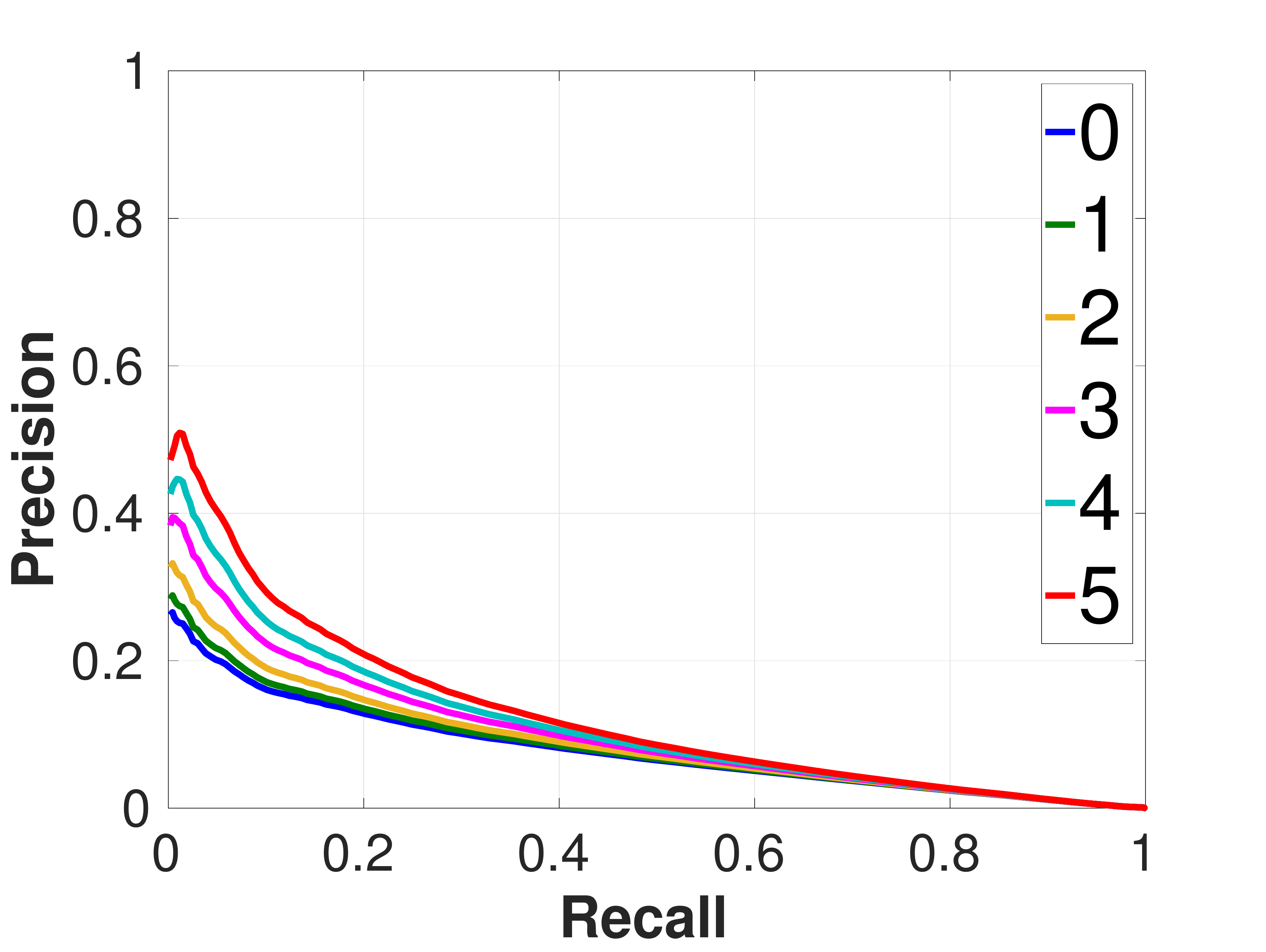}}
            \put(1.0,2.0){{AUC=7.9\%}}
        \end{picture}
        \caption{FCN}
        \label{fig:results_sim_detection_fcn}
    \end{subfigure}%
    \begin{subfigure}{4.0cm}
        \begin{picture}(4.0,3.0)
            \put(0.0,0.0){\includegraphics[width=4.0cm]{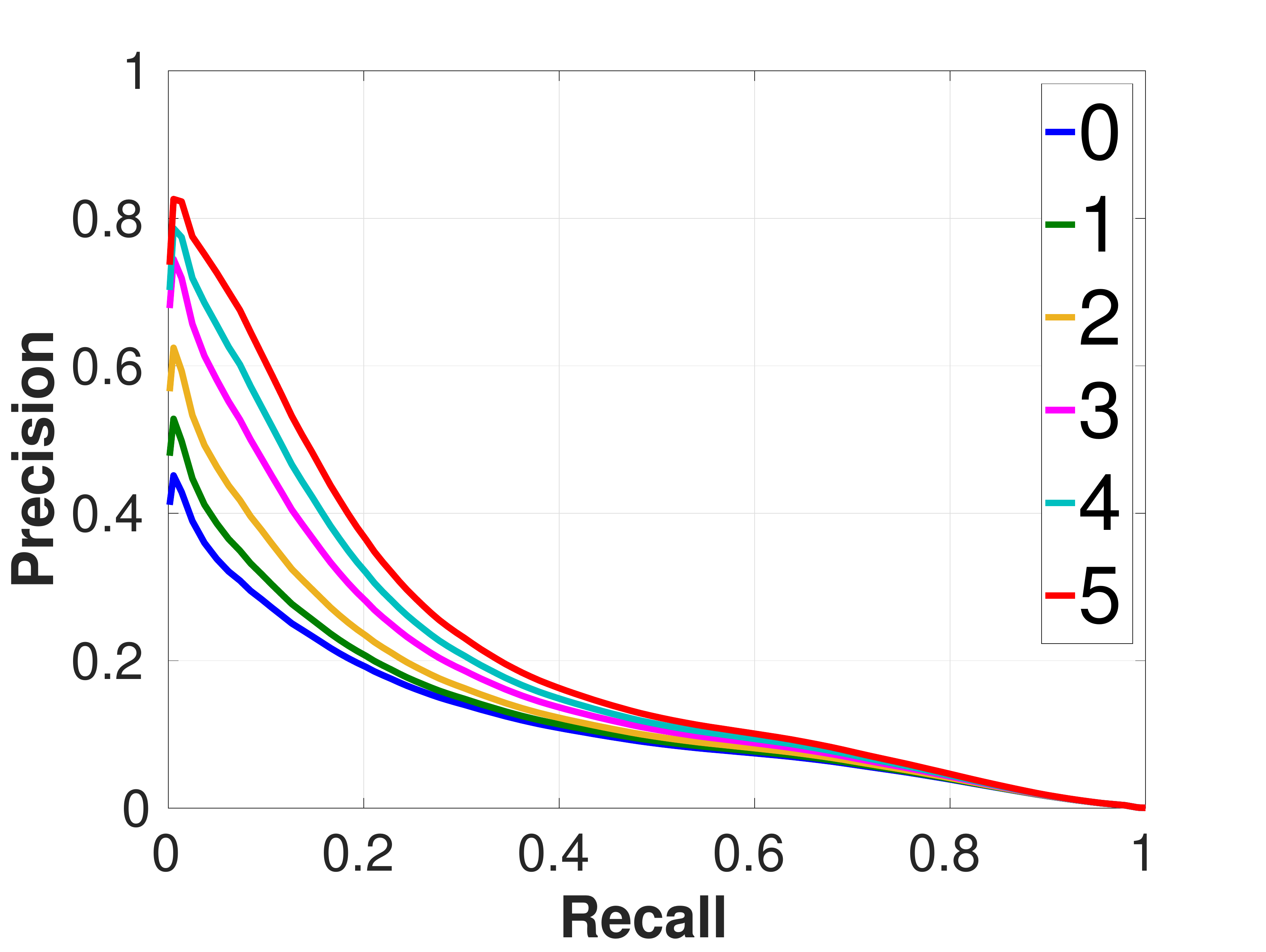}}
            \put(1.0,2.0){{AUC=12.0\%}}
        \end{picture}
        \caption{MF-FCN}
        \label{fig:results_sim_detection_mffcn}
    \end{subfigure}
    
    \begin{subfigure}{4.0cm}
        \begin{picture}(4.0,3.0)
            \put(0.0,0.0){\includegraphics[width=4.0cm]{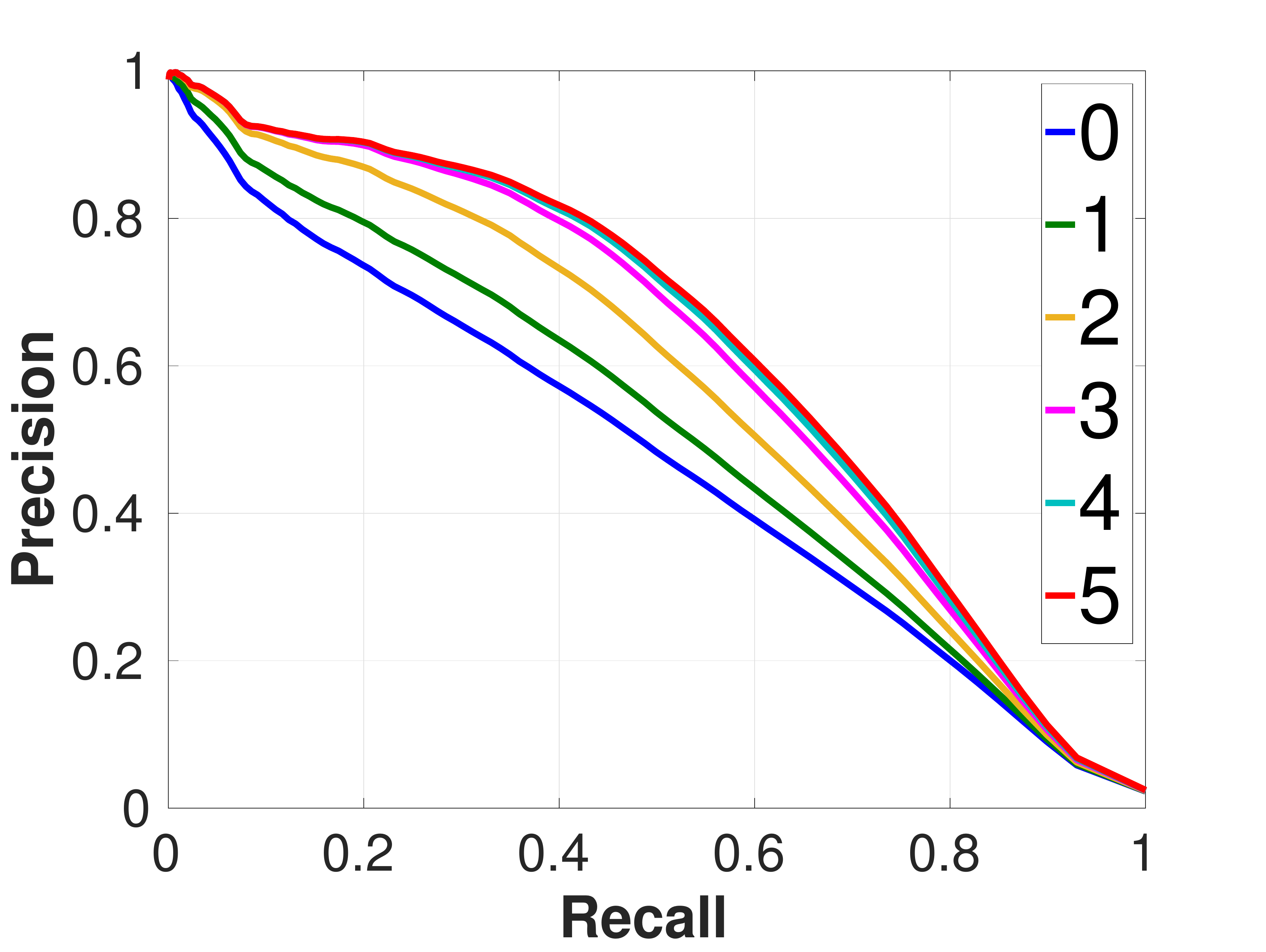}}
            \put(0.6,0.7){{AUC=47.5\%}}
        \end{picture}
        \caption{LSTM-FCN}
        \label{fig:results_sim_detection_lstm}
    \end{subfigure}%
    \begin{subfigure}{4.0cm}
        \begin{picture}(4.0,3.0)
            \put(0.0,0.0){\includegraphics[width=4.0cm]{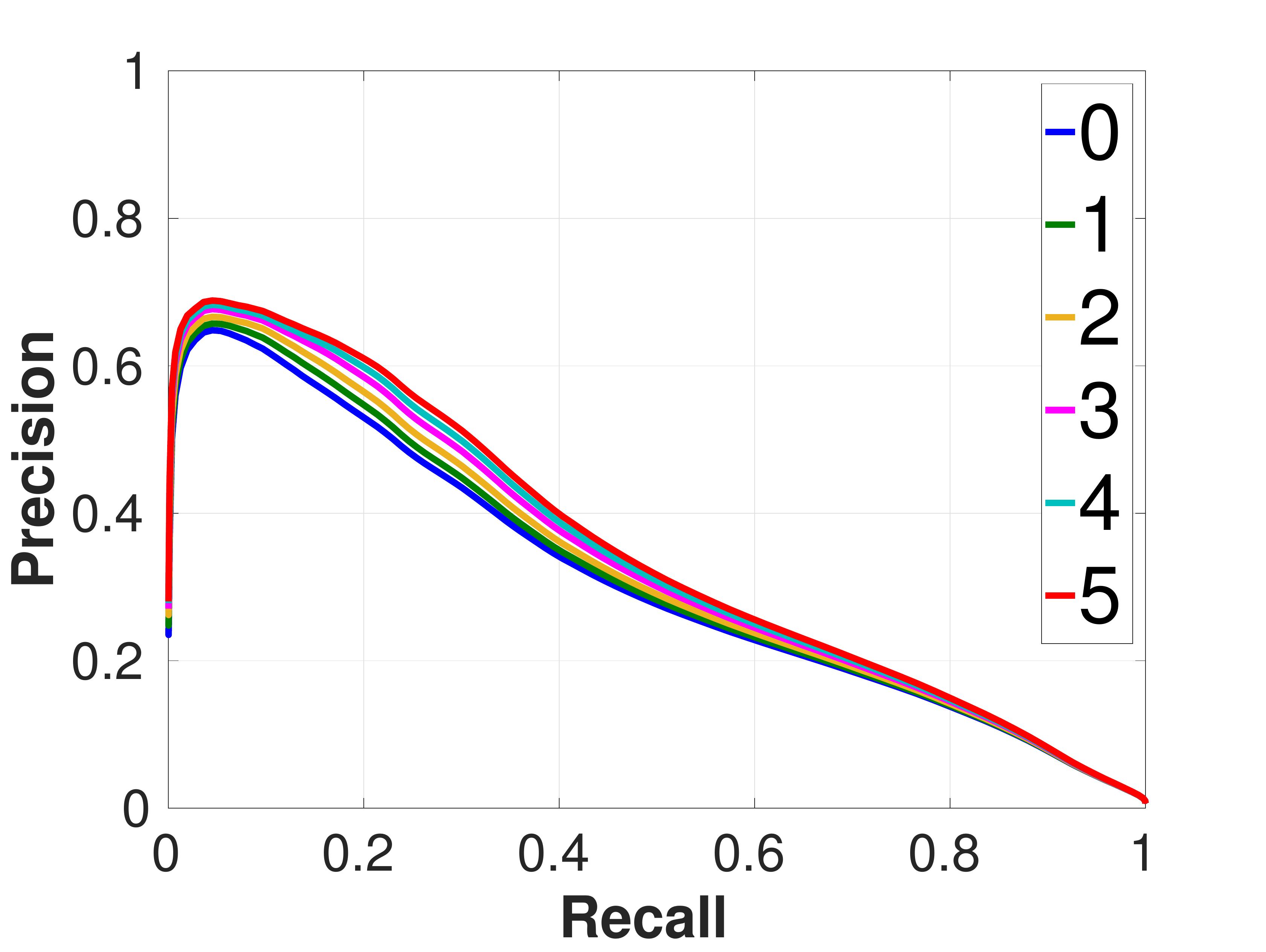}}
            \put(1.0,2.3){{AUC=31.4\%}}
        \end{picture}
        \caption{MV LSTM-FCN}
        \label{fig:results_sim_detection_mvlstm}
    \end{subfigure}
    \caption{Precision-recall curves for {\it detection} on the simulated dataset. The first three show the curves for the three network types on the {\it fixed-view} subset. The last graph shows the performance of the LSTM network on the {\it multi-view} subset. The different lines show the different amounts of slack, i.e., how far a positive classification can be from a true positive to still count as correct. The area under the curve (AUC) is shown for the 0 slack curve.}
    \label{fig:results_sim_detection2}
\end{figure}

% Number of params with RGB input (379,296):
%  FCN-414,336
%  MF-FCN-477,824
%  LSTM-FCN-437,508

Figures \ref{fig:results_sim_detection1} and \ref{fig:results_sim_detection2} show the results of training our networks for the {\it detection} task on the simulated dataset.
Figure \ref{fig:results_sim_detection1} shows the output of each network on example frames.
From this figure it is clear that all networks have the ability to at least detect the presence of liquid.
However, it is also clear that the MF-FCN is superior to the single-frame FCN, and the LSTM-FCN is superior to the MF-FCN.
This aligns with our expectations: As we integrate more temporal information (the FCN sees no temporal information, the MF-FCN sees a small window, and the LSTM-FCN has a full recurrent state), the networks perform better.
The quantitative results in Figure \ref{fig:results_sim_detection2} confirm these qualitative results.
For reference, all the networks have a very similar number of parameters (414,336, 477,824, and 437,508 for the FCN, MF-FCN, and LSTM-FCN networks respectively), so it is clear that the success of the LSTM-FCN is not simply due to having more parameters and ``remembering'' the data better, but that it actually integrates the temporal information better.

Since the LSTM-FCN outperformed the other two network types by a significant margin, we evaluated it on the {\it multi-view} set from the simulated dataset.
The performance is shown in Figure \ref{fig:results_sim_detection_mvlstm}.
Even with the large increase in camera viewpoints, the network is still able to detect liquid with only a relatively small loss in performance.
These results combined with the performance of the LSTM-FCN in Figure \ref{fig:results_sim_detection_lstm} clearly show that it is the best network for performing detection and is the reason we focus on this network for detection on the robot dataset.

\subsubsection{Tracking}

\begin{figure}
    \centering
    \setlength{\unitlength}{1.0cm}
    \begin{subfigure}{4.0cm}
        \begin{picture}(4.0,3.0)
            \put(0.0,0.0){\includegraphics[width=4.0cm]{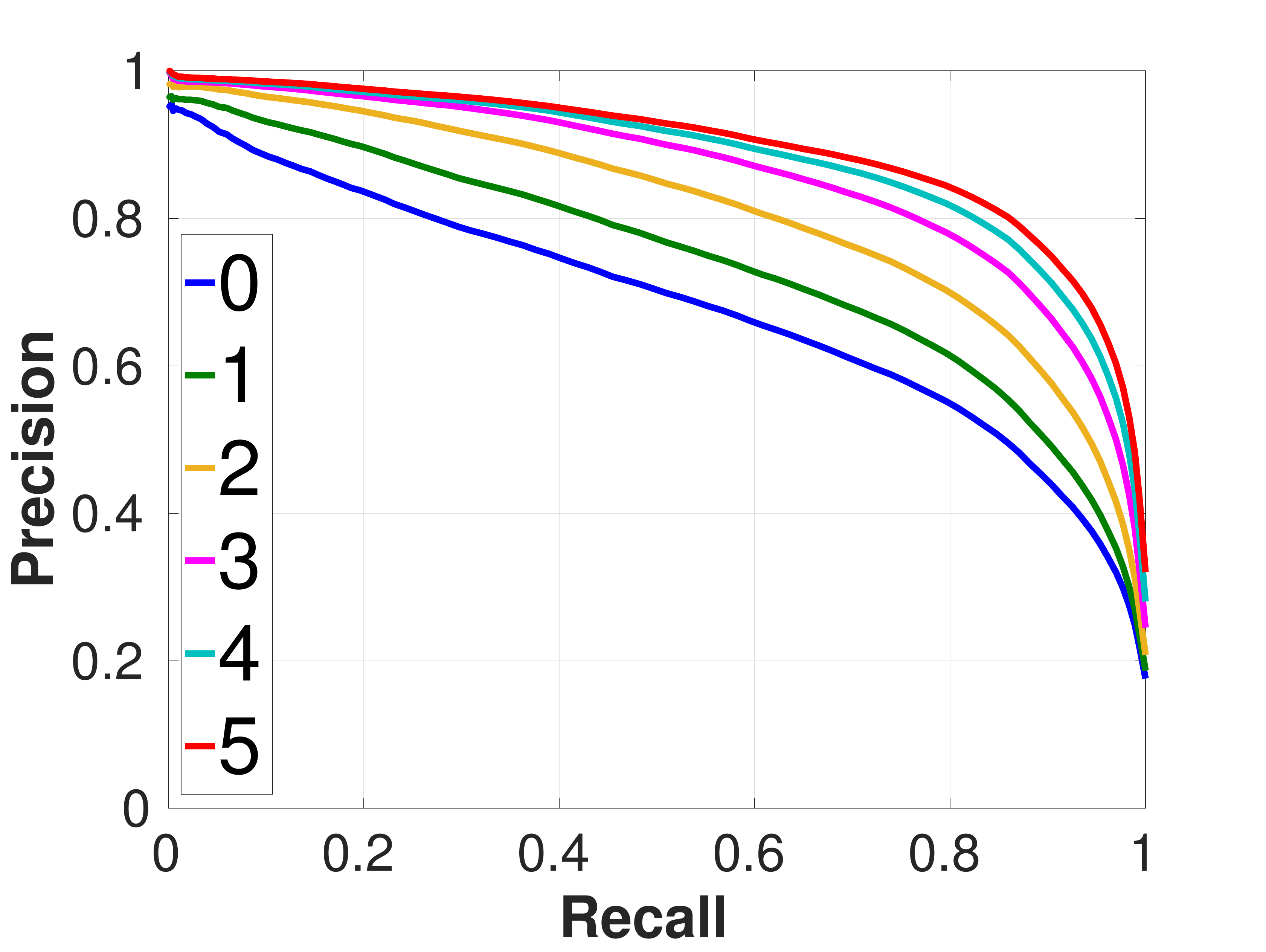}}
            \put(1.0,0.6){{AUC=68.3\%}}
        \end{picture}
        \caption{FCN}
        \label{fig:results_sim_tracking_fcn}
    \end{subfigure}%
    \begin{subfigure}{4.0cm}
        \begin{picture}(4.0,3.0)
            \put(0.0,0.0){\includegraphics[width=4.0cm]{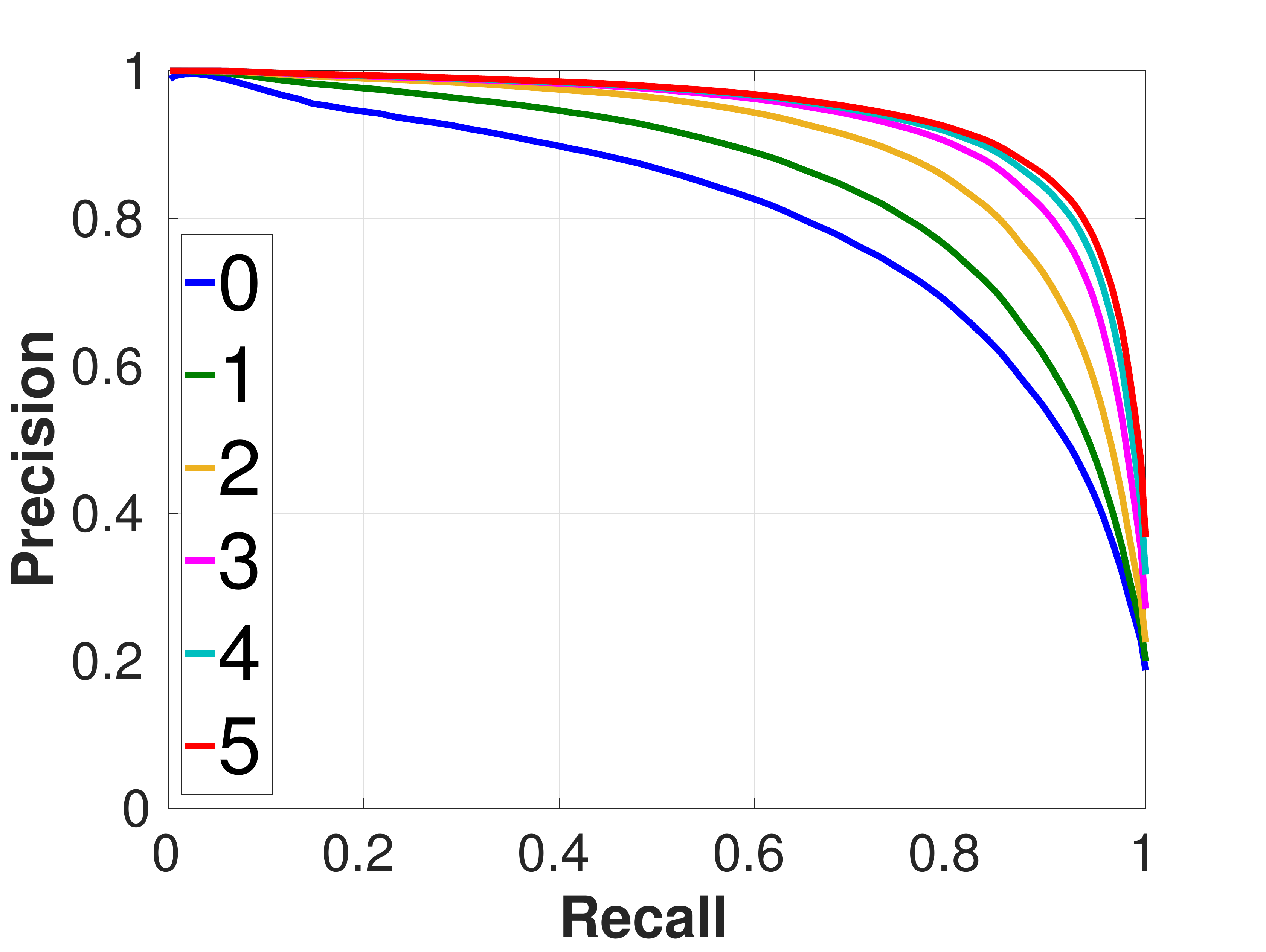}}
            \put(1.0,0.6){{AUC=80.6\%}}
        \end{picture}
        \caption{MF-FCN}
        \label{fig:results_sim_tracking_mffcn}
    \end{subfigure}
    
    \begin{subfigure}{4.0cm}
        \begin{picture}(4.0,3.0)
            \put(0.0,0.0){\includegraphics[width=4.0cm]{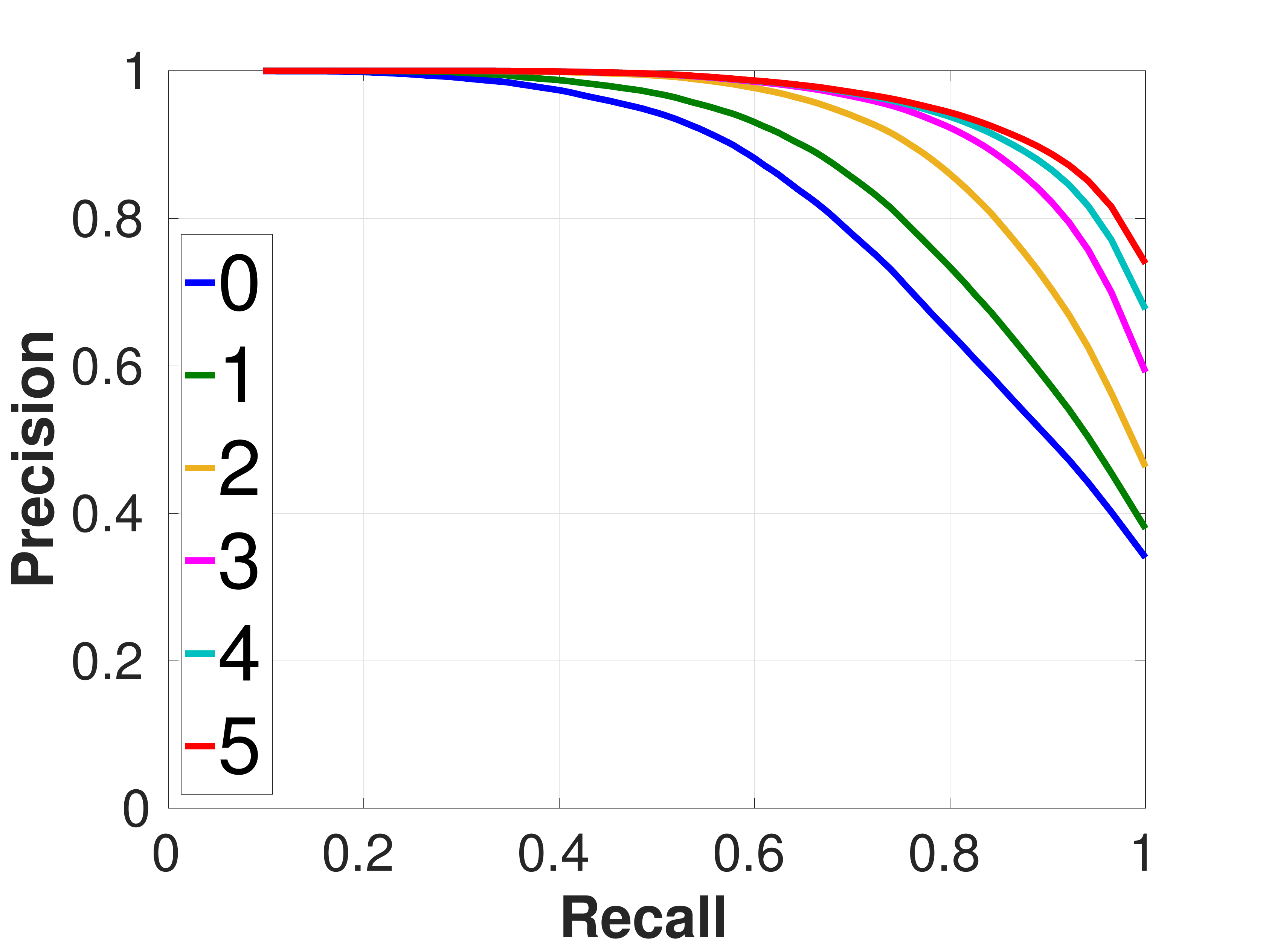}}
            \put(1.0,0.6){{AUC=84.0\%}}
        \end{picture}
        \caption{LSTM-FCN}
        \label{fig:results_sim_tracking_lstm}
    \end{subfigure}%
    \begin{subfigure}{4.0cm}
        \begin{picture}(4.0,3.0)
            \put(0.0,0.0){\includegraphics[width=4.0cm]{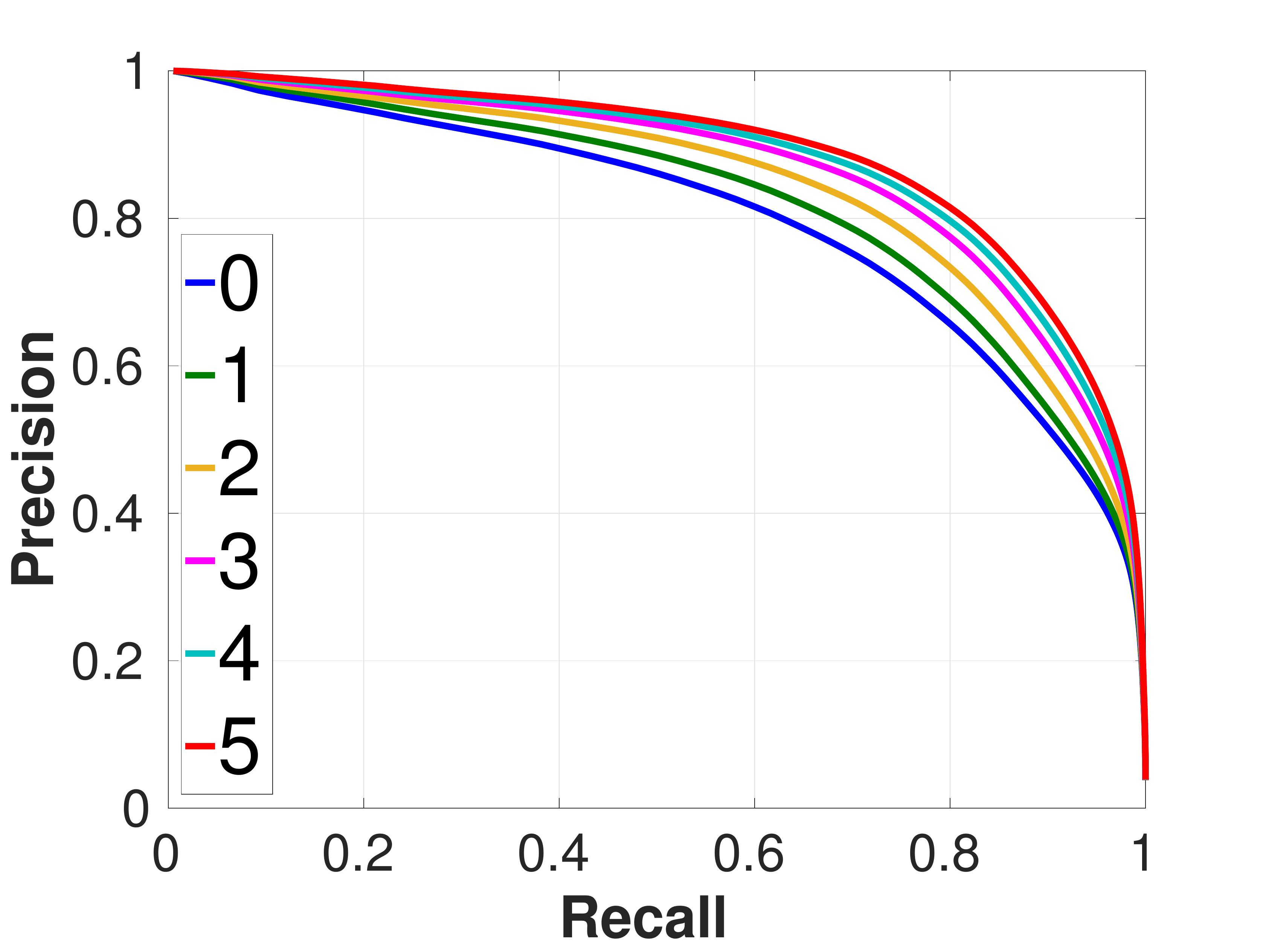}}
            \put(1.0,0.6){{AUC=80.0\%}}
        \end{picture}
        \caption{Combined LSTM-FCN}
        \label{fig:results_detection_tracking}
    \end{subfigure}
    \caption{The precision-recall curves for the {\it tracking} task on the simulated dataset. The first 3 show the performance of the three network types on the {\it tracking} task alone. The last graph shows the performance of the LSTM-FCN on the combined {\it detection} \& {\it tracking} task. Similar to Figure~\ref{fig:results_sim_detection2}, the different lines show the different amounts of slack, i.e., how far a positive classification can be from a true positive to still count as correct. The area under the curve (AUC) is shown for the 0 slack curve.}
    \label{fig:results_sim_tracking}
\end{figure}
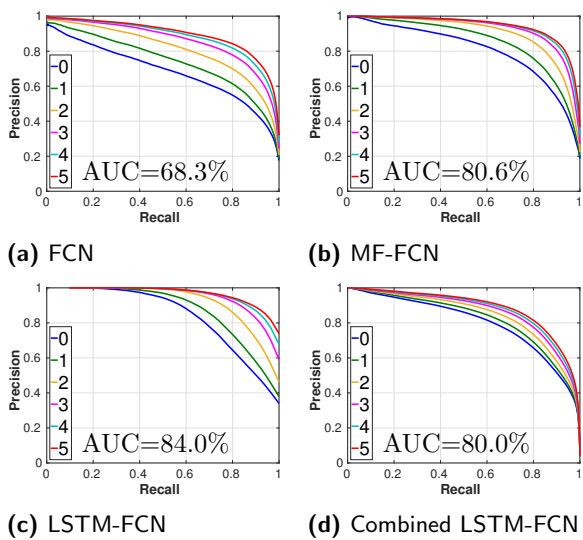

Figure \ref{fig:results_sim_tracking} shows the performance of the 3 network types on the {\it tracking} task.
As expected, the only network with an explicit memory, the LSTM-FCN, performs the best.
However, the other two networks perform better than would be expected of networks with no memory capability.
This is due to the fact that, given segmented input, the networks can infer where some of the liquid {\it likely} is.
Although it is clear that LSTM-FCNs are best suited for this task.

We additionally tested the LSTM-FCN on the combined {\it detection} \& {\it tracking} task.
The results are shown in Figure \ref{fig:results_detection_tracking}.
The network in this case is able to do quite well, with only a minor drop in performance as compared to the LSTM-FCN on the {\it tracking} task alone.

\subsection{Robot Data Set}

\newcommand{\traintest}[2]{$\frac{\textrm{#1}}{\textrm{{\bf #2}}}$}
\begin{figure}
    \setlength{\unitlength}{1.0cm}
    \begin{subfigure}{\textwidth}
        \renewcommand{\arraystretch}{1.5}
        \scalebox{1.025}{
        \begin{tabular}{r | c | c | c |}
        & \multicolumn{3}{c}{Optical Flow} \\\hline
        \traintest{train}{test} & None & Early-Fusion & Late-Fusion \\\hline
        {\bf RGB} & \traintest{76.1\%}{18.2\%} & \traintest{63.3\%}{17.5\%} & \traintest{48.8\%}{25.1\%} \\\hline
        {\bf Gray} & \traintest{70.7\%}{22.0\%} & \traintest{57.0\%}{30.5\%} & \traintest{73.8\%}{24.1\%} \\\hline
        {\bf Neither} & \cellcolor{gray} & \multicolumn{2}{| c |}{\traintest{41.3\%}{35.1\%}} \\\hline
        \end{tabular}
        }
        \caption{FCN}
        \label{fig:results_robot_fcn}
    \end{subfigure}
    
    \begin{subfigure}{\textwidth}
        \renewcommand{\arraystretch}{1.5}
        \scalebox{1.025}{
        \begin{tabular}{r | c | c | c |}
        & \multicolumn{3}{c}{Optical Flow} \\\hline
        \traintest{train}{test} & None & Early-Fusion & Late-Fusion \\\hline
        {\bf RGB} & \traintest{95.1\%}{23.7\%} & \traintest{48.3\%}{13.4\%} & \traintest{93.1\%}{33.1\%} \\\hline
        {\bf Gray} & \traintest{82.9\%}{17.9\%} & \traintest{81.5\%}{49.4\%} & \traintest{92.8\%}{41.7\%} \\\hline
        {\bf Neither} & \cellcolor{gray} & \multicolumn{2}{| c |}{\traintest{82.7\%}{37.4\%}} \\\hline
        \end{tabular}
        }
        \caption{LSTM-FCN}
        \label{fig:results_robot_lstm}
    \end{subfigure}
    
    \caption{The area under the curve (AUC) for the precision-recall curves for the networks for the {\it detection} task on the robot dataset. The top table shows the AUC for the single-frame FCN; the bottom shows the AUC for the LSTM-FCN. The tables show the AUC for different types of input, with rows for different types of image data (RGB or grayscale) and the columns for different types of optical flow (none, early-fusion, or late-fusion). Each cell shows the AUC on the train set (upper) and the AUC on the test set (lower), all computed with 0 slack.}
    \label{fig:results_robot_detection2}
\end{figure}

\newcommand{\figfontsize}{\scriptsize}
\begin{figure}
    \centering
    \setlength{\fboxsep}{0pt}
    \setlength{\fboxrule}{1pt}
    \setlength{\unitlength}{1.0cm}
    \scalebox{0.9}{
    \begin{picture}(9.0,15.5)
        \put(0.0,13.95){\fbox{\includegraphics[width=2.0cm]{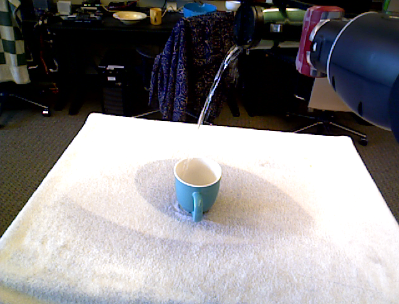}}}
        \put(0.0,12.4){\fbox{\includegraphics[width=2.0cm]{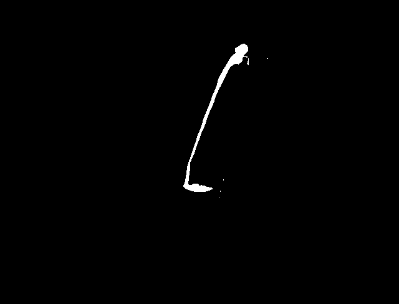}}}
        \put(0.0,10.85){\fbox{\includegraphics[width=2.0cm]{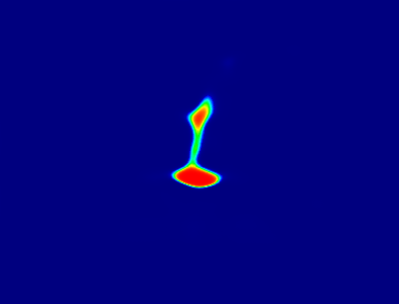}}}
        \put(0.0,9.3){\fbox{\includegraphics[width=2.0cm]{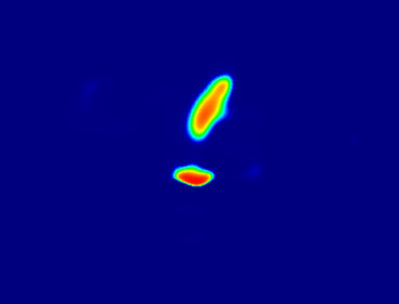}}}
        \put(0.0,7.75){\fbox{\includegraphics[width=2.0cm]{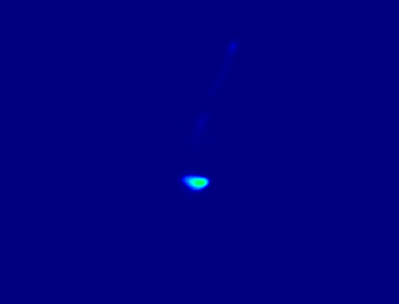}}}
        \put(0.0,6.2){\fbox{\includegraphics[width=2.0cm]{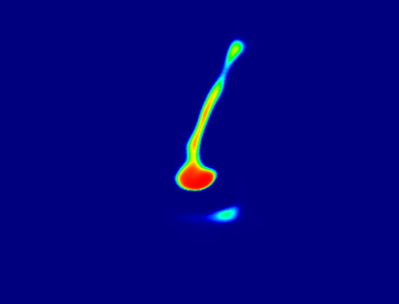}}}
        \put(0.0,4.65){\fbox{\includegraphics[width=2.0cm]{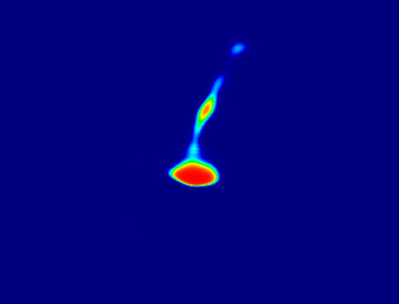}}}
        \put(0.0,3.1){\fbox{\includegraphics[width=2.0cm]{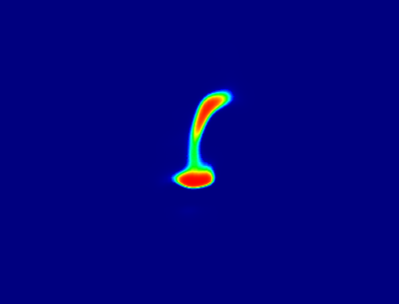}}}
        \put(0.0,1.55){\fbox{\includegraphics[width=2.0cm]{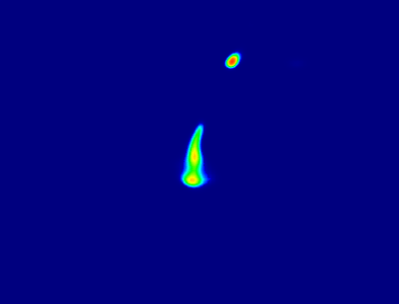}}}
        \put(0.0,0.0){\fbox{\includegraphics[width=2.0cm]{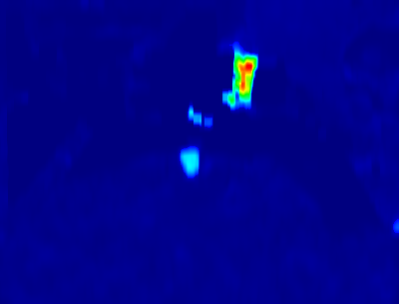}}}
        
        \put(2.0,13.95){\fbox{\includegraphics[width=2.0cm]{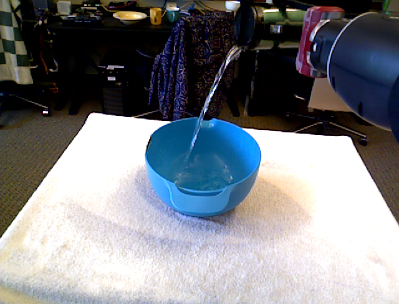}}}
        \put(2.0,12.4){\fbox{\includegraphics[width=2.0cm]{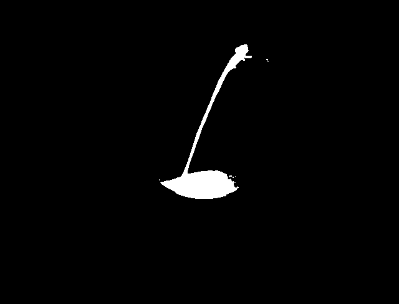}}}
        \put(2.0,10.85){\fbox{\includegraphics[width=2.0cm]{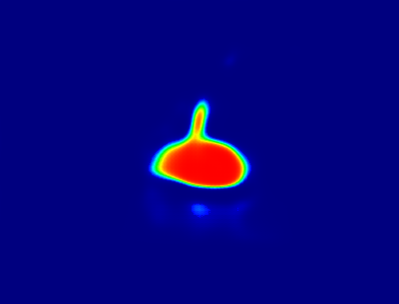}}}
        \put(2.0,9.3){\fbox{\includegraphics[width=2.0cm]{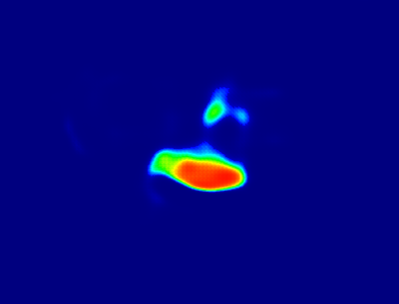}}}
        \put(2.0,7.75){\fbox{\includegraphics[width=2.0cm]{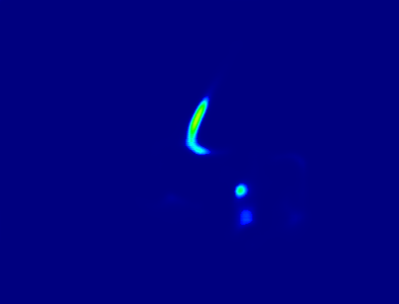}}}
        \put(2.0,6.2){\fbox{\includegraphics[width=2.0cm]{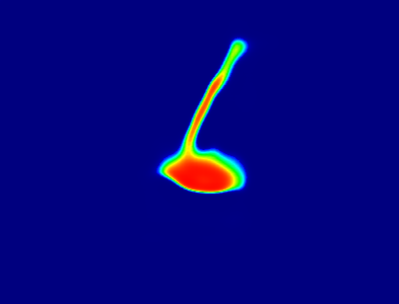}}}
        \put(2.0,4.65){\fbox{\includegraphics[width=2.0cm]{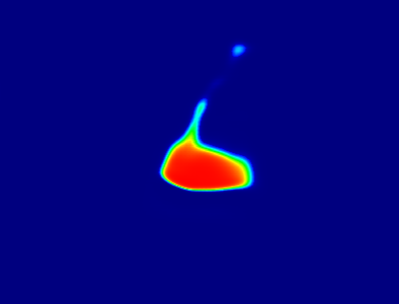}}}
        \put(2.0,3.1){\fbox{\includegraphics[width=2.0cm]{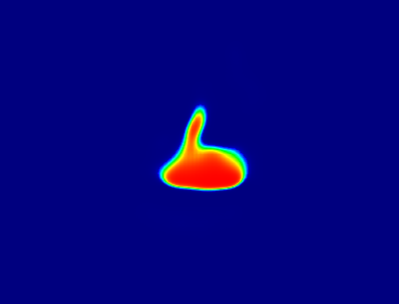}}}
        \put(2.0,1.55){\fbox{\includegraphics[width=2.0cm]{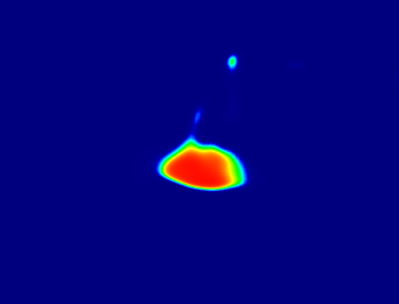}}}
        \put(2.0,0.0){\fbox{\includegraphics[width=2.0cm]{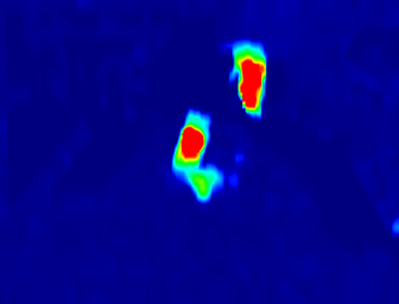}}}

        \put(4.0,13.95){\fbox{\includegraphics[width=2.0cm]{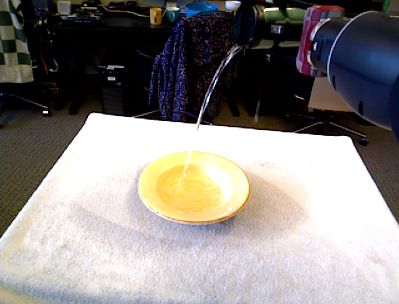}}}
        \put(4.0,12.4){\fbox{\includegraphics[width=2.0cm]{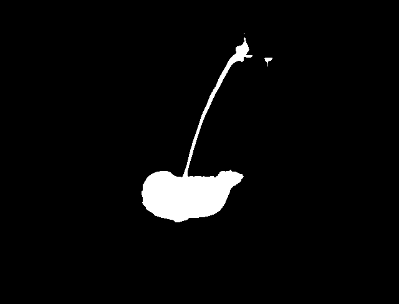}}}
        \put(4.0,10.85){\fbox{\includegraphics[width=2.0cm]{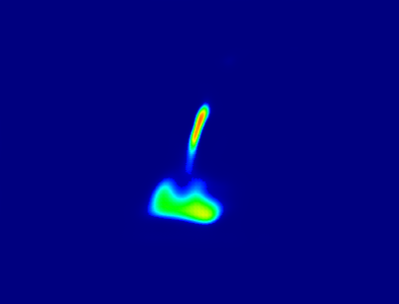}}}
        \put(4.0,9.3){\fbox{\includegraphics[width=2.0cm]{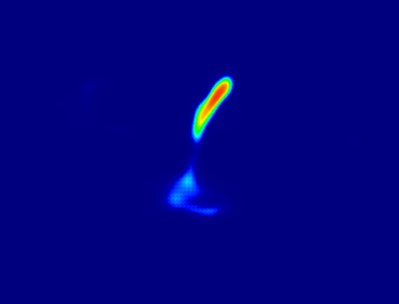}}}
        \put(4.0,7.75){\fbox{\includegraphics[width=2.0cm]{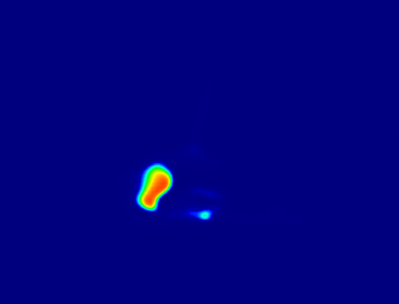}}}
        \put(4.0,6.2){\fbox{\includegraphics[width=2.0cm]{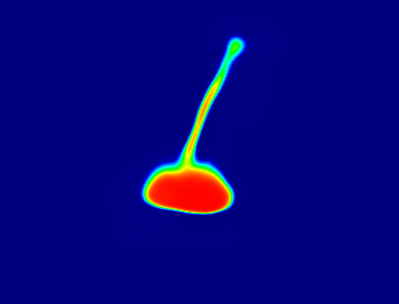}}}
        \put(4.0,4.65){\fbox{\includegraphics[width=2.0cm]{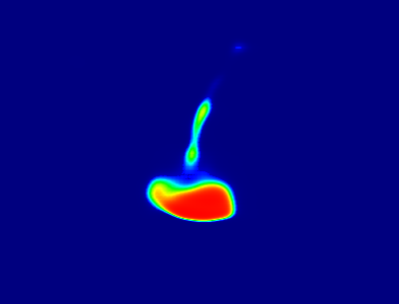}}}
        \put(4.0,3.1){\fbox{\includegraphics[width=2.0cm]{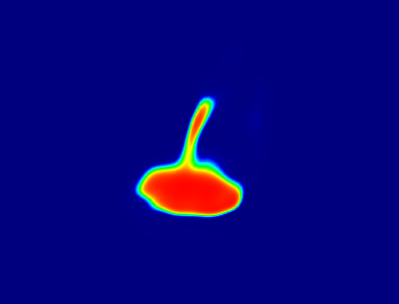}}}
        \put(4.0,1.55){\fbox{\includegraphics[width=2.0cm]{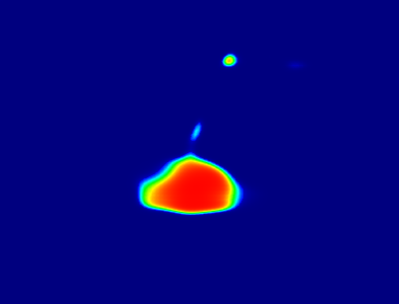}}}
        \put(4.0,0.0){\fbox{\includegraphics[width=2.0cm]{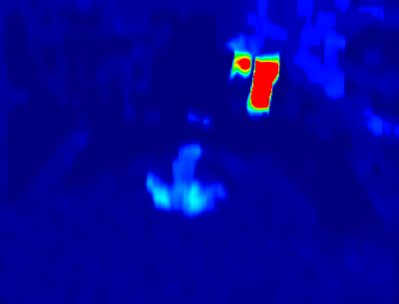}}}
        
        \put(6.0,13.95){\fbox{\includegraphics[width=2.0cm]{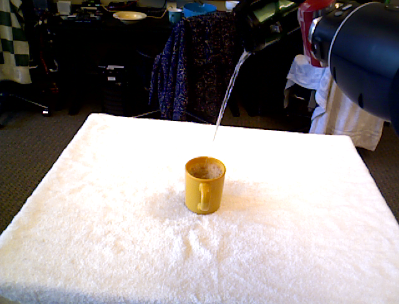}}}
        \put(6.0,12.4){\fbox{\includegraphics[width=2.0cm]{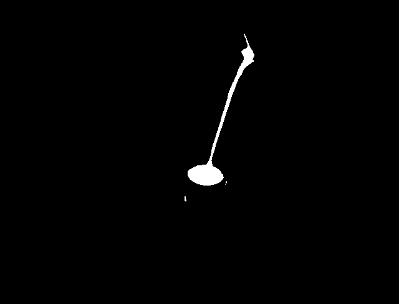}}}
        \put(6.0,10.85){\fbox{\includegraphics[width=2.0cm]{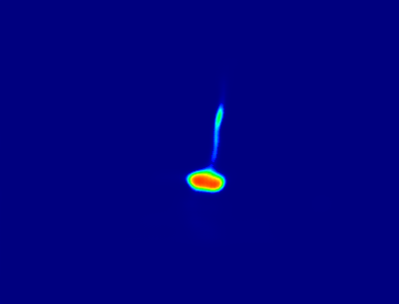}}}
        \put(6.0,9.3){\fbox{\includegraphics[width=2.0cm]{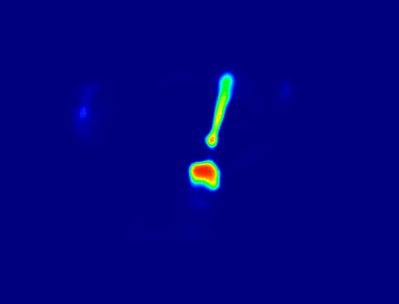}}}
        \put(6.0,7.75){\fbox{\includegraphics[width=2.0cm]{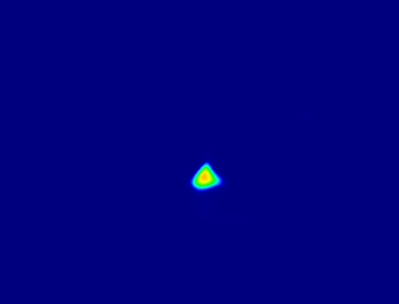}}}
        \put(6.0,6.2){\fbox{\includegraphics[width=2.0cm]{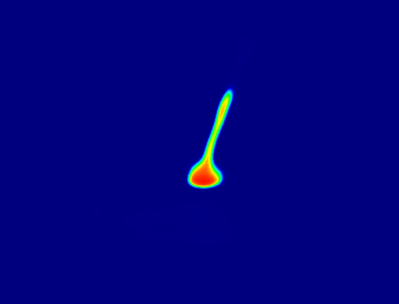}}}
        \put(6.0,4.65){\fbox{\includegraphics[width=2.0cm]{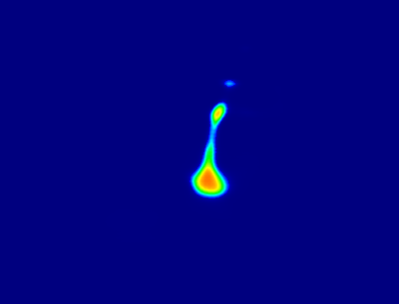}}}
        \put(6.0,3.1){\fbox{\includegraphics[width=2.0cm]{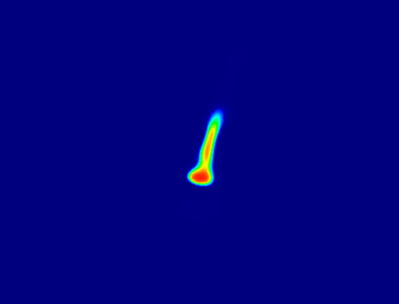}}}
        \put(6.0,1.55){\fbox{\includegraphics[width=2.0cm]{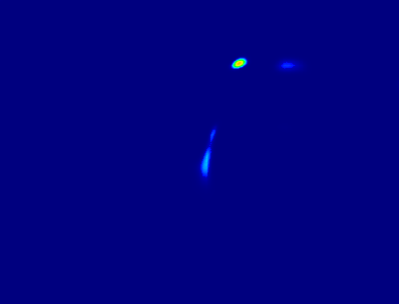}}}
        \put(6.0,0.0){\fbox{\includegraphics[width=2.0cm]{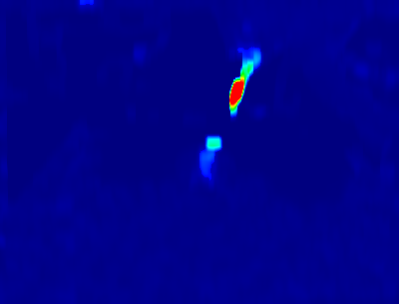}}}
        
        \put(7.75,15.5){\rotatebox{270}{\parbox{1.55cm}{\begin{center}{\bf Color}\end{center}}}}
        \put(7.75,13.95){\rotatebox{270}{\parbox{1.55cm}{\begin{center}{\bf Labels}\end{center}}}}
        \put(7.75,12.4){\rotatebox{270}{\parbox{1.55cm}{{\begin{center}\figfontsize RGB\end{center}}}}}
        \put(7.75,10.85){\rotatebox{270}{\parbox{1.55cm}{{\begin{center}\figfontsize Gray\end{center}}}}}

        \put(7.75,9.3){\rotatebox{270}{\parbox{1.55cm}{{\begin{center}\figfontsize RGB+Flow\\Early-Fus.\end{center}}}}}
        \put(7.75,7.75){\rotatebox{270}{\parbox{1.55cm}{{\begin{center}\figfontsize Gray+Flow\\Early-Fus.\end{center}}}}}
        \put(7.75,6.2){\rotatebox{270}{\parbox{1.55cm}{{\begin{center}\figfontsize RGB+Flow\\Late-Fus.\end{center}}}}}
        \put(7.75,4.65){\rotatebox{270}{\parbox{1.55cm}{{\begin{center}\figfontsize Gray+Flow\\Late-Fus.\end{center}}}}}
        \put(7.75,3.1){\rotatebox{270}{\parbox{1.55cm}{{\begin{center}\figfontsize Flow\end{center}}}}}
        \put(7.75,1.55){\rotatebox{270}{\parbox{1.55cm}{{\begin{center}\figfontsize Baseline\end{center}}}}}

    \end{picture}
    }
    \caption{Example frames for the LSTM-FCN for the {\it detection} task on the robot dataset with different types of input images. The first row shows the color image for reference and the row immediately below it shows the ground truth. All these images are from the test set with target object not seen in the train set. The last row shows the output of the baseline.}
    \label{fig:results_robot_detection1}
\end{figure}
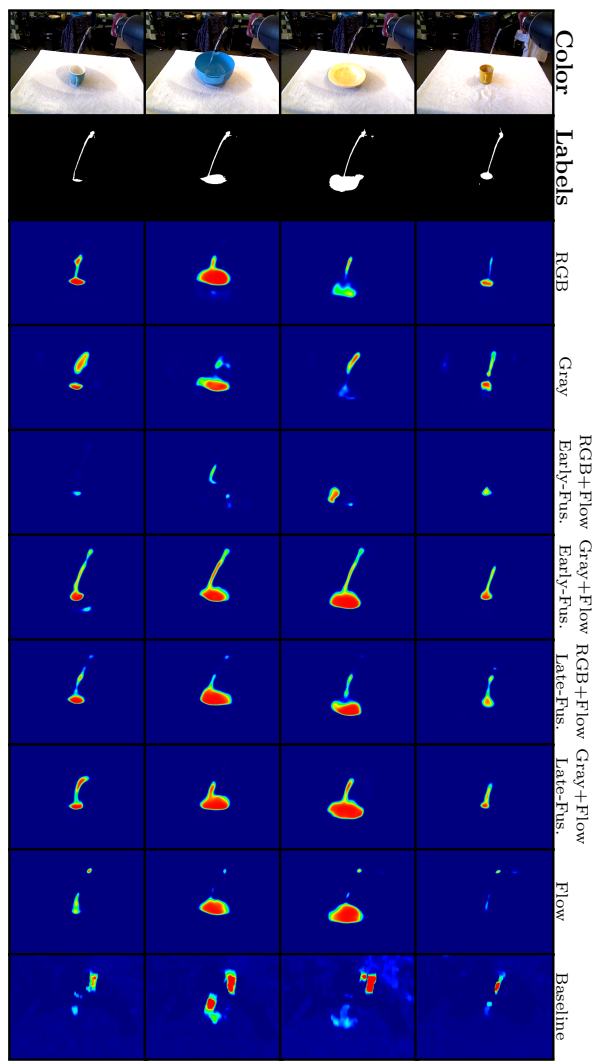

Figure \ref{fig:results_robot_detection1} shows example output on the test set of the LSTM-FCN with different types of input.
From this figure, it appears that the best performing network is the one that takes as input grayscale images plus optical flow with the early-fusion input block.
Indeed, the numbers in the table in Figure \ref{fig:results_robot_lstm} confirm this.
Interestingly, the grayscale + optical flow early-fusion network is the second worst performing network on the train set, but performs the best on the test set.
This suggests that the other networks tend to overfit more to the training distribution and as a result don't generalize to new data very well.

The table in Figure \ref{fig:results_robot_fcn} reflects a similar, albeit slightly different, result for single-frame FCNs.
While the grayscale plus optical flow early-fusion network has one of the highest performances on the test set, it is outperformed by the network that takes only optical flow as input.
As counter-intuitive as it may seem, this makes some sense.
The single-frame FCN does not have the ability to view any temporal information, however since optical flow is computed between two frames, it implicitly encodes temporal information in the input to the network.
As we saw in the section on detection on the simulated dataset, temporal information is very important for the {\it detection} task and the network that takes only optical flow is forced to only use temporal information, thus allowing it to generalize to new data better.
In the case of the LSTM-FCN, this effect is less pronounced because the network can store temporal information in its recurrent state, although performance of the optical flow only network is still better than performance of the networks that do not use optical flow in any way.

\subsubsection{Baseline Comparison}

We also computed the performance of the baseline method based on the methodology of \citep{yamaguchi2016}.
It achieved 5.9\% AUC on the training set and 8.3\% AUC on the testing set.
The last row of Figure \ref{fig:results_robot_detection1} shows some examples of the output of the baseline.
While it is clear from the figure that the baseline is at least somewhat able to detect liquid, it does not perform nearly as well as the neural network based methods.
However, it is important to note that this method was developed by \citep{yamaguchi2016} for a slightly different task in a slightly different environment and using stereo cameras rather than monocular, so it would not be expected to perform as well on this task.
Nonetheless, it still provides a good baseline to compare our methods against.

The biggest advantage of the baseline method over learning-based methods is it's resilience to overfitting due to its lack of trained parameters.
However, this lack of learning also means it can't adapt to the problem as well.
Inspired by the resilience of the baseline method, we combined it with our deep neural network architectures to soften the effect of overfitting while maintaining the adaptability of learning-based methods.
As shown in Figure \ref{fig:results_robot_lstm}, the methods using optical flow as an input tended to have a smaller disparity between their training set and testing set performance.
While this didn't completely alleviate all overfitting, it is clear that combining these two methods is superior to using either alone.

\subsubsection{Initializing on Simulated Data}

\begin{figure}
    \centering
    \setlength{\unitlength}{1.0cm}
   
    \begin{picture}(8.0,6.0)
        \put(0.0,0.0){\includegraphics[width=8.0cm]{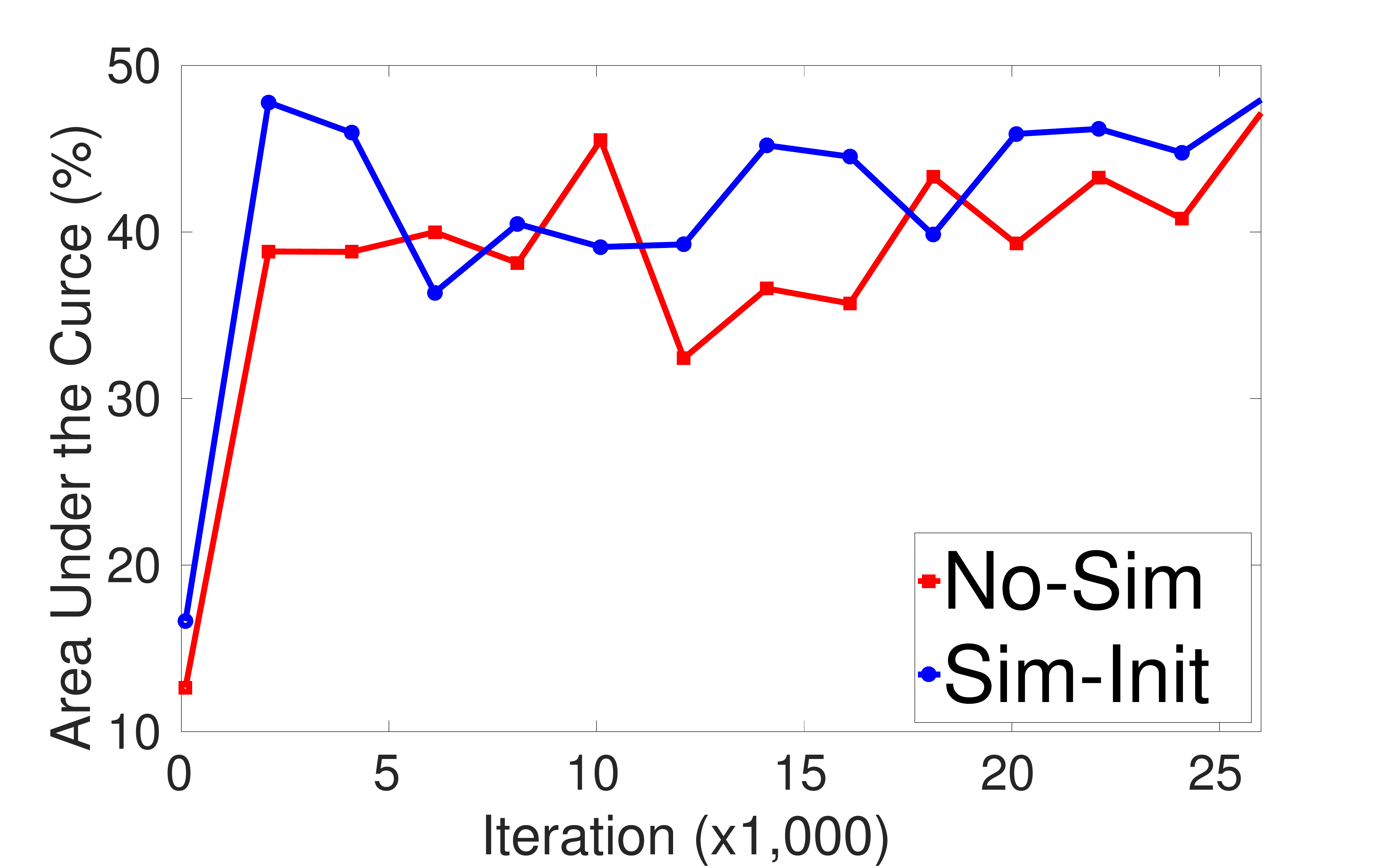}}
    \end{picture}
        
    \caption{The area under the curve on the test set at each iteration of training. The red line shows the performance of the LSTM-FCN trained solely on the robot dataset; the blue line shows the performance of the LSTM-FCN initialized from the simulated dataset.}
    \label{fig:nosim_vs_siminit}
\end{figure}

We evaluated whether or not the simulated data set, with its larger size, could be used to pre-train the weights of a network that would then be trained on the robot data.
Since the LSTM-FCN with grayscale plus optical flow early-fusion as input generalized the best in the previous section, we trained another LSTM-FCN on the same type of input. 
However, instead of pre-training it on cropped images from the robot dataset, we went through the full training process for a detection network on the simulated dataset, and used those weights to initialize this network, which was then trained on the robot dataset.
The networks converged to the same performance after 61,000 iterations of training.
Figure \ref{fig:nosim_vs_siminit} shows the performance of both the network not initialized with any simulated data as compared to the performance of this network on the test set at each iteration.
The network initialized with simulated data does seem to converge slightly faster, although not by a large amount.

\subsection{Tracking Revisited}

\begin{figure}
    \setlength{\unitlength}{1.0cm}
   
    \begin{subfigure}{\textwidth}
    \begin{picture}(8.0,6.0)
        \put(0.0,0.0){\includegraphics[width=8.0cm]{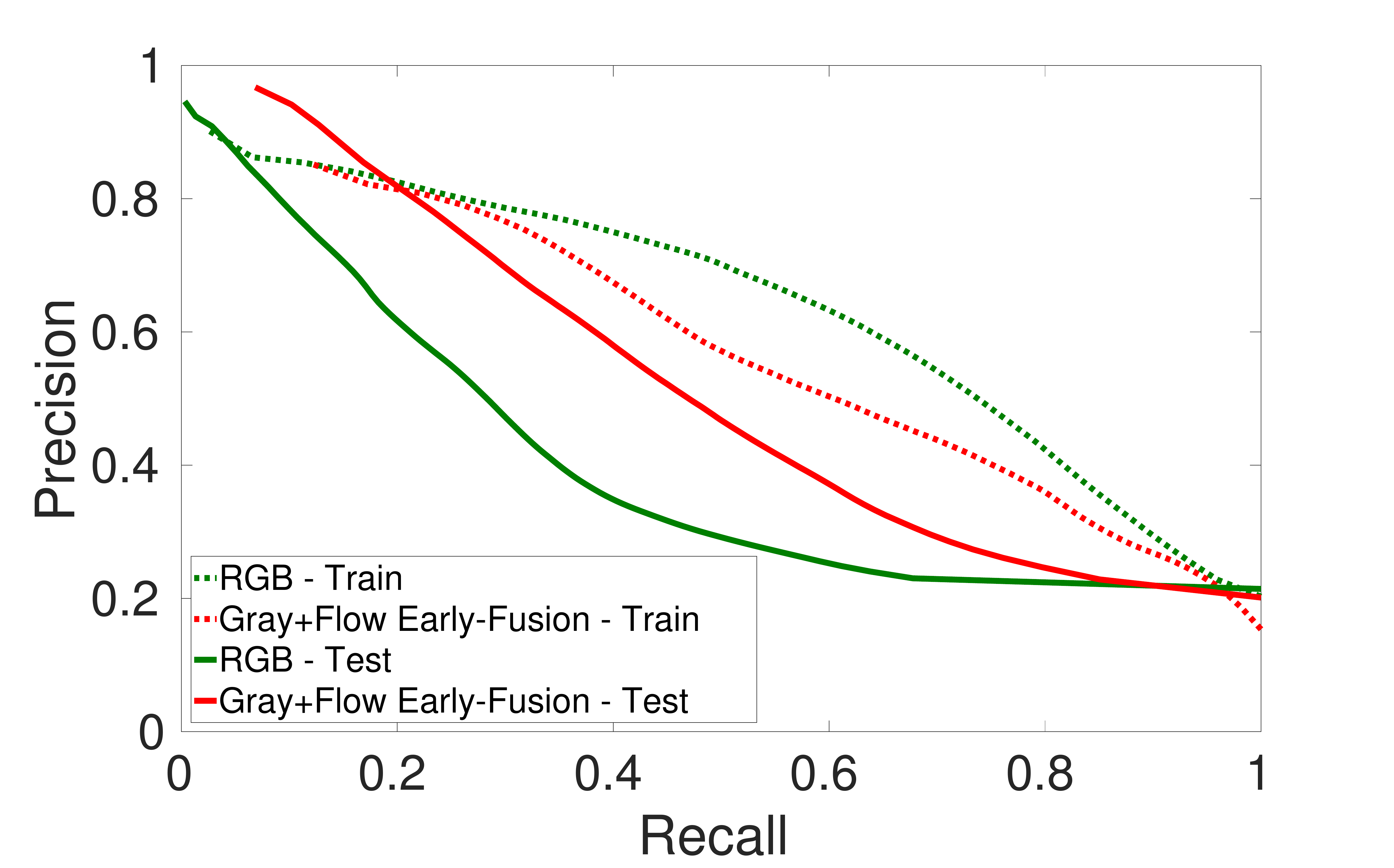}}
    \end{picture}
    \caption{Precision-Recall}
    \end{subfigure}
    
    \begin{subfigure}{\textwidth}
        \begin{tabular}{r | c | c |}
        & Train & Test \\\hline
        {\bf RGB} & 63.6\% & 40.1\% \\\hline
        {\bf Gray+Flow Early-Fusion} & 57.6\% & 52.2\% \\\hline
        \end{tabular}
        \caption{AUC}
    \end{subfigure}
    
    \caption{The performance on the combined {\it detection \& tracking} task of the LSTM-FCN that takes as input RGB images compared to the performance of the LSTM-FCN that takes as input grayscale images plus optical flow using the early-fusion input block. The upper graph shows the precision-recall plot for both networks on both the train and test sets. The lower table shows the corresponding area under the curve for each curve. Note that here we don't use any slack unlike in Figures \ref{fig:results_sim_detection2} and \ref{fig:results_sim_tracking} (equivalent to a slack of 0).}
    \label{fig:combined_grayflow_vs_rgb}
\end{figure}

The prior section showed that an LSTM-FCN taking as input grayscale early-fused with optical flow has the ability to generalize better than any other type of network we evaluated.
This result was achieved on the {\it detection} task, and we wanted to see if this translates to the {\it tracking} task.
However, the robot data set does not contain the ground truth for tracking, so we return to the simulated dataset to test this hypothesis.

We train two networks: one that takes the default input of RGB images and one that takes grayscale images early-fused with optical flow.
We train them on the combined {\it detection} \& {\it tracking} task.
They are trained in the same manner as described previously for doing combined {\it detection} \& {\it tracking}.
The advantage of using this alternative input type is its ability to generalize to new data, so we hold out all pouring sequences with one of the target containers (the {\it dog dish}) during training.
This includes during training of all pre-trained networks such that the final weights of the networks were never influenced by any data containing the test object.

Figure \ref{fig:combined_grayflow_vs_rgb} shows the performance of the two networks on both the train and test sets.
From this figure it is clear that the RGB network outperforms the other on the training set, however, the gray+flow early-fusion network outperforms the RGB network on the test set.
This confirms the results we found in the previous section: Networks trained with grayscale early-fused with optical flow generalize better to new situations.

\section{Application to a Control Task}

\begin{figure}
\begin{center}
\scalebox{0.5}{
\begin{tikzpicture}[->,>=stealth,auto,node distance=1.6cm,thick,
      input node/.style={rectangle,draw,anchor=west,align=center,inner sep=0,outer sep=0},
      output node/.style={draw=none,anchor=west,align=center,inner sep=0,outer sep=0},
      conv node/.style={rectangle,fill=red!60,draw,font=\sffamily\scriptsize\bfseries,align=center,anchor=west,minimum height=1.5cm},
      blob node/.style={ellipse,fill=gray!20,draw,font=\sffamily\scriptsize\bfseries,align=center,inner sep=0},
      elps node/.style={fill=none,draw=none,font=\sffamily\LARGE\bfseries}]
      
      %%%%%%%%%%%%%%%%%%% Detection net %%%%%%%%%%%%%%%%%%%%%
      \node[input node] (lstm_In3) at (0.0,9.0) {\includegraphics[width=2cm]{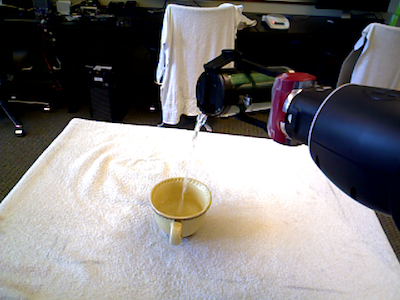}};
      \node[input node] (lstm_Rec1) at (0.0,7.225) {\includegraphics[width=2cm]{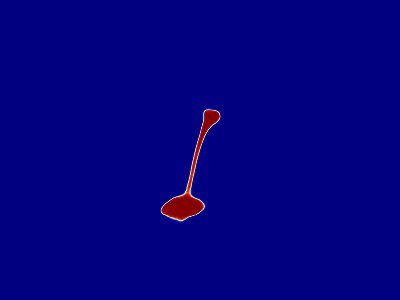}};

      \node[conv node] (lstm_Conv1) at (2.2,8.975) {\rotatebox{270}{\parbox[c]{1.45cm}{\centering Convolution\\{\normalfont\tiny\it 32 $5{\times}5$ kernels}}}};
      \node[conv node] (lstm_Conv2) at (3.1,8.975) {\rotatebox{270}{\parbox[c]{1.45cm}{\centering Convolution\\{\normalfont\tiny\it 32 $5{\times}5$ kernels}}}};
      \node[conv node] (lstm_Conv3) at (4.0,8.975) {\rotatebox{270}{\parbox[c]{1.45cm}{\centering Convolution\\{\normalfont\tiny\it 32 $5{\times}5$ kernels}}}};
      \node[conv node] (lstm_Conv4) at (4.9,8.975) {\rotatebox{270}{\parbox[c]{1.45cm}{\centering Convolution\\{\normalfont\tiny\it 32 $5{\times}5$ kernels}}}};
      \node[conv node] (lstm_Conv5) at (5.8,8.975) {\rotatebox{270}{\parbox[c]{1.7cm}{\centering Convolution\\{\normalfont\tiny\it 32 $17{\times}17$ kernels}}}};
      
      \node[conv node] (lstm_rec_conv1) at (2.2,7.2) {\rotatebox{270}{\parbox[c]{1.45cm}{\centering Convolution\\{\normalfont\tiny\it 20 $5{\times}5$ kernels}}}};
      \node[conv node] (lstm_rec_conv2) at (3.1,7.2) {\rotatebox{270}{\parbox[c]{1.45cm}{\centering Convolution\\{\normalfont\tiny\it 20 $5{\times}5$ kernels}}}};
      \node[conv node] (lstm_rec_conv3) at (4.0,7.2) {\rotatebox{270}{\parbox[c]{1.45cm}{\centering Convolution\\{\normalfont\tiny\it 20 $5{\times}5$ kernels}}}};
      
      \node[conv node] (lstm_lstm1) at (7.0,8.0) [fill=green!60]{\parbox[c]{0.65cm}{\centering LSTM\\{\normalfont\tiny\it 20~$1{\times}1$ kernels \\\vspace{-0.1cm}per~gate}}};
      \node[conv node] (lstm_fc_conv1) at (8.4,8.0) [fill=blue!60]{\rotatebox{270}{\parbox[c]{2.1cm}{\centering $\mathbf{1{\times}1}$ Convolution\\{\normalfont\tiny\it 64 kernels}}}};
      \node[conv node] (lstm_deconv) at (9.3,8.0) [fill=orange!60]{\rotatebox{270}{\parbox[c]{2.0cm}{\centering $Conv^\top$\\{\normalfont\tiny\it 64 $16{\times}16$ kernels}}}};
      \node[input node] (lstm_Out1) at (10.4,8.0) {\includegraphics[width=2cm]{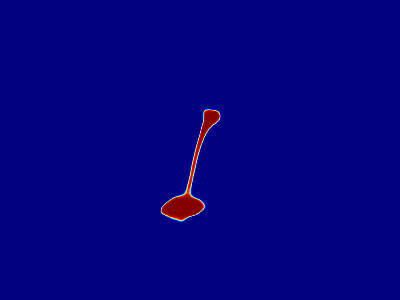}};

      \node[blob node] (lstm_rec_in2) at (6.0, 7.0) {Recurrent\\State};
      \node[blob node] (lstm_rec_in3) at (7.45, 6.7) {Cell\\State};
      
      \node[blob node] (lstm_rec_out2) at (8.7, 9.5) {Recurrent\\State};
      \node[blob node] (lstm_rec_out3) at (7.45, 9.3) {Cell\\State};
      
      \draw (lstm_In3) -- (lstm_Conv1);
      \draw (lstm_Conv1) -- (lstm_Conv2);
      \draw (lstm_Conv2) -- (lstm_Conv3);
      \draw (lstm_Conv3) -- (lstm_Conv4);
      \draw (lstm_Conv4) -- (lstm_Conv5);
      
      \draw (lstm_Rec1) -- (lstm_rec_conv1);
      \draw (lstm_rec_conv1) -- (lstm_rec_conv2);
      \draw (lstm_rec_conv2) -- (lstm_rec_conv3);

      \draw (lstm_Conv5.east) -- (lstm_lstm1.west);
      \draw (lstm_rec_conv3.east) -- (lstm_lstm1.west);
      
      \draw (lstm_lstm1) -- (lstm_fc_conv1);
      \draw (lstm_fc_conv1) -- (lstm_deconv);
      \draw (lstm_deconv) -- (lstm_Out1);
      
      \draw (lstm_rec_in2) -- (lstm_lstm1.west);

      \draw (lstm_rec_in3) -- (lstm_lstm1.south);
      \draw (lstm_lstm1.east) -- (lstm_rec_out2.215);
      \draw (lstm_lstm1.north) -- (lstm_rec_out3);

    %%%%%%%%%%%%%%%%% MF-CNN %%%%%%%%%%%%%%%%%%%
      \node[conv node] (concat) at (7.5,3.525) [fill=gray!60]{\rotatebox{270}{\parbox[c]{2.0cm}{\centering Concatenation\\{\normalfont\tiny\it channel-wise}}}};
      \node[conv node] (Conv6) at (8.5,3.525) {\rotatebox{270}{\parbox[c]{1.45cm}{\centering Convolution\\{\normalfont\tiny\it 32 $5{\times}5$ kernels}}}};
      \node[conv node] (fc_conv1) at (9.5,3.525) [fill=blue!60]{\rotatebox{270}{\parbox[c]{2.0cm}{\centering Fully Connected\\{\normalfont\tiny\it 256 nodes}}}};
      \node[conv node] (fc_conv2) at (10.5,3.525) [fill=blue!60]{\rotatebox{270}{\parbox[c]{2.0cm}{\centering Fully Connected\\{\normalfont\tiny\it 256 nodes}}}};
      \node[conv node] (fc_conv3) at (11.5,3.525) [fill=blue!60]{\rotatebox{270}{\parbox[c]{2.0cm}{\centering Fully Connected\\{\normalfont\tiny\it 100 nodes}}}};

      \node[output node] (Out1) at (12.5,3.525) {\includegraphics[width=2cm]{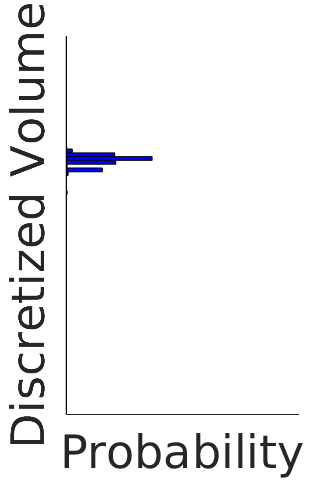}};

      \node[input node] (In3a) at (0.0,4.0) {\includegraphics[width=2cm]{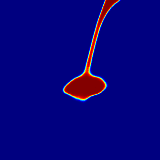}};
      \node[conv node] (Conv1a) at (2.2,3.975) {\rotatebox{270}{\parbox[c]{1.45cm}{\centering Convolution\\{\normalfont\tiny\it 32 $5{\times}5$ kernels}}}};
      \node[conv node] (Conv2a) at (3.1,3.975) {\rotatebox{270}{\parbox[c]{1.45cm}{\centering Convolution\\{\normalfont\tiny\it 32 $5{\times}5$ kernels}}}};

      \node[conv node] (Conv3a) at (4.0,3.975) {\rotatebox{270}{\parbox[c]{1.45cm}{\centering Convolution\\{\normalfont\tiny\it 32 $5{\times}5$ kernels}}}};
      \node[conv node] (Conv4a) at (4.9,3.975) {\rotatebox{270}{\parbox[c]{1.45cm}{\centering Convolution\\{\normalfont\tiny\it 32 $5{\times}5$ kernels}}}};
      \node[conv node] (Conv5a) at (5.8,3.975) {\rotatebox{270}{\parbox[c]{1.7cm}{\centering Convolution\\{\normalfont\tiny\it 32 $17{\times}17$ kernels}}}};
      \draw (In3a) -- (Conv1a);

      \draw (Conv1a) -- (Conv2a);
      \draw (Conv2a) -- (Conv3a);
      \draw (Conv3a) -- (Conv4a);
      \draw (Conv4a) -- (Conv5a);
      \draw (Conv5a) -- (concat);

      \node (elps_node) at (0.22,2.5) [fill=none,draw=none]{\rotatebox{295}{............}};

      \node[input node] (In3b) at (0.5,3.0) {\includegraphics[width=2cm]{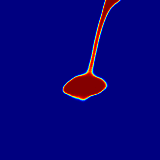}};
      \node[conv node] (Conv1b) at (2.7,3.075) {\rotatebox{270}{\parbox[c]{1.45cm}{\centering Convolution\\{\normalfont\tiny\it 32 $5{\times}5$ kernels}}}};
      \node[conv node] (Conv2b) at (3.6,3.075) {\rotatebox{270}{\parbox[c]{1.45cm}{\centering Convolution\\{\normalfont\tiny\it 32 $5{\times}5$ kernels}}}};
      \node[conv node] (Conv3b) at (4.5,3.075) {\rotatebox{270}{\parbox[c]{1.45cm}{\centering Convolution\\{\normalfont\tiny\it 32 $5{\times}5$ kernels}}}};

      \node[conv node] (Conv4b) at (5.4,3.075) {\rotatebox{270}{\parbox[c]{1.45cm}{\centering Convolution\\{\normalfont\tiny\it 32 $5{\times}5$ kernels}}}};
      \node[conv node] (Conv5b) at (6.3,3.075) {\rotatebox{270}{\parbox[c]{1.7cm}{\centering Convolution\\{\normalfont\tiny\it 32 $17{\times}17$ kernels}}}};
      \draw (In3b) -- (Conv1b);
      \draw (Conv1b) -- (Conv2b);
      \draw (Conv2b) -- (Conv3b);

      \draw (Conv3b) -- (Conv4b);
      \draw (Conv4b) -- (Conv5b);
      \draw (Conv5b) -- (concat);

      \draw (concat) -- (Conv6);
      \draw (Conv6) -- (fc_conv1);
      \draw (fc_conv1) -- (fc_conv2);
      \draw (fc_conv2) -- (fc_conv3);
      \draw (fc_conv3) -- (Out1);
      %\draw[draw=black] (0,2) -- (14.5,2) -- (14.5,5.0) -- (0.0,5.0) -- (0.0,2.0); 
      
      %%%%%%%%%%%%% Other stuff %%%%%%%%%%%%%%%%  
      
      \node[conv node] (crop) at (5.0,5.7) [fill=magenta!60,minimum height=0.7cm]{\parbox[c]{2.0cm}{\centering Crop\\{\normalfont\tiny\it $160{\times}160$}}};
      \draw[line width=0.1cm] (lstm_Out1.south) to [out=270,in=0] (crop.east);
      \draw[line width=0.1cm] (crop.west) to [out=180,in=90] (In3a.north);
      
      \node[conv node] (hmm) at (4.0,0.0) [fill=cyan!60,minimum height=0.7cm]{\parbox[c]{2.0cm}{\centering HMM}};
      \node[conv node] (pid) at (8.0,0.0) [fill=violet!60,minimum height=0.7cm]{\parbox[c]{2.0cm}{\centering PID \\Controller}};
      \node[draw=none] (control) at (12.0,0.0) {\parbox[c]{1.2cm}{\centering Robot\\Control\\Signal}};
      \draw[line width=0.1cm] (Out1.south) to [out=270,in=90] (hmm.north);
      \draw[line width=0.1cm] (hmm.east) -- (pid.west);
      \draw[line width=0.1cm] (pid.east) -- (control.west);
      \draw[line width=0.1cm] (hmm.east) -- (7.0,0.0) -- (7.0,-1.0) -- (3.0,-1.0) -- (3.0,0.0) -- (hmm.west);
      \node[draw=none] at (4.7,0.7) {{\bf\Large$z_t$}}; 
      \node[draw=none] at (6.6,0.4) {{\bf\Large$v_t$}};
      
\end{tikzpicture}
}

\caption{The entire robot control system using the recurrent neural network for detections and the multi-frame network for volume estimation. The recurrent detection network (top) takes both the color image and its own detections from the previous time step and produces a liquid detection heatmap. The multi-frame network (center) takes a sequence of detections cropped around the target container and outputs a distribution over volumes in the container. The output of this network is fed into a HMM, which estimates the volume of the container. This is passed into a PID controller, which computes the robot's control signal.}
\label{fig:icra_model}
\end{center}

\end{figure}
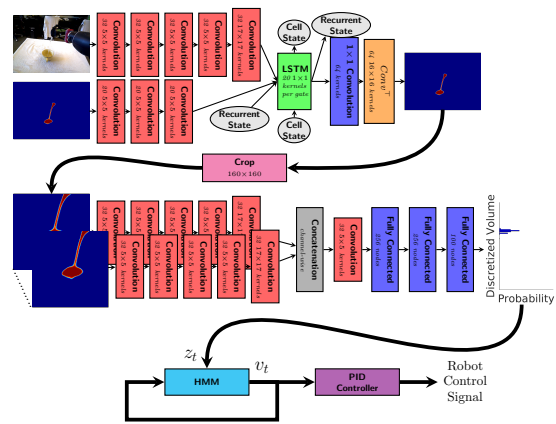

We applied our methodology described in this paper to a robotic control task involving liquids.
This application is described in our concurrent work \citep{schenckc2016c}.
We briefly summarize it here to illustrate the efficacy of our method.
Please refer to that paper for more details.

\subsection{Task}

We utilize our liquid detection framework described here as input to a control algorithm for the pouring task.
The robot's goal was to pour a specific amount of liquid from a source container into a target container using only its RGBD camera as sensory input.
The robot was given a target amount in milliliters and a source container with an unknown initial amount of liquid (but always more than the target).
The robot then used visual closed-loop feedback to pour liquid from the source into the target until the correct amount was reached.

\subsection{Methodology}

\begin{figure}
    \centering
    \setlength{\unitlength}{1.0cm}
    \begin{picture}(7.0,5.5)
        \put(0.0,0.0){\includegraphics[width=7.0cm]{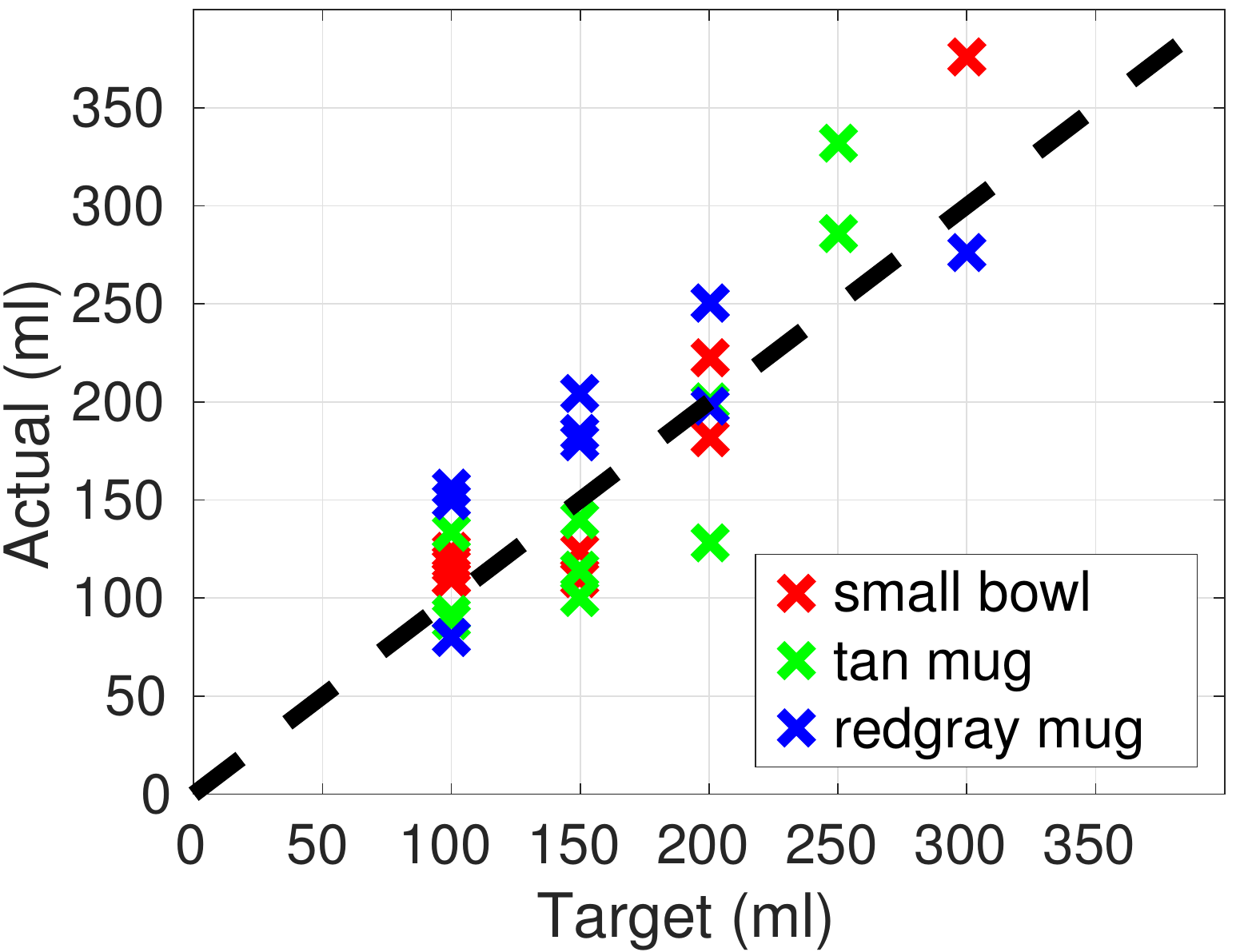}}
    \end{picture}
    \caption{Plot of the result of each pour. The x-axis is the target amount that the robot was attempting to reach, and the y-axis is the actual amount the robot poured. The points are color-coded by the target container. The black dashed line shows a 1:1 correspondence for reference.}
    \label{fig:control_points}
\end{figure}

Figure \ref{fig:icra_model} shows the robot control system.
The robot's gripper with the source container was fixed over the target and it controlled the angle of its wrist to control the angle of the container.
First, we trained a LSTM-FCN to classify pixels as {\it liquid} or {\it not-liquid} from RGB images\footnote{Since this work was concurrent, we did not yet have the results indicating that grayscale images with optical flow performed better than RGB images for detection, and so we used RGB images as the default.}.
The heatmap from this network was cropped around the target container and then passed to another neural network.
This network was a multi-frame network that was trained to take in a series of sequential inputs and output the amount of liquid in milliliters in the target container.
The output of this network is a distribution over the liquid volume.
This distribution was then passed to a hidden Markov model (HMM) that modeled the change in volume over time.
The probability distribution over the liquid volume was represented as a histogram over a set of uniform, discrete bins and the HMM smoothed this distribution's changes over time.
Finally, the difference between the target volume and the volume in the HMM was used by a PID controller to control the pour.
The PID controller adjusted the angular velocity of the robot's wrist joint to control the flow of liquid leaving the source container, returning to upright when the volume reached the target amount.

\subsection{Results}

We evaluated our method on 30 pouring tasks using the objects in Figures \ref{fig:small_bowl}, \ref{fig:tan_mug}, and \ref{fig:redgray_mug} as the target container and the mug in Figure \ref{fig:mug} as the source container.
For each pour, we randomly selected an initial amount of liquid in the source between 300 and 400 ml and a target between 100 and 300 ml (always ensuring at least a 100 ml difference).
At the end of each pour, we compared the target volume given to the robot with the actual volume as measured by a scale.
Note that the our methodlogy here is able to run in real-time (approximately 30 Hertz) on a computer with a modern GPU.

Figure \ref{fig:control_points} shows the results of each pour.
The robot had an average deviation from its target of only 38 ml.
While this may not be sufficient for high precision tasks such as tasks in a wetlab, this is approximately the precision expected in common household tasks such as cooking.
This shows that our liquid perception and reasoning methods developed in this paper are indeed precise and reliable enough to be utilized online in a control task involving liquids, not just for processing data offline.

\section{Conclusion}
\label{sec:conclusion}

In this paper, we showed how a robot can solve the tasks of detection and tracking liquids using deep learning. We evaluated 3 different network architectures, FCN, MF-FCN, and LSTM-FCN, all of which integrated different amounts of temporal information. We also evaluated eight different types of input images to our networks, including RGB and grayscale combined with optical flow. We tested these networks on both data we generated in a realistic liquid simulator and on data we collected from a real robot.

Our results clearly show that integrating temporal information is crucial for perceiving and reasoning about liquids. The multi-frame FCN was able to outperform the single-frame FCN because it incorporated a window of frames, giving it more temporal information.  Furthermore, the LSTM-FCN is able to learn to remember relevant information in its recurrent state, enabling it to outperform the MF-FCN since it keeps information much longer than the fixed window of the MF-FCN. This was true not only for the task of tracking, which requires a notion of memory, but also for the task of detection.

The results also showed that, for the purposes of generalizing to new objects and settings, standard RGB images lead to overfitting and are not as well suited as images converted to grayscale and early-fused with optical flow. Networks trained on RGB images tended to perform very well on sequences drawn from the same distribution as their training set, but their performance dropped considerably when those sequences were drawn from a slightly different setting. However, while networks trained on grayscale early-fused with optical flow did not reach the same level of performance on data taken from the training distribution, their generalization to new settings was significantly better.

Beyond merely demonstrating that these methods work on offline datasets, we also showed results from our concurrent work \citep{schenckc2016c} in which we apply them to a control task. Specifically, we showed how we can combine our deep learning methods with a relatively simple controller to achieve robust results on a robot pouring task. Our robot was able to pour accurate amounts of liquid using only color images for closed-loop feedback. This clearly shows that the methods we describe in this paper are applicable to real, online robot control tasks.

The contributions of this paper are as follows. First, we showed how deep learning can be applied to address the challenging perception task of liquid detection and tracking in the context of pouring. Second, we introduced a novel technique using a thermographic camera and hot water to automatically generate ground truth labels for our real robot dataset. Third, we investigated different deep network structures and showed through experimental evaluation how different types and combinations of inputs affect a network’s ability to solve the detection task. Finally, we showed how our methodology can be applied to a control task on a real robot.

This paper also introduced a new dataset, the University of Washington Liquid Pouring Dataset (UW-LPD), which we make available to the wider research community via the following url: \url{http://rse-lab.cs.washington.edu/lpd/}. The total size of all the data collected is approximately 2.5 terabytes.

This paper opens up various avenues for future work. So far, our deep learning only enables reasoning about liquids in 2D, rather then the 3D volumetric space.  A next step for future work would be to look at ways that enable robots to reason about liquids in full 3D and take advantage of that to do more complex manipulation tasks. One possible direction is to connect the 2D liquid detection introduced in this paper to 3D fluid simulation, as we showed in our initial work on closed-loop simulation~\citep{schenckc2017}.  A promising alternative would be to incorporate fluid simulation into a deep network structure, performing volumetric reasoning using a convolutional structure.  Another avenue for future work is to investigate more ways for networks to generalize to new data. In this paper the test data, while different from the training data, was still collected in the same environment with the same setup. Future work will examine methods for training networks to generalize to different types of liquids across many different environments with many different conditions.

\bibliographystyle{SageH}
\bibliography{ijrr2017}

%\clearpage
%\includepdf[pages=-]{ijrr2017.pdf}

\end{document}